\documentclass[11pt, a4paper]{article}

\usepackage[a4paper, total={170mm,257mm}, left=20mm, top=20mm]{geometry}
\usepackage{newtxtext}
\usepackage{newtxmath}
\usepackage{graphicx}
\usepackage{siunitx}
\usepackage{float}
\usepackage{caption}
\usepackage{subcaption}
\usepackage{titlesec}
\usepackage{fancyhdr}
\usepackage{enumitem}
\usepackage{booktabs}
\usepackage{multicol}
\usepackage[utf8]{inputenc}
\usepackage[T1]{fontenc}
\usepackage{textcomp}
\usepackage{csquotes}
\usepackage{tikz}
\usetikzlibrary{positioning, arrows.meta, shapes.geometric, fit, backgrounds}
\usepackage{microtype}
\usepackage{algorithm}
\usepackage{algpseudocode}
\usepackage{tabularx}

\makeatletter
\newenvironment{breakablealgorithm}
  {\begin{center}
     \refstepcounter{algorithm}
     \hrule height.8pt depth0pt \kern2pt
     \renewcommand{\caption}[2][\relax]{%
       {\raggedright\textbf{\ALG@name~\thealgorithm} ##2\par}%
       \ifx\relax##1\relax
         \addcontentsline{loa}{algorithm}{\protect\numberline{\thealgorithm}##2}%
       \else
         \addcontentsline{loa}{algorithm}{\protect\numberline{\thealgorithm}##1}%
       \fi
       \kern2pt\hrule\kern2pt}
  }{\kern2pt\hrule\relax\end{center}}
\makeatother

\usepackage{bm}
\usepackage[colorlinks=true, linkcolor=blue!50!black, citecolor=blue!50!black, urlcolor=blue!50!black]{hyperref}

\newcommand{\vect}[1]{\vec{\bm{#1}}}
\newcommand{\hatt}[1]{\hat{\bm{#1}}}

\titleformat{\section}
  {\bfseries\LARGE}
  {\thesection.}{0.8em}{}

\titleformat{\subsection}
  {\bfseries\Large}
  {\thesubsection}{0.6em}{}

\titleformat{\subsubsection}
  {\bfseries\itshape\normalsize}
  {\thesubsubsection}{0.6em}{}

\titlespacing*{\section}{0pt}{1.2em plus 0.5em minus 0.3em}{0.8em}
\titlespacing*{\subsection}{0pt}{1em plus 0.3em minus 0.2em}{0.5em}
\titlespacing*{\subsubsection}{0pt}{0.8em plus 0.3em minus 0.2em}{0em}

\title{\vspace{-1cm}\textbf{Technical Report of Auriga's 1:10 Scale Autonomous Vehicle Robot}}

\author{
\begin{minipage}[t]{0.95\textwidth}\centering
Marya Sabouri Dodaran,
Ali Taheri Garakani,
Mandana Shabanzadeh Nasrollah-Abadi,
Yousef Ashrafi,
Mohammad Mahdi Azimi,
Erfan Nekouei Ghahfarokhi,
Amirhossein Kheiri Holighi,
Seyyed Sobhan Hosseini Hajibakandeh,
Matin Jamshidi,
Amirhossein Gholizadeh Behbahani,
Mohammad Hossein Moaiyeri\\[1em]
\textit{Team Auriga, Faculty of Electrical Engineering, Shahid Beheshti University, Tehran, Iran}
\end{minipage}
}
\date{November, 2026}

\begin{document}
\maketitle
\thispagestyle{empty}

\vspace{-1.2cm}
\begin{center}
\small
\texttt{
\href{mailto:h_moaiyeri@sbu.ac.ir}{h\_moaiyeri@sbu.ac.ir}, \\
\{\href{mailto:m.sabouridodaran@mail.sbu.ac.ir}{m.sabouridodaran},
\href{mailto:a.taherigarakani@mail.sbu.ac.ir}{a.taherigarakani},
\href{mailto:m.shabanzadeh@mail.sbu.ac.ir}{m.shabanzadeh},
\href{mailto:y.ashrafi@mail.sbu.ac.ir}{y.ashrafi},
\href{mailto:moham.azimi@mail.sbu.ac.ir}{moham.azimi},
\href{mailto:e.nekouei@mail.sbu.ac.ir}{e.nekouei},
\href{mailto:a.kheiriholighi@mail.sbu.ac.ir}{a.kheiriholighi},
\href{mailto:s.hoseinihajibekande@mail.sbu.ac.ir}{s.hoseinihajibekande},
\href{mailto:mat.jamshidi@mail.sbu.ac.ir}{mat.jamshidi},
\href{mailto:am.gholizadeh@mail.sbu.ac.ir}{am.gholizadeh}\}@mail.sbu.ac.ir
}
\end{center}

\thispagestyle{empty}

\begin{abstract}
\noindent This report presents the design and implementation of Auriga Robotics' 1:10 scale autonomous vehicle, developed at Shahid Beheshti University's Robotics and Intelligent Automation Laboratory. The platform integrates a custom mechanical chassis, a distributed embedded system based on ESP32 microcontrollers, and a Jetson Orin NX module for high-level perception and control. The embedded layer manages sensor interfacing, real-time data acquisition, and actuator control, while the Jetson handles environment perception through deep-learning--based lane, sign, and traffic light detection, as well as decision-making and trajectory control. System communication is achieved via structured UART links, ensuring synchronized operation between perception and control subsystems. The vehicle's software stack incorporates ROS, modular firmware design, and multiple control algorithms---Stanley, PID, and Pure Pursuit---to achieve stable autonomous navigation in structured driving scenarios. The overall system demonstrates robust performance and reliability in urban-like environments, serving as a modular research platform for future autonomous driving studies.

\vspace{0.5em}
\end{abstract}

\newpage
\tableofcontents
\newpage

\section{Introduction}

\begin{figure}[H]
    \centering
    \includegraphics[width=0.95\linewidth]{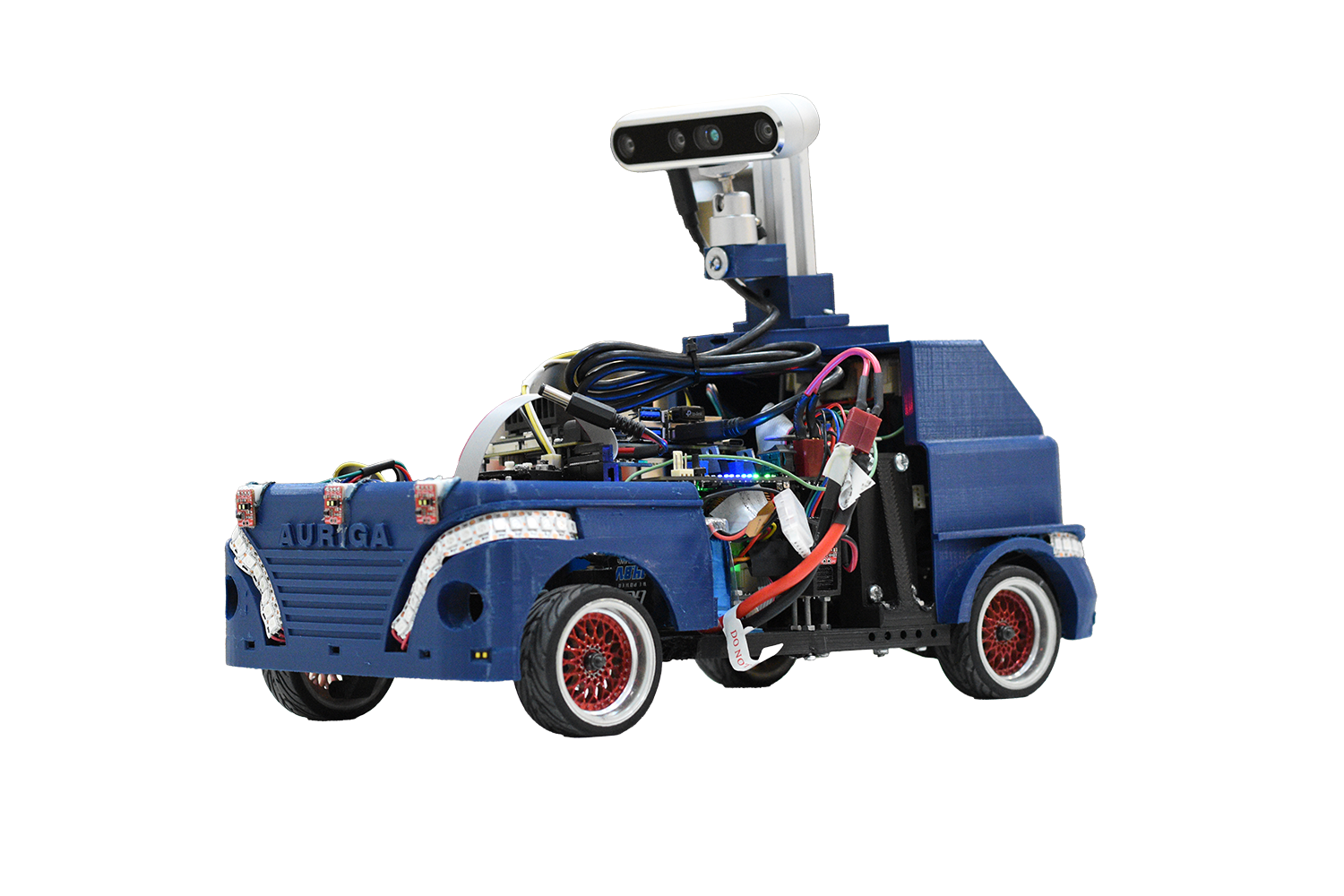}
    \caption{Auriga's 1:10 scale autonomous vehicle.}
    \label{fig:robotpic}
\end{figure}

Autonomous driving represents one of the most advanced integrations of control theory and computer vision, and is of high importance in self-driving cars in the current age. To explore these concepts in a manageable and scalable way, Auriga Robotics at Shahid Beheshti University developed a 1:10 scale autonomous vehicle designed to perform perception, decision-making, and control in real time. The project was carried out as part of the team's ongoing research in intelligent robotics and autonomous navigation systems, with the goal of building a modular platform suitable for academic experimentation. An image of the robot is attached in Fig.~\ref{fig:robotpic}.

The primary objective of the project was to design a complete autonomous driving stack, that includes hardware, embedded firmware, and high-level perception algorithms. The vehicle was required to detect and follow lanes, recognize traffic signs and lights, and maintain stable longitudinal and lateral control under various driving scenarios. To achieve this, the system was divided into three processing layers: low-level sensor integration and actuation was handled by two ESP32 microcontrollers (called CONTROL and ZERO), and high-level perception and decision-making executed on an NVIDIA Jetson Orin NX mini-computer, in which all three units maintained stable communication with each other. On the software side, implemented on the Jetson Orin NX, the architecture was built around ROS (Robot Operating System)~\cite{quigley2009ros} to allow modular algorithm development, parallel debugging, and integration of computer vision and control pipelines. 

The remainder of this report is organized as follows. Section~\ref{sec:hardware} presents the hardware design, including the mechanical structure, sensors, and electronics. Section~\ref{sec:embedded} details the embedded and high-level software architecture, covering sensor integration, communication between the processing units, and the Robot Operating System. Section~\ref{sec:cv} discusses the perception and computer-vision algorithms, together with the training strategy and experimental validation of the lane-detection model. Section~\ref{sec:control} covers vehicle localization, navigation, and the control algorithms used for lane-keeping, decision-making, parking, and intersection handling.

This report describes the 2025 platform of Team Auriga, which builds and improves on our earlier 2024 vehicle~\cite{auriga2024}.

\section{Hardware Description}
\label{sec:hardware}

\subsection{Mechanical Design}

The mechanical structure of the autonomous vehicle was fully custom-designed to balance robustness, modularity, and scalability. It consists of three main parts: the chassis, the steering and drivetrain system, and the vehicle body. Each component was designed and fabricated to ensure reliable operation, efficient integration with electronic systems, and easy maintenance.

\paragraph{Car Chassis.}
The chassis of the vehicle is designed at a 1:10 scale relative to a full-sized car. It consists of an aluminum base plate and a 3D-printed plastic upper layer. The chassis layout was designed to accommodate the DC motor, servo motor, rotary encoder, electronic control boards, and other system components.  
The primary motivation behind developing a custom chassis, rather than using a prebuilt model, was to integrate specific mechanical features required for improved performance. These include the addition of a more powerful DC motor for greater torque and speed, an increased ground clearance to allow traversal over small obstacles, and custom mounting points for sensors and electronics. The vehicle dimensions are mentioned in Table~\ref{tab:vehicle_dims}.

\begin{table}[H]
    \centering
    \caption{Vehicle dimensions.}
    \label{tab:vehicle_dims}
    \renewcommand{\arraystretch}{1.15}
    \begin{tabular}{@{}lc@{}}
    \toprule
    \textbf{Dimension} & \textbf{Value (cm)} \\ \midrule
    Wheelbase $L$ (front axle to rear axle)     & $25.0 \pm 1.0$ \\
    Vehicle width                               & $21.5 \pm 0.5$ \\
    Overall length (front bumper to rear bumper) & $39.5 \pm 0.5$ \\
    Height                                      & $29.5 \pm 0.5$ \\
    \bottomrule
    \end{tabular}
    \end{table}

\paragraph{Steering and Drivetrain System.}
The vehicle employs a single differential rear-wheel drive system, where propulsion is provided by a DC motor mounted on the rear axle. The steering mechanism follows an Ackermann configuration~\cite{rajamani2012} and is actuated by a servo motor. This arrangement enables the vehicle to achieve appropriate turning angles and efficiently cover a wide range of trajectories.

\paragraph{Car Body.}
The robot's body consists of two main parts: the front bumper, which houses the headlights and front distance sensors, and the rear section, which includes the taillights and the camera mount along with the rear distance sensors. Both parts were designed and fabricated using 3D printing to ensure precision fitting and lightweight structure.

\subsection{Electrical Design}

The electrical system of Auriga's autonomous vehicle is centered around a main control board and three auxiliary side boards. Two of these side boards are positioned at the front and roof of the vehicle, while the third board is dedicated to servo motor and DC motor data acquisition. This separation minimizes high-current noise coupling from the motor driver into sensitive digital communication lines.  
The side boards facilitate organized routing of sensor wires, ensuring efficient connections and reduced signal interference. Power is supplied by a 5000 mAh 4-cell lithium-polymer (Li-Po) battery, which delivers energy to both the drive and electronic subsystems, maintaining stable operation during extended runs.  
Voltage regulation is achieved through a combination of switching and linear regulators that maintain all subsystem voltages within safe operational limits. Over-voltage and over-current protection circuits are included to enhance reliability and prevent component damage. Several indicator LEDs are placed around the board to provide visual confirmation of correct voltage levels and regulator status, simplifying debugging and system validation.

\begin{figure}[H]
    \centering
    \begin{subfigure}[t]{0.8\linewidth}
        \centering
        \includegraphics[width=\linewidth]{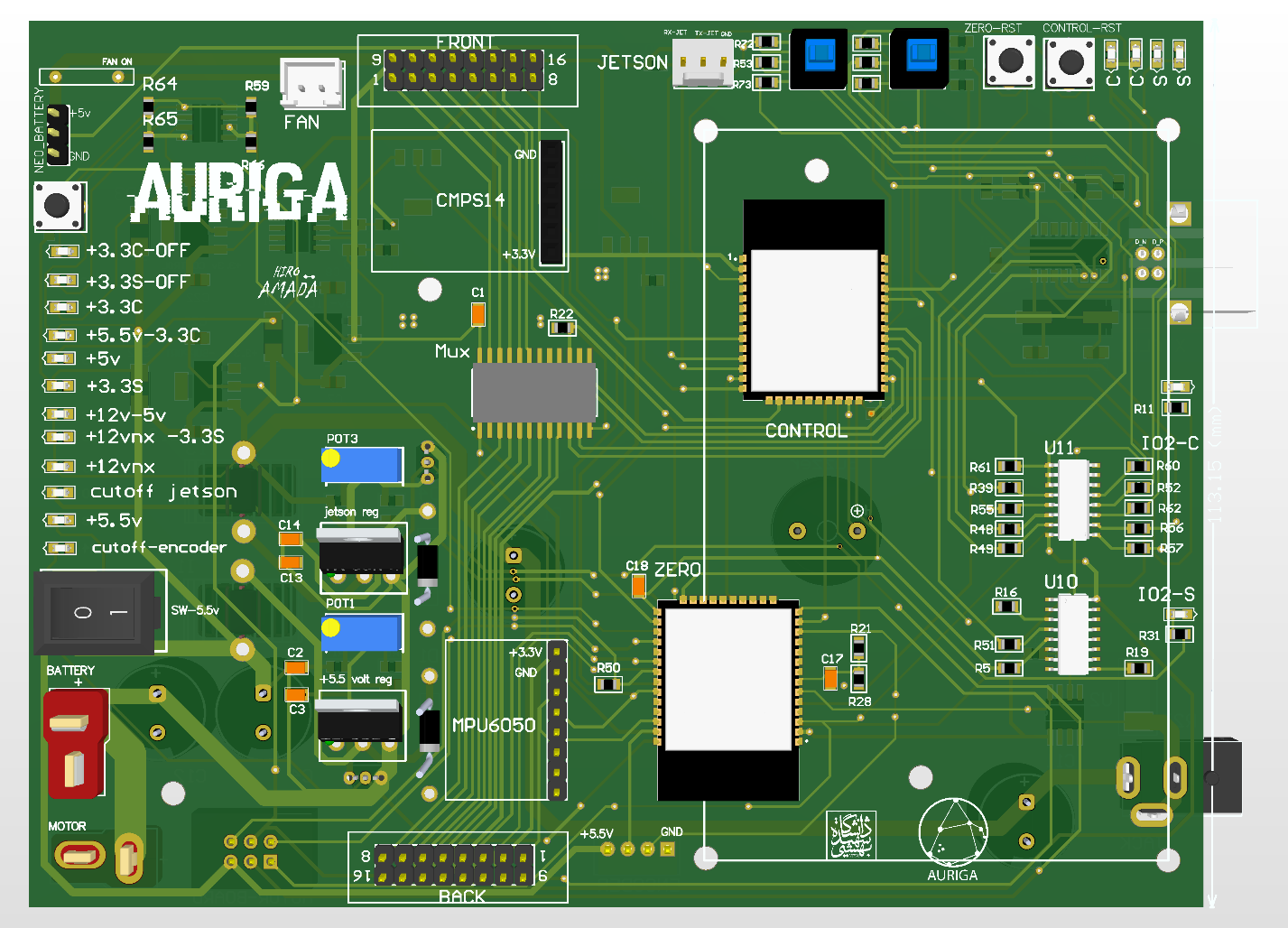}
        \caption{Top layer of the control board.}
        \label{fig:controlboard_top}
    \end{subfigure}
    \vspace{1em} 
    \begin{subfigure}[t]{0.8\linewidth}
        \centering
        \includegraphics[width=\linewidth]{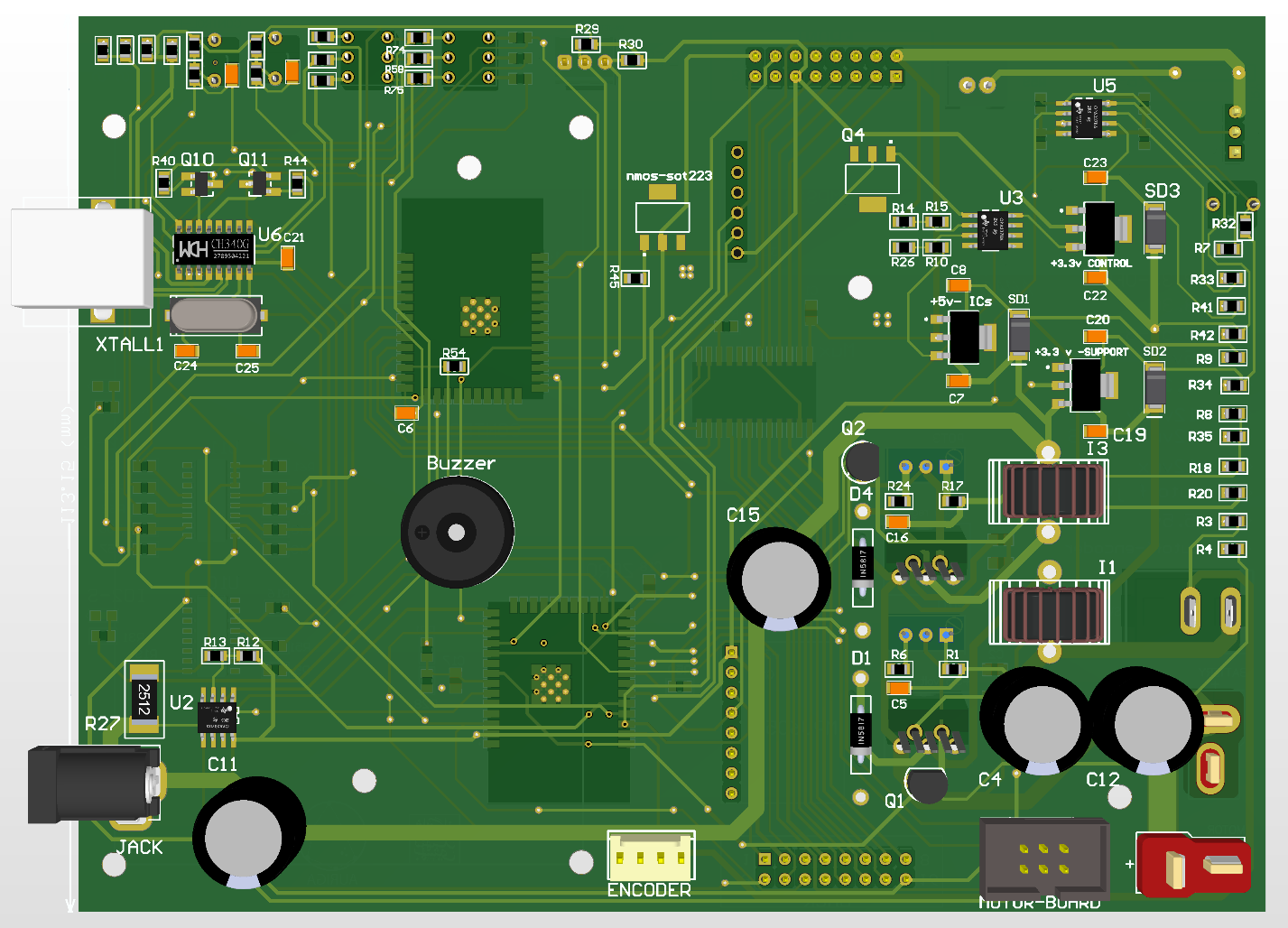}
        \caption{Bottom layer of the control board.}
        \label{fig:controlboard_bottom}
    \end{subfigure}
    \caption{Control board PCB layout showing (a) the top layer and (b) the bottom layer.}
    \label{fig:controlboard}
\end{figure}

\paragraph{Control Board.}
As shown in Fig.~\ref{fig:controlboard}, the main control board integrates two ESP32 microcontrollers~\cite{esp32}---\textit{ZERO} and \textit{CONTROL}---that communicate through a UART link. ZERO MCU handles sensor data acquisition and low-level control tasks, while CONTROL MCU manages high-level communication with the onboard computer (NVIDIA Jetson Orin NX~\cite{jetson}) and actuation, enabling synchronized data exchange and command handling.

Power for the control board and peripherals is provided by both switching and linear regulators: two XL4016 switching regulators generate 12~V and 5~V outputs for the Jetson Orin NX, servo motor, encoder, and sensors, while two AMS1117-3.3~V and one AMS1117-5~V linear regulators provide stable voltage for logic-level components.  

\begin{figure}[H]
    \centering
    \includegraphics[width=0.8\linewidth]{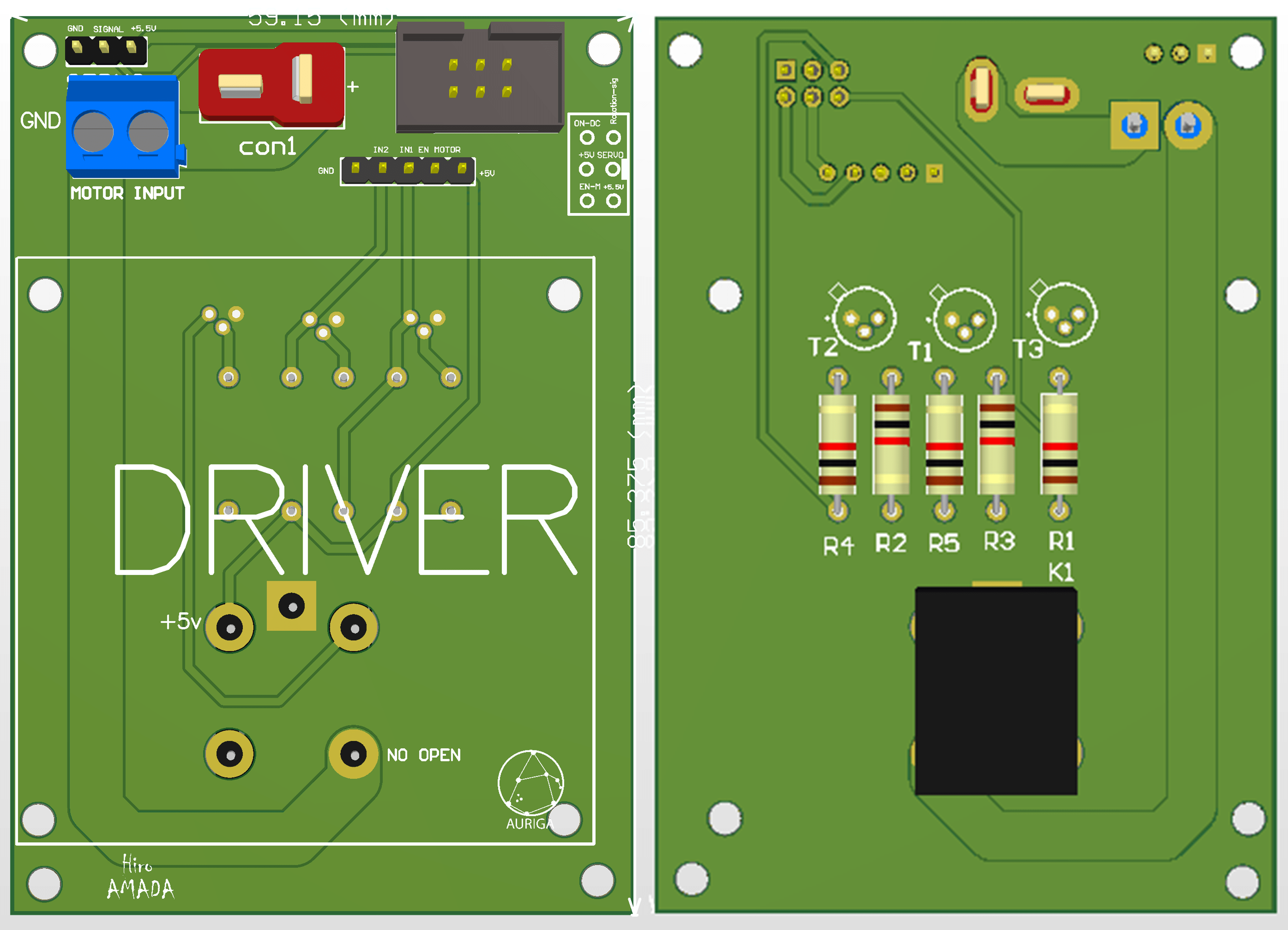}
    \caption{DC motor driver board PCB layout showing the top (left) and bottom (right) layers.}
    \label{fig:driverboard}
\end{figure}

\paragraph{DC Motor Driver Board.}
For DC motor actuation, an L298 dual H-bridge driver~\cite{l298} is employed, delivering sufficient current capacity for precise speed control. The printed circuit board (PCB) layout and assembly were designed and fabricated by the team, resulting in a compact and reliable architecture that integrates all power and control subsystems efficiently. The PCB design for the driver board is shown in Fig.~\ref{fig:driverboard}.

\paragraph{Sensors.}
Range measurement is handled by VL53L1X time-of-flight (ToF) laser-ranging sensors~\cite{vl53l1x} mounted on the front, rear, and sides of the vehicle, with a total of 12 ToF sensors. To avoid I\textsuperscript{2}C~\cite{i2c} interference across these sensors, they are all multiplexed and connected to the ZERO MCU. These sensors provide high-speed and accurate distance measurements for obstacle detection, complementing camera-based perception for enhanced situational awareness.

Orientation is measured by a single compass module---the CMPS14~\cite{cmps14}---which provides a tilt-compensated heading for real-time localization. An MPU-9250~\cite{mpu9250} was developed and tested separately on a breadboard and was not used on the robot. The CMPS14 is read by the ZERO MCU via the I\textsuperscript{2}C protocol.

The DC motor includes an optical rotary encoder that provides wheel speed and position feedback, enabling precise closed-loop control of longitudinal motion.

To enhance the visual feedback of the autonomous vehicle and provide a more intuitive representation of its actions, a string of WS2812B Neopixel LEDs is integrated into the system. These LEDs are used to indicate various behaviors of the vehicle, such as turning left or right, scanning the surrounding environment, and detecting obstacles.

\section{Embedded Design, High-Level Software Architecture, and Robot Operating System (ROS)}
\label{sec:embedded}

To achieve the required computational performance and maintain synchronized control across all subsystems, the vehicle's architecture incorporates three distinct processing units: two ESP32-based embedded controllers, named \textit{ZERO} and \textit{CONTROL}, and an NVIDIA Jetson Orin NX module. Each unit is assigned specific responsibilities within the perception--control hierarchy to ensure modularity, determinism, and real-time responsiveness. The Jetson Orin NX serves as the high-level computing platform, running perception algorithms, decision-making routines, and overall control logic. The CONTROL ESP32 functions as the intermediary layer, managing actuator commands, multiplexing, and structured communication with both the Jetson and the ZERO module. The ZERO ESP32 handles all low-level sensor acquisition tasks, including ToF, compass, and encoder data collection, before transmitting the processed data to the CONTROL unit for integration and forwarding to the Jetson.

\begin{table}[H]
\centering
\caption{Overview of the three processing units and their primary responsibilities.}
\begin{tabular}{@{}lll@{}}
\toprule
\textbf{Unit} & \textbf{Processor} & \textbf{Primary Function} \\ \midrule
Jetson Orin NX & ARM Cortex-A78AE + GPU & Perception and control logic \\
CONTROL ESP32 & Xtensa LX6 Dual-Core & Intermediary communication layer, actuator interfacing \\
ZERO ESP32 & Xtensa LX6 Dual-Core & Sensor acquisition, low-level actuation \\ 
\bottomrule
\end{tabular}
\end{table}

\subsection{Embedded System Software Architecture}

The embedded subsystem forms the low-level operational layer of Auriga's autonomous vehicle. It is responsible for acquiring sensor data, executing actuator commands, and maintaining synchronized communication between all processing units. The subsystem consists of two ESP32 microcontrollers, referred to as \textit{CONTROL} and \textit{ZERO}, that cooperate with the Jetson Orin NX computer. The CONTROL unit interfaces directly with the Jetson, relaying processed sensor data from ZERO and applying actuator commands such as steering and motor control. The ZERO unit manages all time-critical sensor measurements and DC motor control, providing a continuous data stream to CONTROL over UART. Together, these two microcontrollers form a distributed real-time embedded network bridging the vehicle's physical layer and the Jetson's high-level control layer.

\subsubsection{Firmware Architecture}

\begin{figure}[H]
    \centering
    \includegraphics[width=1\linewidth]{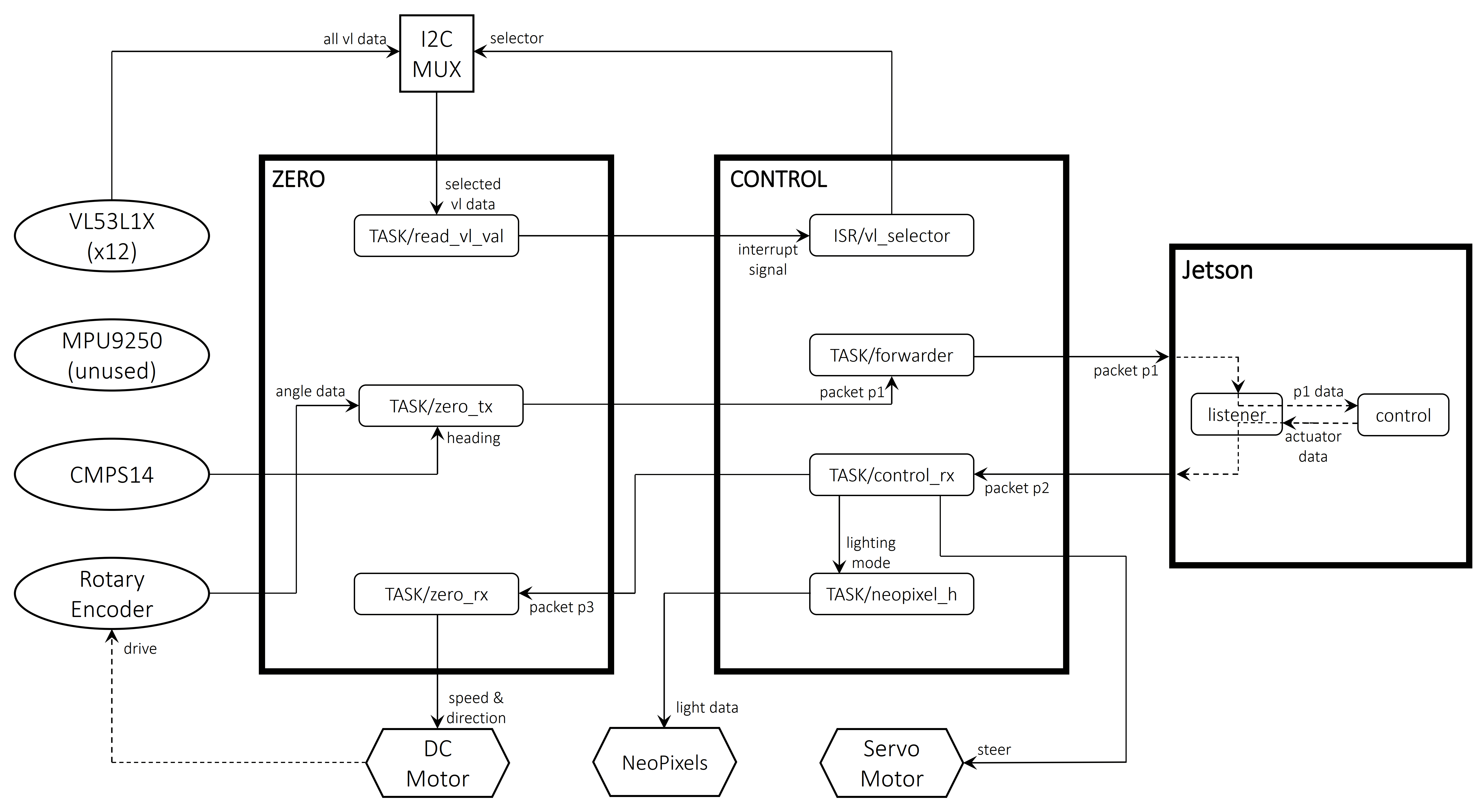}
    \caption{Embedded firmware steady-state architecture.}
    \label{fig:embedded_architecture}
\end{figure}

Both ESP32 units run a FreeRTOS-based firmware~\cite{barry2016freertos} divided into concurrent tasks, each handling an independent subsystem. On the ZERO module, tasks include ToF sensor polling, compass and encoder sampling, and UART data transmission. CONTROL executes complementary tasks for UART reception, servo actuation, lighting and signaling control (NeoPixels and buzzer), and I\textsuperscript{2}C multiplexer channel selection. Inter-task communication relies on FreeRTOS queues to ensure deterministic timing and non-blocking operation. Both cores of each ESP32 are utilized, dedicating time-critical acquisition and actuation tasks to one core while communication and housekeeping tasks run on the other. Fig.~\ref{fig:embedded_architecture} illustrates the steady-state organization of dataflow and task scheduling among CONTROL, ZERO, and Jetson. Task core numbers have been labeled in the form 
$\texttt{TASK-\{core\#\}/task\_name}$, and every arrow from one node to the other represents the data being transmitted, and the direction of transmission. The Robot Operating System architecture for perception and control has been abstracted away as \texttt{perception pipeline} and \texttt{control} in the figure, and will be covered in depth in section~\ref{subsec:ros}.

\begin{table}[H]
    \centering
    \caption{FreeRTOS tasks on the two ESP32 units (priority: higher runs first).}
    \label{tab:tasks}
    \small
    \renewcommand{\arraystretch}{1.1}
    \begin{tabular}{@{}llccp{5.6cm}@{}}
    \toprule
    \textbf{MCU} & \textbf{Task} & \textbf{Core} & \textbf{Prio.} & \textbf{Function / period} \\ \midrule
    ZERO & Sensor read and transmit & 0 & 10 & Reads all sensor data, builds and sends packet \texttt{p1}. \\
    ZERO & Instruction receiver & 0 & 10 & Parses \texttt{p3} (motor command, ToF bitmask); 1-tick polling \\
    ZERO & ToF reader & 1 & 13 & Sweeps the 12 ToF sensors every $\sim$41\,ms; pushes results to a queue \\
    ZERO & Button monitor & 1 & 15 & Debounces the motor cut-off button (ISR-flagged) (this task was not included in Fig.~\ref{fig:embedded_architecture} but is implemented). \\ \midrule
    CONTROL & ZERO$\to$Jetson forwarder & 0 & 10 & Receives \texttt{p1} from ZERO, checks its integrity and forwards it to Jetson \\
    CONTROL & NeoPixel handler & 1 & 10 & Applies the light mode; 5-tick polling, 500\,ms blink toggle \\
    CONTROL & Jetson receiver & 1 & 14 & Validates \texttt{p2}, drives the servo and queues NeoPixel data to NeoPixel handler task, relays \texttt{p3} to ZERO \\
    \bottomrule
    \end{tabular}
    \end{table}
    
    Three queues carry data between tasks (ToF values, ToF bitmask, light mode), and a mutex serialises access to the shared I\textsuperscript{2}C bus between the ToF reader and the compass read. Time-critical events are handled in interrupt service routines placed in IRAM: encoder edges (ZERO), the Jetson over-current signal (ZERO), the motor cut-off button (ZERO), and the ToF multiplexer step (CONTROL).

\subsubsection{Sensor and Actuator Integration}
\label{subsubsec:sensor_integration}

As shown in Fig.~\ref{fig:embedded_architecture}, during steady-state operation the embedded layer continuously acquires and distributes data from fifteen data channels: twelve VL53L1X time-of-flight (ToF) sensors arranged around the vehicle perimeter, one rotary encoder for wheel speed feedback, a CMPS14 tilt-compensated magnetic compass, and the time elapsed since Jetson handshake was made.

To prevent I\textsuperscript{2}C address conflicts, the VL53L1X modules are multiplexed through a hardware selector controlled by the CONTROL ESP32. The ZERO ESP32 requests channel selection and sequentially samples the active sensor, packaging the range data for transmission. The CMPS14 provides a direct heading measurement, while the encoder delivers rotational velocity used for odometry and closed-loop motor control. 

All twelve VL53L1X sensors share the default I\textsuperscript{2}C address, so only one is connected to the bus at a time. After each reading, ZERO sends a 100\,\textmu s pulse on a dedicated line; an interrupt on CONTROL advances a channel counter (modulo 12) and drives the four multiplexer select lines. The sensors run in long-distance mode with a 33\,ms timing budget and a 40\,ms inter-measurement period, giving a full-sweep of about half a second. Because packets are sent faster than sweeps complete, the last ToF values are repeated until a new sweep arrives. The Jetson can disable individual sensors through a 12-bit bitmask in \texttt{p2}; disabled or failed sensors are reported as $-1$. 

The rotary encoder is decoded in x4 mode: an interrupt on every edge of both channels reads both pin levels and updates a signed position counter, giving 1600 counts per revolution. Speed is obtained from the count difference between consecutive packets and the measured time step.

The steering servo is driven at 50\,Hz with a pulse width of 1000--1815\,\textmu s, mapped linearly from the 8-bit command $u\in[0,255]$ (3.2\,\textmu s per step, using a 16-bit PWM timer). The DC motor driver receives an enable line, a direction line, and a 500\,Hz, 10-bit PWM signal; the sign of the motor command sets the direction.

All sensor outputs are collected and structured into a unified serial data packet, transmitted from ZERO to CONTROL via UART. CONTROL verifies packet integrity, forwards the data to the Jetson Orin NX, and awaits actuator commands. These commands are returned to CONTROL, containing servo steering angles, motor drive values, and LED states. CONTROL applies its own actuator commands locally and relays motor control commands to ZERO, which drives the DC motor accordingly. The details of packet structure and synchronization protocol are discussed in section~\ref{subsubsec:commprotocol}.

\subsubsection{Sensor Fusion for Heading Estimation}
\textbf{Note: the MPU-9250 was not used on the 2025 vehicle. Due to integration deadlines, the robot used only the CMPS14 heading during live track testing. The sensor fusion described below was developed and validated offline on a breadboard after the laboratory closures, is independent of the robot platform, and is the intended architecture for future platforms.}

\label{subsubsec:mpu}
Accurate heading estimation is required for stable vehicle control and localization. The test platform was designed to use two complementary modules: an MPU-9250 IMU (3-axis accelerometer, 3-axis gyroscope, 3-axis magnetometer) and a CMPS14 tilt-compensated compass (BNO080-based) which provides a fused, tilt-compensated heading output. The relevant module specifications and internal fusion capability are documented in the manufacturers' datasheets.

A two-stage fusion pipeline was implemented: (1) internal Kalman filtering~\cite{kalman1960,welch1995} on MPU-9250 measurements to produce a drift-compensated yaw estimate \(\psi_{\mathrm{MPU}}\) and (2) a rate-dependent blend between \(\psi_{\mathrm{MPU}}\) and the CMPS14 heading \(\psi_{\mathrm{CMPS}}\).

\paragraph{Stage 1 --- Heading estimation using the MPU-9250 data.}
The MPU-9250 sensor only provides raw vector readings for the relative acceleration $\vect{a}$, magnetic field strength $\vect{m}$, and the angular velocity $\vect{\omega}$. Here, we combine the three to get an estimate on $\psi$, the heading angle.

\begin{figure}[H]
    \centering
    \includegraphics[width=0.8\linewidth]{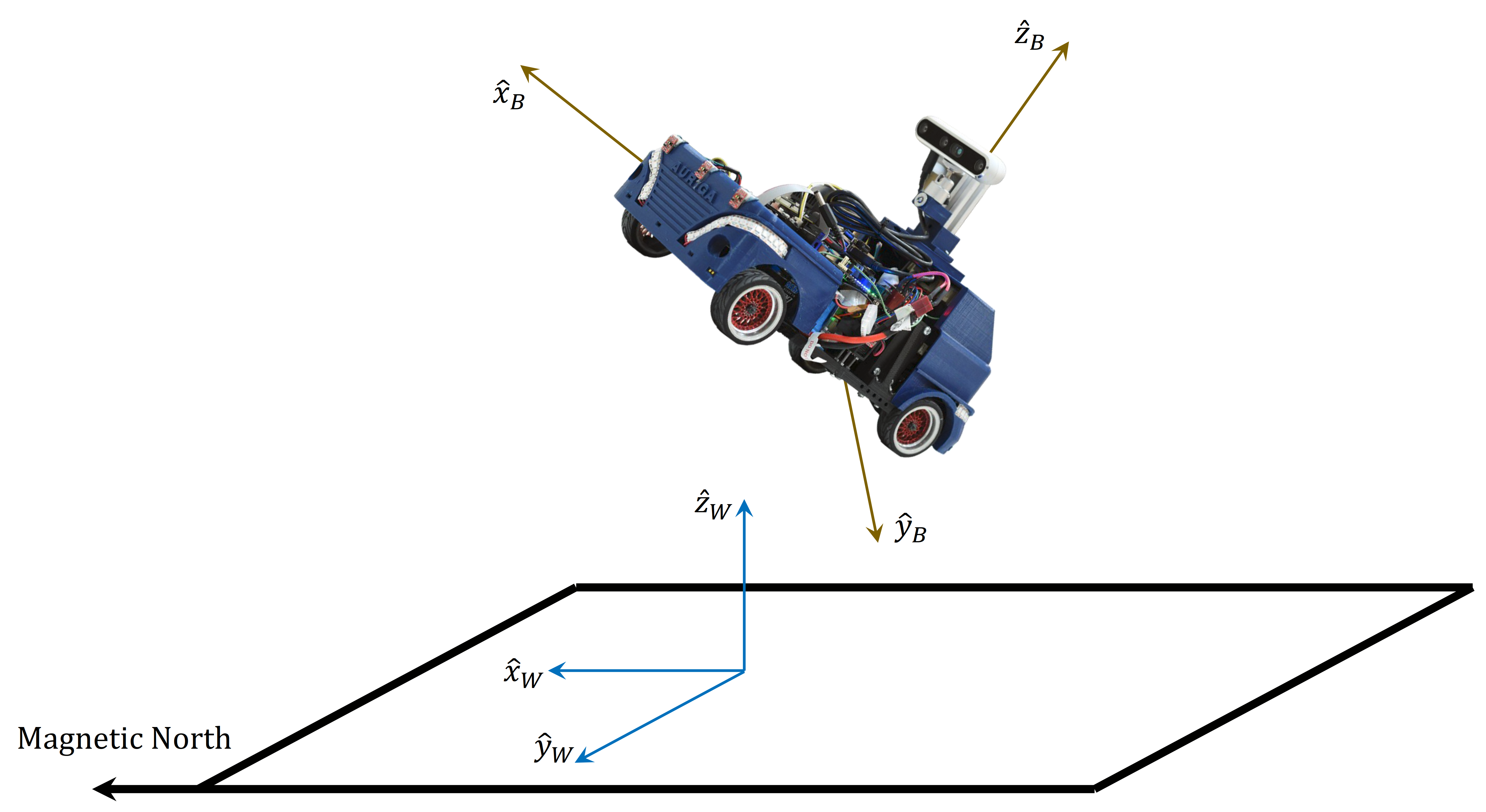}
    \caption{World $\mathcal{B}_W$ and body $\mathcal{B}_B$ coordinate systems.}
    \label{fig:bodyworld}
\end{figure}

\textbf{Definitions.} Define the \textit{world} orthonormal vector basis $\mathcal{B}_W=\{\hatt{x}_W,\hatt{y}_W,\hatt{z}_W\}$, where $\hatt{x}_W$ points towards the \textit{north magnetic pole}, $\hatt{y}_W$ points to the \textit{magnetic west}, and $\hatt{z}_W$ points upwards from the horizontal plane.
This basis is universal and fixed. Parallel to the fixed basis, define the time-varying \textit{body} orthonormal basis $\mathcal{B}_B=\{\hatt{x}_B,\hatt{y}_B,\hatt{z}_B\}$, where $\hatt{x}_B$ points to the front of the vehicle, $\hatt{y}_B$ points to the left of the vehicle, and $\hatt{z}_B$ points upwards with respect to the vehicle.
The basis system is shown in Fig.~\ref{fig:bodyworld}.

For any pure vector $\vect{v}$, let the representation in the world and body coordinates be $\vect{v}^W$ and $\vect{v}^B$ respectively. These two representations are related to each other with a 
rotation matrix $\bm{R}_{WB}$, such that 
\begin{equation}
    \vect{v}^W=\bm{R}_{WB} \vect{v}^B, \quad \vect{v}^B=\bm{R}_{WB}^T \vect{v}^W.
    \label{eq:bodyworld_coordinate_transformation}
\end{equation}
Where $\bm{R}_{WB}$ encodes the full orientation of the robot. Here, since the robot only moves in the horizontal plane, we only care about the heading angle and a full orientation state of the robot is not computed. 
The heading angle $\psi$ is defined as the angle between the horizontal projection (projection onto the plane spanned by $\{\hatt{x}_W,\hatt{y}_W\}$) of
$\hatt{x}_B$ and the magnetic north axis $\hatt{x}_W$, measured clockwise
about $\hatt{z}_W$:
\begin{equation}
    \psi = \operatorname{atan2}\!\left(-\hatt{x}_B\cdot\hatt{y}_W,\; \hatt{x}_B\cdot\hatt{x}_W\right)
     = \operatorname{atan2}\!\left(-[\bm{R}_{WB}]_{21},\; [\bm{R}_{WB}]_{11}\right),
\quad \psi\in[0,2\pi),
    \label{eq:heading_def}
\end{equation}
Additionally, we define the operators $\bm P_{\mathcal H}$ for projecting any vector to the horizontal plane $\mathcal H=\operatorname{span}(\hatt{x}_W,\hatt{y}_W)$ in both coordinate systems:
\begin{equation}
    \bm P_{\mathcal H}^B=\bm{1}-\hatt{z}_W^B\left(\hatt{z}_W^B\right)^T, \quad \bm P_{\mathcal H}^W=\bm{1}-\hatt{z}_W^W\left(\hatt{z}_W^W\right)^T.
    \label{eq:projection_ops}
\end{equation} 
Where $\bm 1$ is the identity matrix. Therefore for any vector $\vect{v}$, let $\vect{v}_{\mathcal H}=\bm P_{\mathcal H}\vect{v}$ be the vector projected onto the horizontal plane. 

\textbf{Accelerometer.} The accelerometer on the MPU-9250 measures the relative acceleration vector $\vect{a}_m=\vect{a}_{\mathrm{real}}-\vect{g}$ in the body coordinate (the subscript $m$ denotes the measured result: the final value the sensor gives), where $\vect{g}$ is the gravitational acceleration. Therefore, the ideal readings are of the form
\begin{equation}
    \vect{a}_m^B=\bm{R}_{WB}^T \vect{a}_{\mathrm{real}}^W-\vect{g}^B.
    \label{eq:accelerometer_model}
\end{equation}
For the stationary robot, from Eq.~\ref{eq:accelerometer_model} we get $\vect{a}_m^B\approx -\vect{g}^B$. The real acceleration vector $\vect{a}_{\mathrm{real}}(t)$ is fast-changing in time,
hence for the discretized model of the accelerometer $\vect{a}^B_m[n]$, a low-pass filter is implemented to arrive at an estimate of $\vect{g}^B$ in the form 
\begin{equation}
    -\vect{g}^B[n]=-(1-\alpha)\vect{g}^B[n-1]+\alpha\vect{a}^B_m[n], \quad \alpha\in[0,1].
    \label{eq:gravity_lpf}
\end{equation}
The value $\alpha=0.05$ was chosen for this platform. From Eq.~\ref{eq:gravity_lpf}, we get an estimate of the universal up direction $\hatt{z}_W$, expressed in the body coordinates (Eq.~\ref{eq:universal_up}).
\begin{equation}
    \hatt{z}_W^B=\frac{-\vect{g}^B}{\lVert\vect{g}^B\rVert}
    \label{eq:universal_up}
\end{equation}
From this, we can find the horizontal plane in the body coordinates. Define the forward direction unit vector $\hatt{f}$, which is the projection of the body front unit vector $\hatt{x}_B$ onto the horizontal plane, 
calculated as 
\begin{equation}
    \vect{f}^B=\hatt{x}_B^B-(\hatt{x}_B^B\cdot \hatt{z}_W^B)\hatt{z}_W^B=\begin{bmatrix}
    1-(z_{W,x}^B)^2 \\
    -z_{W,x}^B z_{W,y}^B \\
    -z_{W,x}^Bz_{W,z}^B
    \end{bmatrix}_{\text{Body}}, \quad \hatt{f}^B=\frac{\vect{f}^B}{\lVert\vect{f}^B\rVert},
    \label{eq:front_vec}
\end{equation}
and the left direction unit vector in the horizontal plane is then simply obtained by
\begin{equation}
    \hatt{l}^B=\hatt{z}_W^B\times\hatt{f}^B=\begin{bmatrix}
        z_{W,y}^Bf_z^B-z_{W,z}^Bf_y^B \\
        z_{W,z}^Bf_x^B-z_{W,x}^Bf_z^B \\
        z_{W,x}^Bf_y^B-z_{W,y}^Bf_x^B
    \end{bmatrix}_{\text{Body}}.
\end{equation}
\textbf{Gyroscope.} The gyroscope measures the angular velocity vector $\vect{\omega}^B $ in the body coordinates. The heading rate change is 
\begin{equation}
    -\frac{d\psi}{dt}=\vect{\omega}^W\cdot\hatt{z}_W^W=\left(\vect{\omega}^B\right)^T\bm{R}_{WB}^T\bm{R}_{WB} \hatt{z}_W^B=\vect{\omega}^B\cdot\hatt{z}_W^B.
    \label{eq:gyro_headingrate}
\end{equation}
Projecting onto the estimated vertical rather than the board's up axis ($\hatt{z}_B$) keeps the yaw rate correct when the vehicle pitches or rolls.

The chip gives a 16-bit integer per direction. The conversion factor, specified in the datasheet, is $S = 65.5\ \mathrm{LSB/(^\circ/s)}$.
Discretizing Eq.~\ref{eq:gyro_headingrate}, the gyroscope output is the noisy, biased measurement
\begin{equation}
    \vect{\omega}_m^B[n] = \vect{\omega}^B[n]+\vect{b}^B+\vect{\eta}_g^B[n],
\end{equation}
with slowly varying bias $\vect{b}$ and white noise $\vect{\eta}_g$. A startup average $\vect{b}^B[0]$ of 200 stationary samples is subtracted, giving $\vect{\omega}_m^B[n]-\vect{b}^B[0]$ as the measured angular velocity. The measured heading rate is therefore
\begin{equation}
    \dot{\psi}_{\mathrm{gyro}}[n] = -\left(\vect{\omega}^B[n]-\vect{b}^B[0]\right)\cdot\hatt{z}_W^B[n],
    \label{eq:rate_meas}
\end{equation}
for small $n$ near the robot startup (the update rule for $\vect{b}$ is discussed in Stage 2). Here, $\hatt{z}_W^B[n]$ comes from Eq.~\ref{eq:universal_up}.  
Integrating with the measured step $\Delta t[n]$ (about 10~ms, calculated in each iteration) gives
\begin{equation}
    \psi[n] = \operatorname{wrap}_{[0,2\pi)} \big(\psi[n-1] + \dot{\psi}_{\mathrm{gyro}}[n]\,\Delta t[n]\big),
    \label{eq:euler_integration}
\end{equation}
Where $\operatorname{wrap}_{I}(\cdot)$ maps any angle into the principal interval $I$. Hence, the gyroscope gives a final noisy and biased estimate $\psi_{\mathrm{gyro}}$.

\textbf{Magnetometer.} The magnetometer sensor is useful for measuring Earth's magnetic field, which provides a universally fixed coordinate system. The magnetometer measurement model is 
\begin{equation}
    \vect{m}_m^B=\bm \Gamma\vect{M}_{\oplus}^B-\vect{O}^B,
    \label{eq:mag_model}
\end{equation}
where $\vect{M}_{\oplus}$ is the Earth's magnetic field, $\vect{O}$ is the magnetic offset vector, and $\bm \Gamma$ is the magnetic field distortion matrix. 
The methods required to extract $\vect{M}_{\oplus}$ from $\vect{m}_m$ are discussed in Stage 4. For the ideal magnetometer with 
$\vect{m}_m=\vect{M}_{\oplus}$, we can get a direct estimate on the heading angle.

Decomposing Earth's magnetic field as $\vect{M}^W_{\oplus}=\vect{M}^W_{\oplus,\mathcal H}+M^W_z \hatt{z}_W^W$ (Eq.~\ref{eq:projection_ops}), we have 
\begin{equation}
    \vect{M}_{\oplus,\mathcal H}^W=\bm P_{\mathcal H}^W\vect{M}_{\oplus}=\bm P_{\mathcal H}^W\bm R_{WB}\vect{m}^B_m,
\end{equation}
and since $\bm P_{\mathcal H}^W\bm{R}_{WB}=\bm{R}_{WB}\bm P_{\mathcal H}^B$, we get 
\begin{equation}
    \vect{M}^W_{\oplus,\mathcal H}=\bm R_{WB}\vect{m}^B_{m,\mathcal H}.
    \label{eq:tilt_compensation}
\end{equation}
This is the equation for tilt-compensation. Since, by definition, 
\begin{equation}
    \frac{\vect{M}^W_{\oplus,\mathcal H}}{\lVert \vect{M}^W_{\oplus,\mathcal H} \rVert}=\hatt{x}_W^W,
\end{equation}
and by definition for the heading angle, since 
\begin{equation}
    \hatt{f}^W=\cos\psi \hatt{x}_W^W-\sin\psi \hatt{y}_W^W,
\end{equation}
we conclude that the heading angle is given by 
\begin{equation}
    \psi=\operatorname{atan2}\big(\vect{m}^B_{m,\mathcal H}\cdot \hatt{l}^B, \, \vect{m}^B_{m,\mathcal H}\cdot\hatt{f}^B \big),
    \label{eq:mag_heading}
\end{equation}
where $\vect{m}^B_{m,\mathcal H}=\bm P_\mathcal{H}^B \vect{m}^B_m$. Hence, we get a direct estimate on the heading angle $\psi_{\mathrm{mag}}$.
\paragraph{Stage 2 --- Kalman filtering.}
Integrating the measured heading rate alone (Eq.~\ref{eq:euler_integration}) is dead reckoning: the white noise $\vect{\eta}_g^B$ accumulates as a random walk in $\psi$, and any bias left over after subtracting $\vect{b}^B[0]$ accumulates as a linear drift. The magnetometer heading $\psi_{\mathrm{mag}}$ (Eq.~\ref{eq:mag_heading}) has the opposite property: it does not drift, but it is noisy and is disturbed by steel and motor currents. Since the two errors are complementary, the two heading sources are fused with a Kalman filter~\cite{kalman1960,welch1995,simon2006}, which is closely related to a complementary filter, except that the gain is computed from the noise statistics instead of being fixed.

\textbf{State and model.} The state is
\begin{equation}
    \vect{x}[n]=\begin{bmatrix}\psi[n]\\ b_\psi[n]\end{bmatrix},
\end{equation}
where $b_\psi[n]$ is the \textit{residual} heading-rate bias that remains after the startup average $\vect{b}^B[0]$ has been removed in Eq.~\ref{eq:rate_meas}. The true heading rate is therefore $\dot{\psi}_{\mathrm{gyro}}[n]-b_\psi[n]$, where $\dot{\psi}_{\mathrm{gyro}}[n]$ is the measured heading rate of Eq.~\ref{eq:rate_meas}. The measured $\dot{\psi}_{\mathrm{gyro}}[n]$ acts as a known input to the model, and the residual bias is modeled as a random walk:
\begin{equation}
    \vect{x}[n]=\underbrace{\begin{bmatrix}1&-\Delta t[n]\\0&1\end{bmatrix}}_{\bm F[n]}\vect{x}[n-1]
    +\underbrace{\begin{bmatrix}\Delta t[n]\\0\end{bmatrix}}_{\bm G[n]}\dot{\psi}_{\mathrm{gyro}}[n]+\vect{w}[n],
    \qquad \vect{w}[n]\sim\mathcal N\big(\vect 0,\bm Q[n]\big).
    \label{eq:kf_process}
\end{equation}
These correspond to the continuous-time update rules of $\dot{\psi}(t)=\dot{\psi}_{\mathrm{gyro}}(t)-b(t)+w_{\psi}(t)$ and $\dot b(t)=w_{b_\psi}(t)$. The bias is only affected by noise in this model. The first row of Eq.~\ref{eq:kf_process} reads $\psi[n]=\psi[n-1]+\big(\dot{\psi}_{\mathrm{gyro}}[n]-b_\psi[n-1]\big)\Delta t[n]$, which is Eq.~\ref{eq:euler_integration} with the bias removed, and the second row says $b_\psi$ stays constant between steps. The process noise is $\bm Q[n]=\operatorname{diag}\big(q_\psi\Delta t[n],\ q_b\Delta t[n]\big)$, where $q_\psi$ accounts for the gyroscope noise and integration error in $\psi$, and $q_b$ is the rate at which the bias wanders. The magnetometer heading is the measurement:
\begin{equation}
    \psi_{\mathrm{mag}}[n]=\bm H\vect{x}[n]+v[n],\qquad \bm H=\begin{bmatrix}1&0\end{bmatrix},\quad v[n]\sim\mathcal N\big(0,R[n]\big).
    \label{eq:kf_meas}
\end{equation}
Here $\psi_{\mathrm{mag}}[n]$ is Eq.~\ref{eq:mag_heading} evaluated with the calibrated magnetic vector $\vect{m}_c^B$ of Stage 4 in place of $\vect{m}_m^B$. The accelerometer is neither an input nor a measurement of the filter. It only provides $\hatt{z}_W^B$ (Eq.~\ref{eq:universal_up}), which is used in $\dot{\psi}_{\mathrm{gyro}}[n]$ and in the tilt compensation of $\psi_{\mathrm{mag}}[n]$.

From here, for any quantity $u$ (vector or not) let $\hat{u}[n|n-1]$ be the quantity estimated at step $n$ using the data up to $n-1$ (the prediction section), and let $\hat{u}[n|n]$ be the updated prediction after acquiring the magnetometer heading estimate, $\psi_{\mathrm{mag}}$. Furthermore, 
if $u$ is the true reading, define $\tilde{u}=u-\hat{u}$ as the error in the quantity $u$.

\textbf{Prediction.} Let $\bm P=\mathbb{E}\left[\tilde{\vect{x}}\tilde{\vect{x}}^T\right]$ be the $2\times 2$ covariance matrix. The prediction runs once per gyroscope sample, whether or not a magnetometer sample is available:
\begin{align}
    \hat{\vect{x}}[n|n-1]&=\bm F[n]\,\hat{\vect{x}}[n-1|n-1]+\bm G[n]\,\dot{\psi}_{\mathrm{gyro}}[n],\label{eq:kf_pred_x}\\
    \bm P[n|n-1]&=\bm F[n]\,\bm P[n-1|n-1]\,\bm F[n]^T+\bm Q[n],\label{eq:kf_pred_P}
\end{align}
followed by $\hat\psi[n|n-1]\leftarrow\operatorname{wrap}_{[0,2\pi)}\big(\hat\psi[n|n-1]\big)$. Here, $\bm P[n|n-1]=\mathbb{E}\left[\tilde{\vect{x}}[n|n-1]\tilde{\vect{x}}^T[n|n-1]\right]$. Without magnetometer samples this reduces to dead reckoning, and $\bm P$ grows. This is fixed when 
calculating $\bm P[n|n]$ in the update section.

\textbf{Update.} When a valid magnetometer heading $\psi_{\mathrm{mag}}[n]$ is available, the innovation
\begin{equation}
    y[n]=\operatorname{wrap}_{(-\pi,\pi]}\!\big(\psi_{\mathrm{mag}}[n]-\hat\psi[n|n-1]\big)
    \label{eq:innovation}
\end{equation}
is calculated as the difference between the predicted value and the directly measured value. Then, the update algorithm updates the estimated values as below.
\begin{align}
    \sigma_y^2[n]&=\bm H\bm P[n|n-1]\bm H^T+R[n],\label{eq:kf_innov}\\
    \bm K[n]&=\bm P[n|n-1]\,\bm H^T\big/\sigma_y^2[n],\label{eq:kf_gain}\\
    \hat{\vect{x}}[n|n]&=\hat{\vect{x}}[n|n-1]+\bm K[n]\,y[n],\\
    \bm P[n|n]&=\big(\bm 1-\bm K[n]\bm H\big)\bm P[n|n-1].\label{eq:kf_update}
\end{align}
The innovation $y[n]$ uses the signed wrap so that $359^\circ\to1^\circ$ counts as $+2^\circ$ instead of $-358^\circ$. The gain $\bm K[n]$ splits the correction between the two states according to their uncertainties: a large $R[n]$ makes the filter trust the gyroscope, and a large $P_{00}$ makes it trust the magnetometer. Afterwards, $\hat\psi$ is wrapped to $[0,2\pi)$ and $\bm P$ is symmetrized, to avoid asymmetry in single-precision arithmetic.

$\bm H$ observes only $\psi$, but the pair $(\bm F,\bm H)$ is observable, since $\begin{bmatrix}\bm H\\ \bm H\bm F\end{bmatrix}=\begin{bmatrix}1&0\\1&-\Delta t[n]\end{bmatrix}$ has rank two. In practice, if the gyroscope has a residual bias, the innovation $y[n]$ keeps the same sign. The cross-covariance $P_{10}$ links this to $b_\psi$, so the second gain component $K_1=P_{10}/\sigma_y^2$ moves $\hat b_\psi$ until the innovation averages to zero. The bias is therefore observable only through the magnetometer.

\textbf{Adaptive measurement noise.} The Earth's field strength $B_0=\lVert\vect{M}_\oplus\rVert$ is constant at a fixed location, so a measured strength that deviates from $B_0$ indicates a disturbance (motors, nearby steel). Define
\begin{equation}
    \delta_B[n]=\left|\frac{\lVert\vect{m}_{c,\mathcal H}^B[n]\rVert}{B_0}-1\right|,\qquad
    R[n]=R_0\left[1+\left(\frac{\delta_B[n]}{\delta_0}\right)^2\right],
    \label{eq:kf_adaptiveR}
\end{equation}
where $\vect{m}_{c,\mathcal H}^B=\bm P_{\mathcal H}^B\vect{m}_c^B$ is the horizontal part of the calibrated field (in the 3D calibration mode of Stage 4, the full vector $\vect{m}_c^B$ is used instead). A disturbed field therefore makes the magnetometer less trusted, and if $\delta_B[n]>0.3$ the sample is discarded and only the prediction step is run.

\textbf{Initialization.} The first call of the update loop only stores a time reference, and $\Delta t[n]$ is measured from a hardware timer at every iteration and limited to 0.2\,s. Before the first accepted magnetometer sample, the heading is unknown. At that sample, the filter is initialized as $\hat\psi=\psi_{\mathrm{mag}}$, $P_{00}=R_0$, $P_{01}=P_{10}=0$, $P_{11}=P_{b,0}$ and $\hat b_\psi=0$. The same re-initialization occurs if the magnetometer calibration becomes valid for the first time or becomes invalid. If the startup average $\vect{b}^B[0]$ was rejected because the vehicle moved during the measurement (spread larger than $3^\circ$/s), it is set to zero, and $b_\psi$ then has to absorb the full gyroscope bias.

\begin{table}[H]
\centering
\caption{Kalman filter parameters.}
\label{tab:kf_params}
\renewcommand{\arraystretch}{1.1}
\begin{tabular}{@{}llc@{}}
\toprule
\textbf{Parameter} & \textbf{Meaning} & \textbf{Value} \\ \midrule
$q_\psi$ & heading process noise & $10^{-5}\ \mathrm{rad^2/s}$ \\
$q_b$ & residual-bias random-walk noise & $10^{-7}\ \mathrm{rad^2/s^3}$ \\
$R_0$ & magnetometer heading noise & $3\times10^{-3}\ \mathrm{rad^2}$ ($\approx 3^\circ$) \\
$\delta_0$ & field-deviation scale in Eq.~\ref{eq:kf_adaptiveR} & $0.1$ \\
$P_{00}$ before first update & initial heading variance & $3\ \mathrm{rad^2}$ \\
$P_{b,0}$ & initial residual-bias variance & $10^{-2}\ \mathrm{(rad/s)^2}$ \\
\bottomrule
\end{tabular}
\end{table}

Approximating the steady state with the continuous-time algebraic Riccati equation, with the measurement noise intensity $R_0\Delta t$ for $\Delta t=10$\,ms, gives a heading standard deviation of about $0.26^\circ$, a residual-bias standard deviation of about $0.06^\circ/\mathrm{s}$ and a heading gain $K_0\approx7\times10^{-3}$ per update. The heading therefore converges to the magnetometer with a time constant of about $\Delta t/K_0\approx1.5$\,s, while the gyroscope carries the heading between updates.

\paragraph{Stage 3 --- Rate-dependent CMPS14/MPU blend.}
The CMPS14 provides a direct tilt-compensated heading \(\psi_{\mathrm{CMPS}}\). We expected this sensor to drift considerably more at standstill than the Kalman-filtered MPU-9250; our tests did not confirm this (Section~\ref{subsubsec:heading_validation}). To exploit the fast reaction of the onboard fusion while retaining the low-noise, bias-corrected estimate from the MPU Kalman filter at low dynamics, we blend the two headings with a scalar weight \(w(|\omega_z|)\) that is a monotonic function of the yaw rate magnitude:
\begin{equation}
    \psi_{\mathrm{fused}} = w(|\omega_z|)\,\psi_{\mathrm{CMPS}} + [1-w(|\omega_z|)]\,\psi_{\mathrm{MPU}}.
\end{equation}
The weight is implemented as a smooth piecewise linear function in our baseline:
\begin{equation}
    w(|\omega_z|)=
    \begin{cases}
        0, & |\omega_z|\le \omega_1,\\[4pt]
        \dfrac{|\omega_z|-\omega_1}{\omega_2-\omega_1}, & \omega_1<|\omega_z|<\omega_2,\\[6pt]
        1, & |\omega_z|\ge \omega_2.
    \end{cases}
\end{equation}
Thresholds \(\omega_1,\omega_2\) are design parameters and must be tuned to the vehicle dynamics; the angular velocity thresholds were empirically selected for our platform. The two sensors do not share the same zero reference, so a constant offset measured with the board at rest is subtracted from $\psi_{\mathrm{CMPS}}$, and the difference is wrapped to $(-\pi,\pi]$ before blending. In our tests the blend did not measurably reduce stationary drift compared with using the CMPS14 alone (Section~\ref{subsubsec:heading_validation}).

\paragraph{Stage 4 --- Magnetometer calibration and startup.}
So far the magnetometer was assumed ideal. In reality the measured vector follows Eq.~\ref{eq:mag_model}, $\vect{m}_m^B=\bm\Gamma\vect{M}_\oplus^B-\vect{O}^B$, where $\bm\Gamma$ (soft iron: scaling, cross-axis coupling) and $\vect{O}$ (hard iron: constant offset) depend on the vehicle's wiring, batteries and mounting. They are therefore estimated on the vehicle itself, from the sensor readings only~\cite{vasconcelos2011,renaudin2010}.

\textbf{Raw data.} The AK8963 magnetometer is read at 100\,Hz in 16-bit continuous mode through the I\textsuperscript{2}C bypass of the MPU-9250. Each chip axis is scaled as $m_{\mathrm{AK},i}=\mathrm{raw}_i\,a_i\,(0.15\ \mu\mathrm{T})$, where $a_i=(\mathrm{ASA}_i-128)/256+1$ is a factory sensitivity adjustment read once from the chip's fuse ROM~\cite{ak8963}. The chip axes are not aligned with the gyroscope and accelerometer (its $x$ and $y$ are swapped and $z$ is inverted), so $\vect{m}_m^B=\big[m_{\mathrm{AK},y},\ m_{\mathrm{AK},x},\ -m_{\mathrm{AK},z}\big]^T$.

\textbf{Earth's magnetic field.} At a fixed location $\lVert\vect{M}_\oplus\rVert=B_0$ is constant, so if the vehicle is rotated, $\vect{M}_\oplus^B$ traces a sphere of radius $B_0$ in the body frame. Eq.~\ref{eq:mag_model} turns this sphere into an ellipsoid: the measured vectors satisfy
\begin{equation}
    \big(\vect{m}_m^B+\vect{O}^B\big)^T\bm A\,\big(\vect{m}_m^B+\vect{O}^B\big)=1,\qquad
    \bm A=\frac{\bm\Gamma^{-T}\bm\Gamma^{-1}}{B_0^2},
    \label{eq:ellipsoid}
\end{equation}
so $-\vect{O}^B$ is the centre of the ellipsoid, and $\bm A$ describes its shape. The calibration fits this ellipsoid to recorded samples and then maps it back to a sphere:
\begin{equation}
    \vect{m}_c^B=\bm W\big(\vect{m}_m^B+\vect{O}^B\big),\qquad \bm W=B_0\bm A^{1/2}=\big(\bm\Gamma\bm\Gamma^T\big)^{-1/2},
    \label{eq:mag_correct}
\end{equation}
where $\bm A^{1/2}$ is the symmetric square root of $\bm A$. The product $\bm W\bm\Gamma$ is orthogonal, so $\vect{m}_c^B$ is the Earth's field $\vect{M}_\oplus^B$ up to a rotation, with $\lVert\vect{m}_c^B\rVert=B_0$. This is the vector used in Eq.~\ref{eq:mag_heading} and Eq.~\ref{eq:kf_adaptiveR}. Two limits follow. First, $B_0$ is the mean size of the fitted ellipsoid (the geometric mean of its semi-axes), not an independent measurement of the local field. Second, an ellipsoid fit cannot observe a rotation of the magnetometer axes relative to the body frame, which would appear as a constant heading error.

\textbf{Least-squares fit.} Let $\vect{m}_{i}$, $i=1,\dots,N$, be the stored samples with centroid $\bar{\vect{m}}$. They are centred and scaled, $\vect{p}_i=(\vect{m}_i-\bar{\vect{m}})/s$ with $s=50\ \mu$T. Centring puts the origin inside the ellipsoid, and scaling keeps the equations well conditioned. Writing the ellipsoid as $\vect{p}^T\bm C\vect{p}+\vect{d}^T\vect{p}=1$, the unknowns enter linearly,
\begin{equation}
    \vect{\phi}_i^T\vect{\xi}=1,
    \label{eq:normal_eq}
\end{equation}
\begin{equation}
    \vect{\phi}_i=\big[p_x^2,\,p_y^2,\,p_z^2,\,2p_xp_y,\,2p_xp_z,\,2p_yp_z,\,p_x,\,p_y,\,p_z\big]_i^T,\
\end{equation}
\begin{equation}
    \vect{\xi}=\big[C_{xx},C_{yy},C_{zz},C_{xy},C_{xz},C_{yz},d_x,d_y,d_z\big]^T,
\end{equation}
and are obtained from the regularized normal equations $\big(\sum_i\vect{\phi}_i\vect{\phi}_i^T+\epsilon\bm 1\big)\vect{\xi}=\sum_i\vect{\phi}_i$ with $\epsilon=10^{-9}N$, solved by Gaussian elimination with partial pivoting~\cite{golub2013}. Completing the square gives the centre $\vect{p}_0=-\tfrac12\bm C^{-1}\vect{d}$ and the shape matrix $\bm A_s=\bm C/\gamma$ with $\gamma=1+\vect{p}_0^T\bm C\vect{p}_0$. The hard-iron vector follows from $-\vect{O}^B=\bar{\vect{m}}+s\,\vect{p}_0$. With the eigendecomposition $\bm A_s=\bm V\operatorname{diag}(w_k)\bm V^T$ (Jacobi method~\cite{golub2013}),
\begin{equation}
    r_{\mathrm{eff}}=(\det\bm A_s)^{-1/6},\qquad B_0=s\,r_{\mathrm{eff}},\qquad
    \bm W=r_{\mathrm{eff}}\,\bm V\operatorname{diag}\big(\sqrt{w_k}\big)\bm V^T.
\end{equation}
This is an unconstrained algebraic fit, as opposed to the ellipse-specific constrained fit of~\cite{fitzgibbon1999}, so the validity of the result is enforced by the acceptance tests below. The computation is done in double precision (software-emulated on the ESP32), which is why it runs in a separate low-priority task.

\textbf{2D and 3D modes.} A ground vehicle only yaws, so the $z$ component of the field is not excited and cannot be calibrated. By default, only the $x$--$y$ components are fitted, with the five unknowns $C_{xx},C_{yy},C_{xy},d_x,d_y$ and $r_{\mathrm{eff}}=(\det\bm A_{s,xy})^{-1/4}$. In this mode, $\bm W$ has no coupling with $z$ and $W_{zz}=1$, the $z$ component of $\vect{O}^B$ keeps its previous value, and the field strength in Eq.~\ref{eq:kf_adaptiveR} is evaluated on the horizontal part only. The full 3D fit takes over when at least 80 samples are stored and their $z$ span is at least half of the mean $x$--$y$ span, which only happens if the vehicle has been tilted. In the 2D mode, an error in the $z$ offset leaks into the tilt-compensated heading and grows with the tilt angle, so it is small as long as the vehicle stays level.

\textbf{On-the-fly operation.}
\begin{enumerate}[leftmargin=*,itemsep=0pt]
    \item Every valid $\vect{m}_m^B$ sample is stored in a 128-sample circular buffer only if it lies farther than $1.5\ \mu$T from all stored samples. A stationary vehicle therefore adds nothing, and the oldest samples are replaced when the buffer is full.
    \item Once per second a background task copies the buffer and attempts a fit, provided there are at least 30 samples and they cover at least 11 of 12 equal angular sectors of the $x$--$y$ plane around their centroid (the vehicle has turned nearly a full circle).
    \item The fit is rejected if $\bm A_s$ is not positive definite, if the ratio of the largest to the smallest semi-axis exceeds 2, if $B_0\notin[10,120]\ \mu$T, or if the RMS relative error
    \begin{equation}
        \varepsilon=\sqrt{\frac1N\sum_i\left(\frac{\lVert\bm W(\vect{m}_i+\vect{O}^B)\rVert}{B_0}-1\right)^2}
    \end{equation}
    exceeds $6\%$ (horizontal norm in the 2D mode).
    \item A new fit replaces the current calibration only if its $\varepsilon$ on the current buffer is lower than that of the current one. It is then blended in, $\vect{O}^B,\bm W,B_0\leftarrow\text{old}+0.5\,(\text{new}-\text{old})$, so that the heading does not jump. The result is handed to the filter through a mutex-protected buffer and is picked up at the next iteration.
\end{enumerate}

\textbf{Startup.} At power-up, the MPU-9250 is reset and identified, the AK8963 is enabled through the bypass mode and its $\mathrm{ASA}_i$ values are read, the gyroscope bias $\vect{b}^B[0]$ is averaged over 200 samples (about 1\,s, with the vehicle standing still), and the calibration task is started. The vehicle must then be rotated in place by about one full turn, so that the sector-coverage test is passed. Until the first valid calibration exists, no magnetometer update can be made and the heading is flagged invalid. At that moment the Kalman filter is initialized from the first accepted $\psi_{\mathrm{mag}}$ (Stage 2). From then on, the calibration keeps refining itself while the vehicle drives. The firmware can also be seeded with a stored calibration.

\textbf{Limitations.} The calibration is static: field changes that depend on the motor current are not modeled and are only handled by the field-strength gate of Eq.~\ref{eq:kf_adaptiveR}. The heading is relative to magnetic north, which is distorted by indoor steel and varies between locations. If the vehicle is relocated, the calibration re-adapts as old samples are replaced in the buffer. Furthermore, in preliminary unlogged tests the 3D mode appeared to output much higher bias than the 2D mode, which we suspect is contributed to by the inaccuracy of the up direction estimation in Eq.~\ref{eq:universal_up}.

\subsubsection{Experimental Validation of the Heading Estimator}
\label{subsubsec:heading_validation}
 
The estimator was tested on a breadboard rather than on the vehicle, with an MPU-9250 and a CMPS14 taped side by side on one board and logged over UART at 20\,Hz. All processing was done offline. No external angle reference was available, so the experiments measure agreement between the two sensors and stability at rest, not absolute accuracy. The alignment offset between the two sensors was not captured on-line in the reported run. It was instead computed offline as the circular mean of the CMPS14--MPU difference over the first 23\,min of stationary data (about $-139^\circ$, with a peak-to-peak variation of about $2^\circ$), and the aligned CMPS14 heading and the blended heading were recomputed from it. This is exact for the blended heading, because the blend weight depends only on the MPU yaw rate.
 
\begin{figure}[H]
\centering
\includegraphics[width=0.9\linewidth]{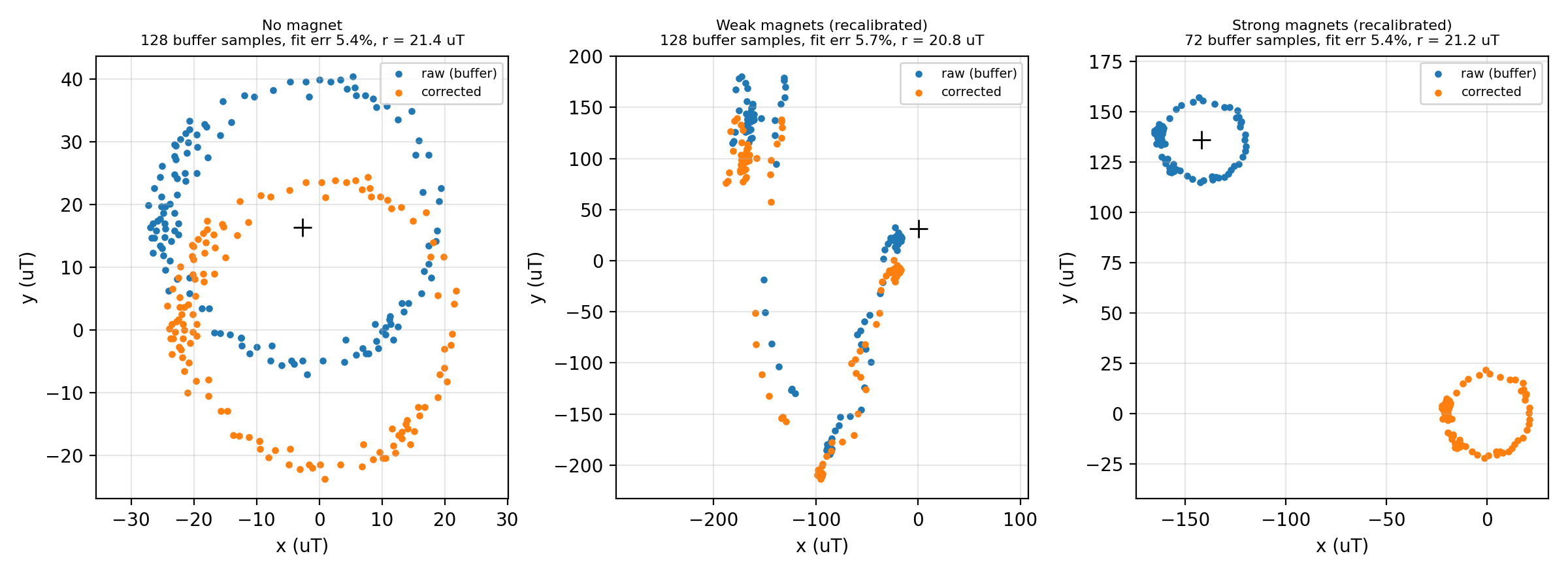}
\caption{Magnetometer calibration buffers. Blue: raw samples; orange: samples corrected with the last accepted calibration of each segment; the black cross marks the fitted hard-iron centre.}
\label{fig:val_cal}
\end{figure}

\begin{figure}[H]
\centering
\begin{subfigure}[t]{0.24\linewidth}
    \centering
    \includegraphics[width=\linewidth]{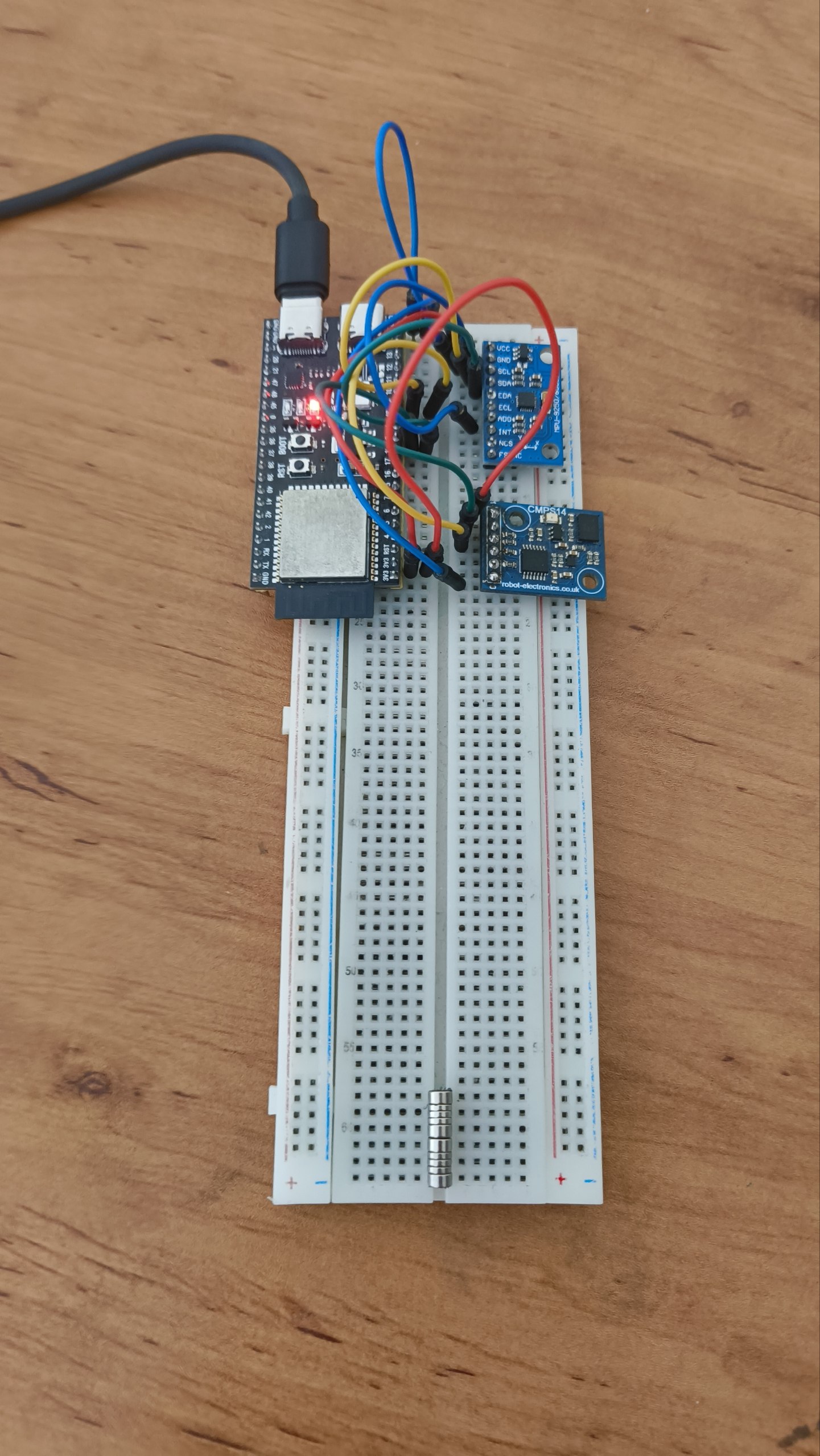}
    \caption{Weak magnets.}
    \label{fig:weak_mag}
\end{subfigure}
\hspace{0.5cm}
\begin{subfigure}[t]{0.24\linewidth}
    \centering
    \includegraphics[width=\linewidth]{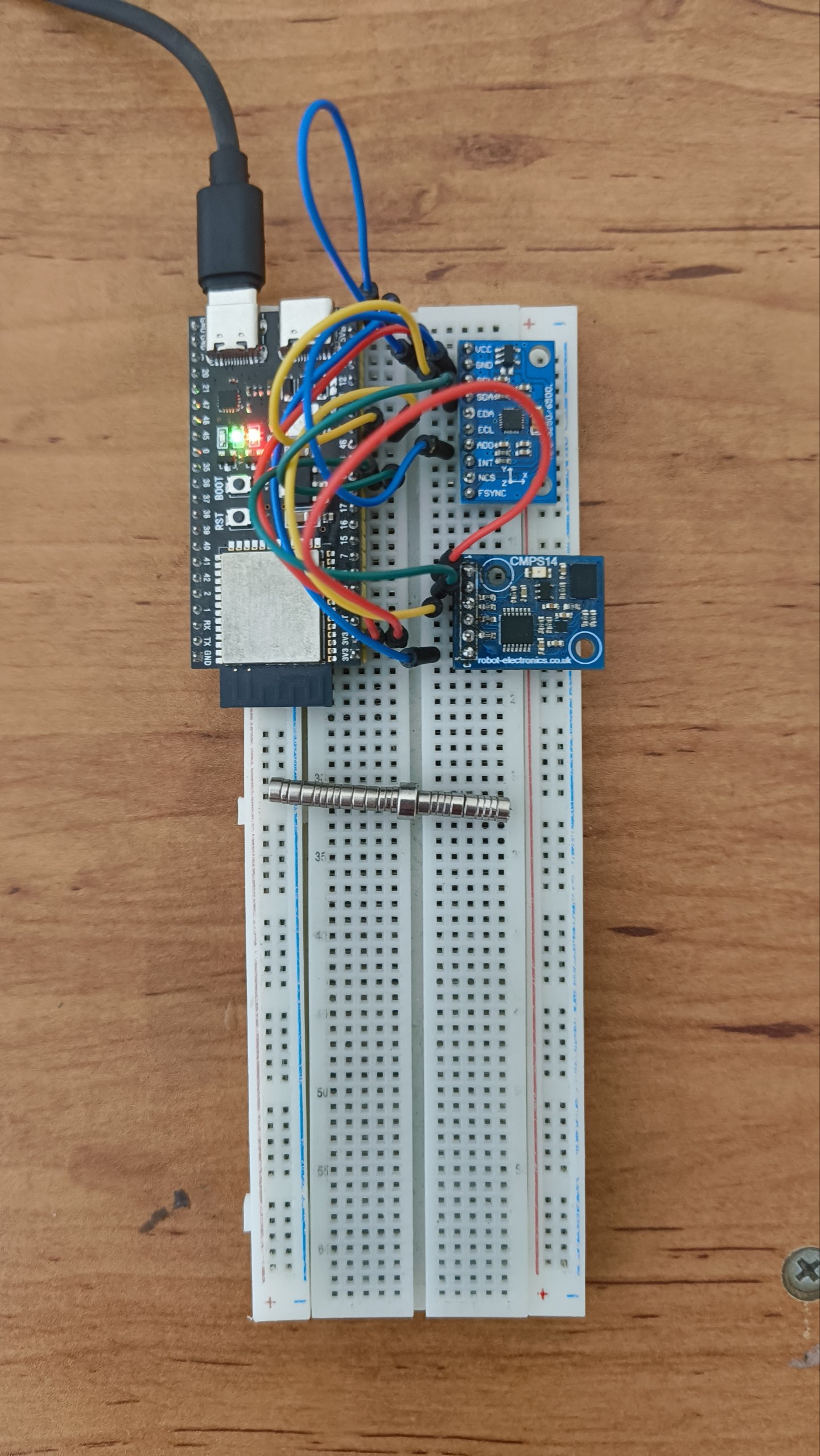}
    \caption{Strong magnets.}
    \label{fig:strong_mag}
\end{subfigure}
\caption{Breadboard setup of the two magnetic disturbance tests.}
\label{fig:val_setup}
\end{figure}

Fig.~\ref{fig:val_cal} shows the magnetometer samples held in the calibration buffer for the three conditions tested, reconstructed offline from the logged raw samples with the same storage rule as the firmware (128-sample circular buffer, minimum spacing of $1.5\,\mu$T). The fit error and field radius of the last accepted calibration of each segment are given in the panel titles. Without magnets the raw samples form an offset ring, and the corrected samples are centred on the origin. The two magnet conditions are shown in Fig.~\ref{fig:val_setup}. The weak magnets leave the horizontal field close to the undisturbed value of about $38\,\mu$T, whereas the strong magnets raise it to about $215\,\mu$T. The corrected weak-magnet samples did not form a circle, while those of the other two cases did. The reason is not yet known.

\begin{figure}[H]
\centering
\includegraphics[width=0.85\linewidth]{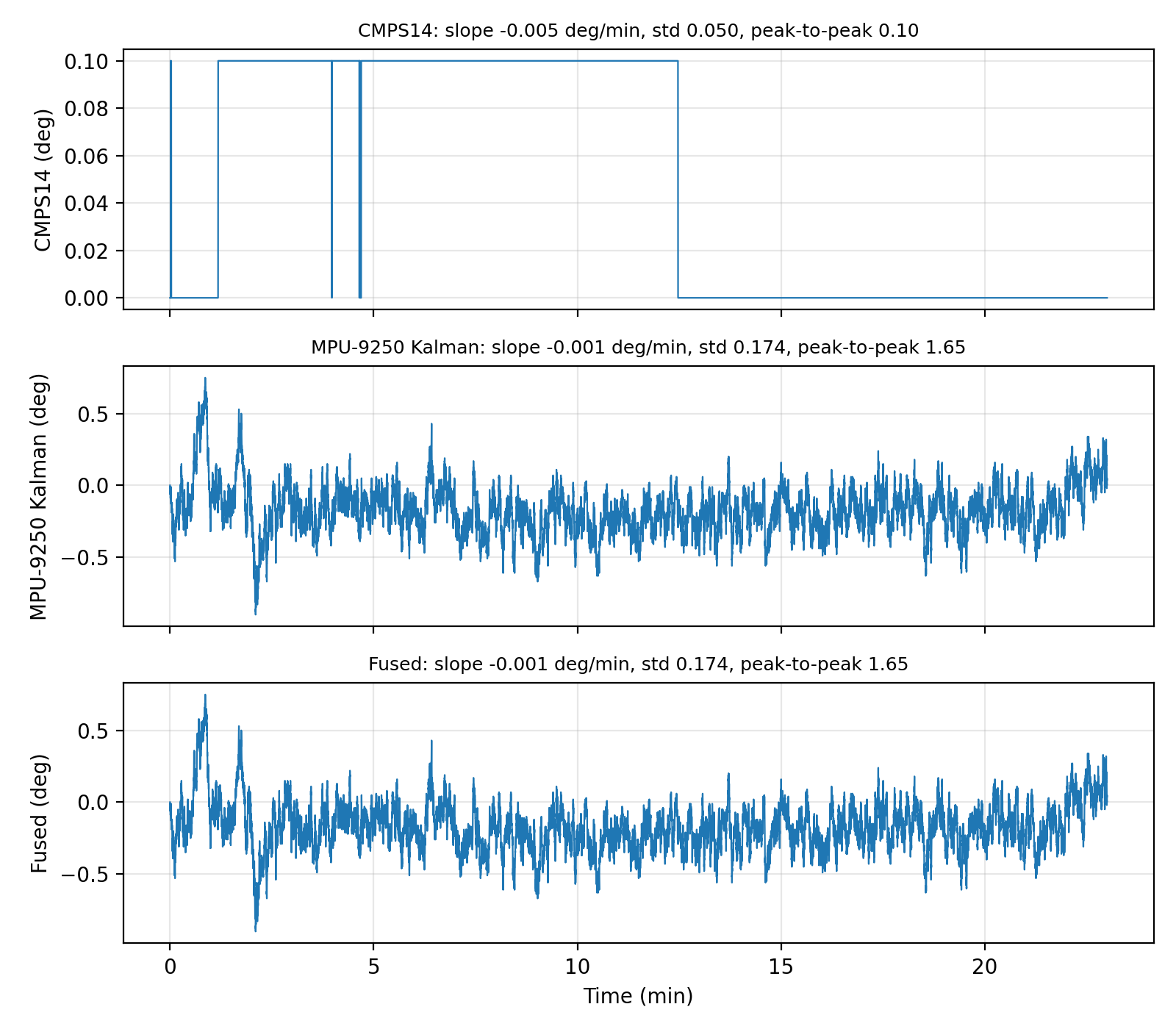}
\caption{Stationary heading change of the CMPS14, the MPU Kalman estimate and the blended heading over 23\,min. The blended heading coincides with the MPU estimate because the blend weight is zero at rest.}
\label{fig:val_drift}
\end{figure}

Fig.~\ref{fig:val_drift} shows 23\,min with the board untouched (the last minutes of the recording, when the operator returned to the desk, are excluded). The MPU Kalman heading had a slope of $-0.001^\circ$/min, a standard deviation of $0.17^\circ$ and a peak-to-peak variation of $1.65^\circ$. The CMPS14 had a slope of $-0.005^\circ$/min and a peak-to-peak variation of $0.10^\circ$, which is its output resolution. At rest the blend weight is zero, so the blended heading is identical to the MPU heading. Our design assumed that the CMPS14 drifts considerably more than the filtered MPU at standstill, and the blend gives the MPU full weight at low rates for that reason. This run does not support the assumption. In a second 14\,min run (during which a cable was briefly touched) the CMPS14 again varied by $0.1^\circ$ against about $1.9^\circ$ for the MPU. In both recordings the CMPS14 was at least as stable as the MPU, so the blend gave no measurable benefit at rest. A further 30\,min run showed about $9^\circ$ of CMPS14 drift, so CMPS14 stability seems to vary between runs. The measured MPU standard deviation is of the same order as the $0.26^\circ$ predicted in Stage~2, although the two quantities are not strictly the same (the prediction is an estimation error with respect to the true heading, which was not measured here).
 
\begin{figure}[H]
\centering
\includegraphics[width=0.85\linewidth]{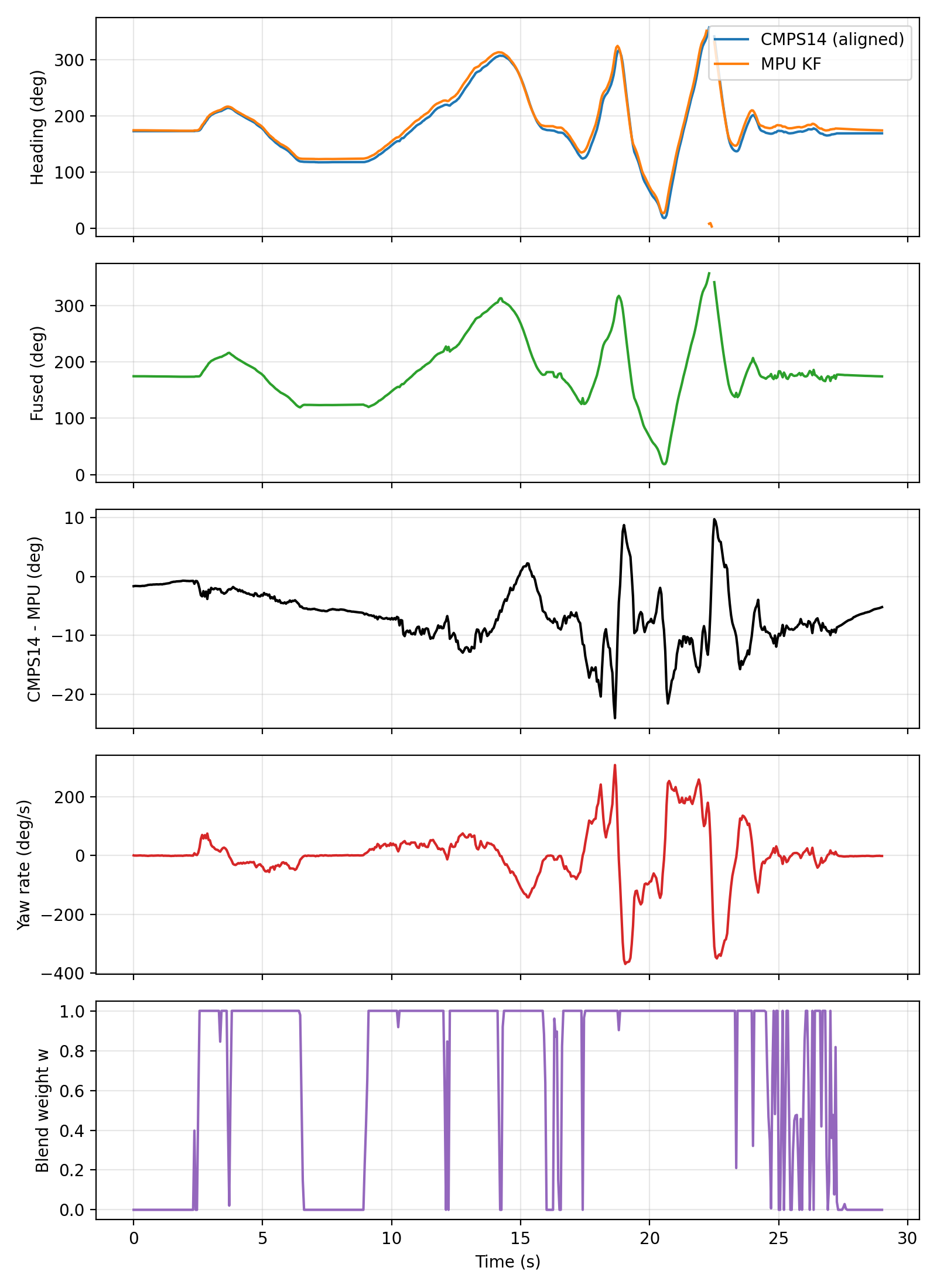}
\caption{Manual rotation: aligned CMPS14 and MPU headings, blended heading, their difference, the yaw rate, and the blend weight $w$.}
\label{fig:val_dyn}
\end{figure}

Fig.~\ref{fig:val_dyn} shows a 29\,s rotation by hand with yaw rates up to about $350^\circ$/s, using blend thresholds of $3^\circ$/s and $15^\circ$/s. After alignment the two headings stay within about $\pm25^\circ$ of each other and mostly within $10^\circ$, and the difference drifts slowly toward $-10^\circ$. The blend weight follows the rate as designed, so the blended heading follows the MPU at low rates and the CMPS14 at high rates. Without a ground-truth angle we cannot say which sensor is closer to the true heading, only that they agree. The experiment therefore does not show that the blend improves on either sensor alone.
 
\begin{figure}[H]
\centering
\includegraphics[width=0.85\linewidth]{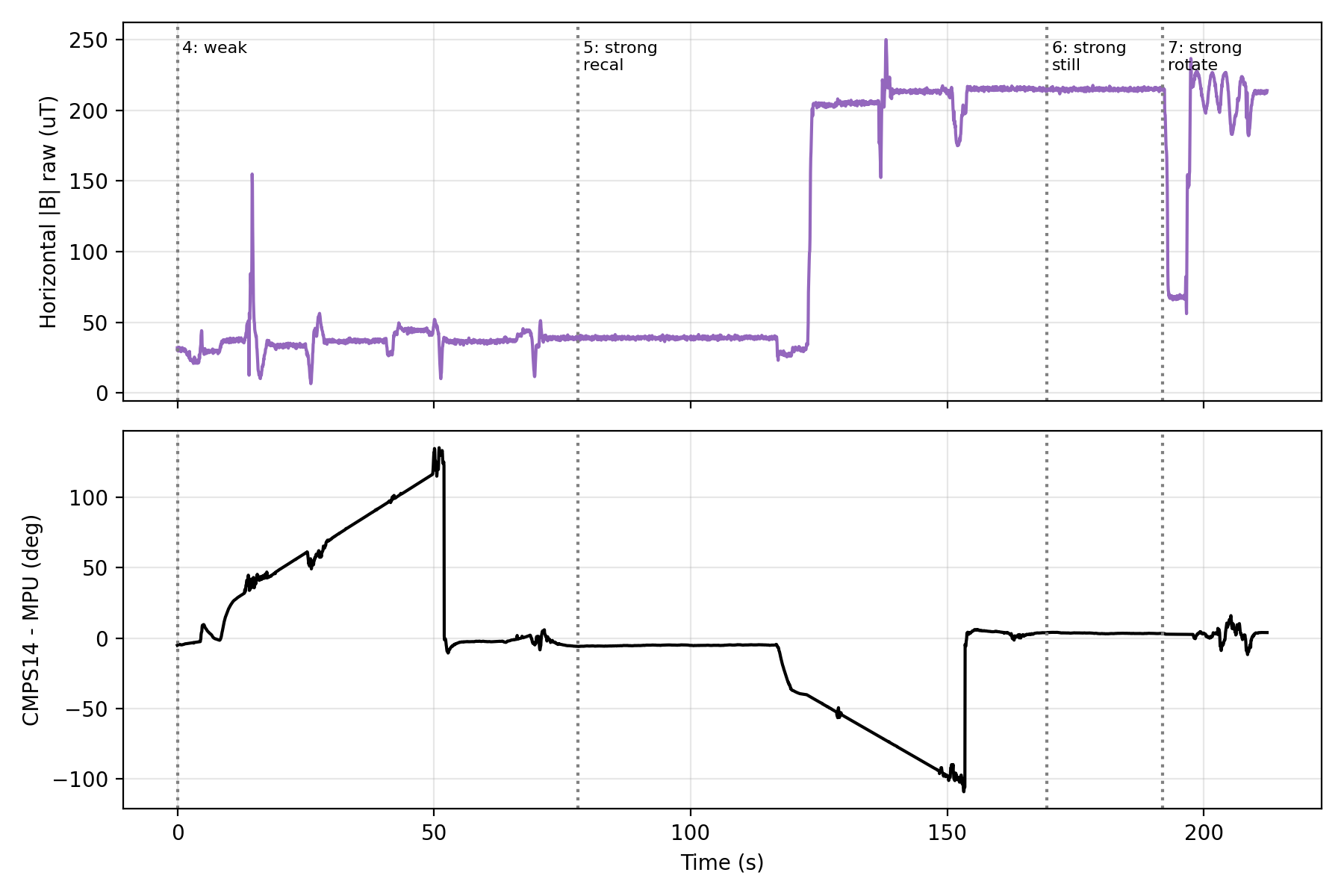}
\caption{Horizontal field strength (top) and CMPS14--MPU difference (bottom) across the magnet tests. The vertical lines mark the start of the weak-magnet test, the start of the strong-magnet phase (the magnets were introduced about 45\,s later), the strong-magnet stationary test and the strong-magnet rotation.}
\label{fig:val_mag}
\end{figure}

The magnet tests are summarised in Fig.~\ref{fig:val_mag}. Placing the magnets near the board leaves the MPU heading uncorrected until the calibration is refitted: in both cases the CMPS14--MPU difference grew almost linearly, by about $100$--$130^\circ$ in $35$--$45$\,s ($\approx 3^\circ$/s) while the board stood still, and returned to a few degrees as soon as a new calibration became valid after a manual reset. We believe the magnetometer samples were being rejected by the field-strength gate during these intervals, so the filter was integrating the gyroscope alone, but we did not verify this directly. With the strong magnets and a new calibration the MPU held its heading within about $1^\circ$ over 22\,s at rest and followed the CMPS14 within a few degrees during rotation. A plausible reason is that magnets fixed to the board act as an additional constant hard-iron offset, which the calibration removes. The calibration does not recover by itself, since the old buffer still holds samples from the undisturbed field and a new fit is accepted only after a full rotation. In the present implementation a lasting disturbance therefore requires a manual reset.
 
These results come from a single board and a small number of runs, without an external reference, and only the planar (2D) calibration mode was exercised. They should be read as a first check of the implementation and not as a characterisation of its accuracy.

\subsubsection{Communication Protocol and Data Packet Structure}
\label{subsubsec:commprotocol}

Data transfer between processing units is implemented over two UART channels, following a structured packet-based protocol. ZERO periodically sends data packets (\texttt{p1}) containing ToF, compass, and encoder readings to CONTROL, which checks packet integrity (using checksum methods) and forwards them as packet \texttt{p1} to Jetson. CONTROL also receives control and lighting commands (\texttt{p2}) from Jetson, which upon arrival extracts data, actuates NeoPixel and Servo Motor directly, and sends DC motor speed as packet \texttt{p3} to ZERO.

Each communication packet follows a fixed structure designed for robustness and low latency. The format is summarized as:
\[
\texttt{\$\textbf{d\textsubscript 1}!\textbf{d\textsubscript 2}!...\textbf{d\textsubscript n}!\textbf{chk8}\#.....}
\]
where \texttt{\$} is the start byte (synchronization marker), \texttt{\textbf{d\textsubscript{i}}} are the data fields separated by \texttt{!}, \texttt{chk8} is an 8-bit additive checksum for data integrity, \texttt{\#} is the end byte (termination marker), and \texttt{.} denotes padding characters used to maintain fixed payload length.

This structure ensures byte-level synchronization and error detection even under variable serial latency conditions. Message acknowledgment is intentionally minimized to reduce round-trip delay, while periodic redundancy provides fault tolerance.

Both UART links use 115200\,baud, 8N1. Table~\ref{tab:packets} lists the packets. A receiver waits for the start byte, reads a fixed-length frame, and then flushes its RX buffer, so only the most recent frame is processed. The integrity field is an 8-bit additive checksum (sum of the numeric fields modulo 256). The sensor packet \texttt{p1} is verified on the Jetson; \texttt{p2} and \texttt{p3} are verified by their receivers (CONTROL and ZERO).

Forward latency (time passed between a ZERO packet transmission and a Jetson packet reception) was measured to have mean \textbf{18.2\,ms} with standard deviation of a \textbf{4.0\,ms}. The backward latency (time passed between a Jetson packet transmission and a ZERO packet reception) was shorter, owing to the smaller packet size. The mean was measured to be around \textbf{9.2\, ms}, with standard deviation of \textbf{3.4\, ms}. These measurements were performed by simultaneously hardware interrupting all three devices, and setting the interrupted time to zero, and continuously polling data transmission and reception time across the three devices.

\begin{table}[H]
\centering
\caption{Packet formats. Wire time is for a fixed-length frame at 115200\,baud, 8N1 (10 bits per byte).}
\label{tab:packets}
\small
\renewcommand{\arraystretch}{1.1}
\begin{tabular}{@{}lcccp{6.3cm}@{}}
\toprule
\textbf{Packet} & \textbf{Direction} & \textbf{Length} & \textbf{Wire time} & \textbf{Fields} \\ \midrule
\texttt{p1} & ZERO$\to$CONTROL$\to$Jetson & 85\,B & 7.4\,ms/hop & 12 ToF distances (mm), compass bearing (0.1$^\circ$), encoder count, speed, motor-enable flag, time elapsed, checksum \\
\texttt{p2} & Jetson$\to$CONTROL & 25\,B & 2.2\,ms & steering command (0--255), buzzer, motor command ($\pm$1023), light mode, ToF bitmask, checksum \\
\texttt{p3} & CONTROL$\to$ZERO & 18\,B & 1.6\,ms & motor command, ToF bitmask, checksum \\
\bottomrule
\end{tabular}
\end{table}

\paragraph{Startup Sequences, Edge Cases, and Fault Protection.}

At startup, CONTROL initiates a handshake with ZERO to verify connectivity and sensor readiness before enabling motor actuation. All sensors are sequentially initialized and validated before control commands are accepted.  
Emergency stop and fail-safe routines are integrated into CONTROL's firmware: if serial communication from Jetson is lost for more than 500\,ms, or if a fault flag is raised by ZERO, CONTROL immediately halts motor drive signals and centers the steering servo. These measures ensure safe degradation of operation during transient or fault conditions.

\subsection{Robot Operating System (ROS) Architecture}
\label{subsec:ros}

The autonomous vehicle software stack is implemented using the Robot Operating System (ROS) Noetic~\cite{quigley2009ros}, running on Ubuntu~20.04 LTS on the Jetson Orin NX onboard computer. ROS provides a modular, distributed middleware that enables perception, decision-making, and actuation to run in parallel while exchanging data through topics, shared memory, and global parameters.

\subsubsection{System Overview}

\begin{figure}[H]
    \centering
    \includegraphics[width=0.85\linewidth]{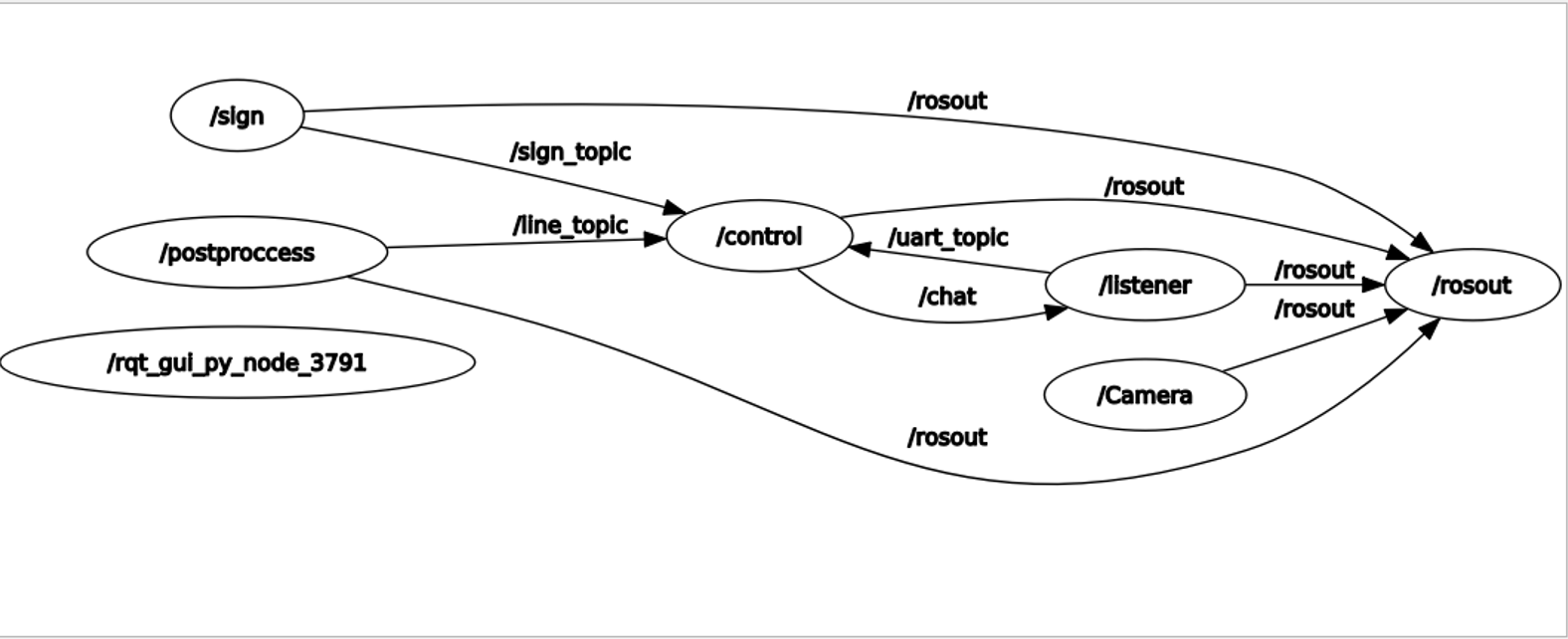}
    \caption{Graph of ROS nodes and topic-level publish--subscribe relationships.}
    \label{fig:rqt_graph}
\end{figure}

The system consists of seven ROS nodes:
\texttt{Camera}, \texttt{Sign}, \texttt{Network}, \texttt{Postprocess}, \texttt{Crosswalk}, \texttt{Control}, and \texttt{Listener}.
Each node executes as an independent process but participates in a unified computation graph, visualized in Fig.~\ref{fig:rqt_graph}.  
Large data objects such as RGB and depth frames are communicated through shared memory to minimize copying overhead, while status, events, and processed features are sent through ROS topics and parameters.

\begin{table}[H]
\centering
\caption{Summary of ROS nodes and their functions}
\renewcommand{\arraystretch}{1.2}
\begin{tabular}{@{}p{2.2cm}p{3.2cm}p{8cm}@{}}
\toprule
\textbf{Node} & \textbf{Inputs} & \textbf{Primary Function / Outputs} \\ \midrule
Camera & Intel RealSense D455 & Captures synchronized RGB and depth frames at $848\times480$ and writes them to shared memory. Publishes a startup-ready parameter after the first valid frame. \\
Sign & Shared memory frames & Performs traffic sign detection. Publishes \texttt{sign\_topic} (sign type and distance) and \texttt{cross\_input} to trigger crosswalk processing. \\
Network & Shared memory RGB frame & Runs lane segmentation model and writes a $(32\times64\times7)$ tensor to shared memory. \\
Postprocess & Network output + RGB frame & Extracts lane offset and heading angle, published on \texttt{line\_topic}. \\
Crosswalk & RGB frame + \texttt{cross\_input} & Computes crosswalk angle and position, and receives the \texttt{intersection} parameter from Control. \\
Control & Perception topics + \texttt{uart\_topic} & Computes steering, throttle, and mode, publishing serial data to the Listener. \\
Listener & \texttt{chat} topic & Converts control outputs to serial packets and republishes hardware feedback on \texttt{uart\_topic}. \\
\bottomrule
\end{tabular}
\end{table}

\subsubsection{Node-Level Data Flow}

The \textbf{Camera} node initializes the Intel D455 at $848\times480$, the operating point providing the highest depth accuracy. It writes RGB and depth frames directly to shared memory, and after the first valid capture sets a ROS parameter indicating that dependent nodes may start processing. This prevents premature access to uninitialized memory.

The \textbf{Sign} node reads the current frames from shared memory and generates two outputs:  
(1) sign type and distance on \texttt{sign\_topic}, and  
(2) a binary crosswalk trigger on \texttt{cross\_input}.

The \textbf{Network} node receives the latest RGB frame and performs neural inference, producing a $(32\times64\times7)$ feature tensor stored in shared memory for downstream use.

The \textbf{Postprocess} node extracts lane geometry from the tensor: lateral offset and heading angle. These are published to \texttt{line\_topic}, consumed by the Control layer.

The \textbf{Crosswalk} node uses shared memory frames and \texttt{cross\_input} to compute crosswalk angle and position. Additionally, it receives a runtime parameter named \texttt{intersection} from the Control node indicating whether the vehicle is navigating a junction.

The \textbf{Control} node subscribes to \texttt{line\_topic}, \texttt{sign\_topic}, crosswalk outputs, and hardware feedback from \texttt{uart\_topic}. It computes high-level actuation commands, which are published as \texttt{chat} messages.

The \textbf{Listener} node interfaces directly with the microcontrollers, translating \texttt{chat} into serial commands and publishing returning sensor feedback back onto \texttt{uart\_topic}.
The \texttt{Listener} node waits for the camera-ready parameter, opens the Jetson UART (\texttt{/dev/ttyTHS0}, 115200\,baud), and sends the latest command packet to CONTROL on every loop iteration after clamping each field to its valid range. Incoming \texttt{p1} packets (section~\ref{subsubsec:commprotocol}) are checked against their checksum and republished as 17 integers on \texttt{uart\_topic}; if corrupted bytes are detected, the port is closed, reopened, and flushed.

\begin{figure}[H]
    \centering
    \includegraphics[width=0.85\linewidth]{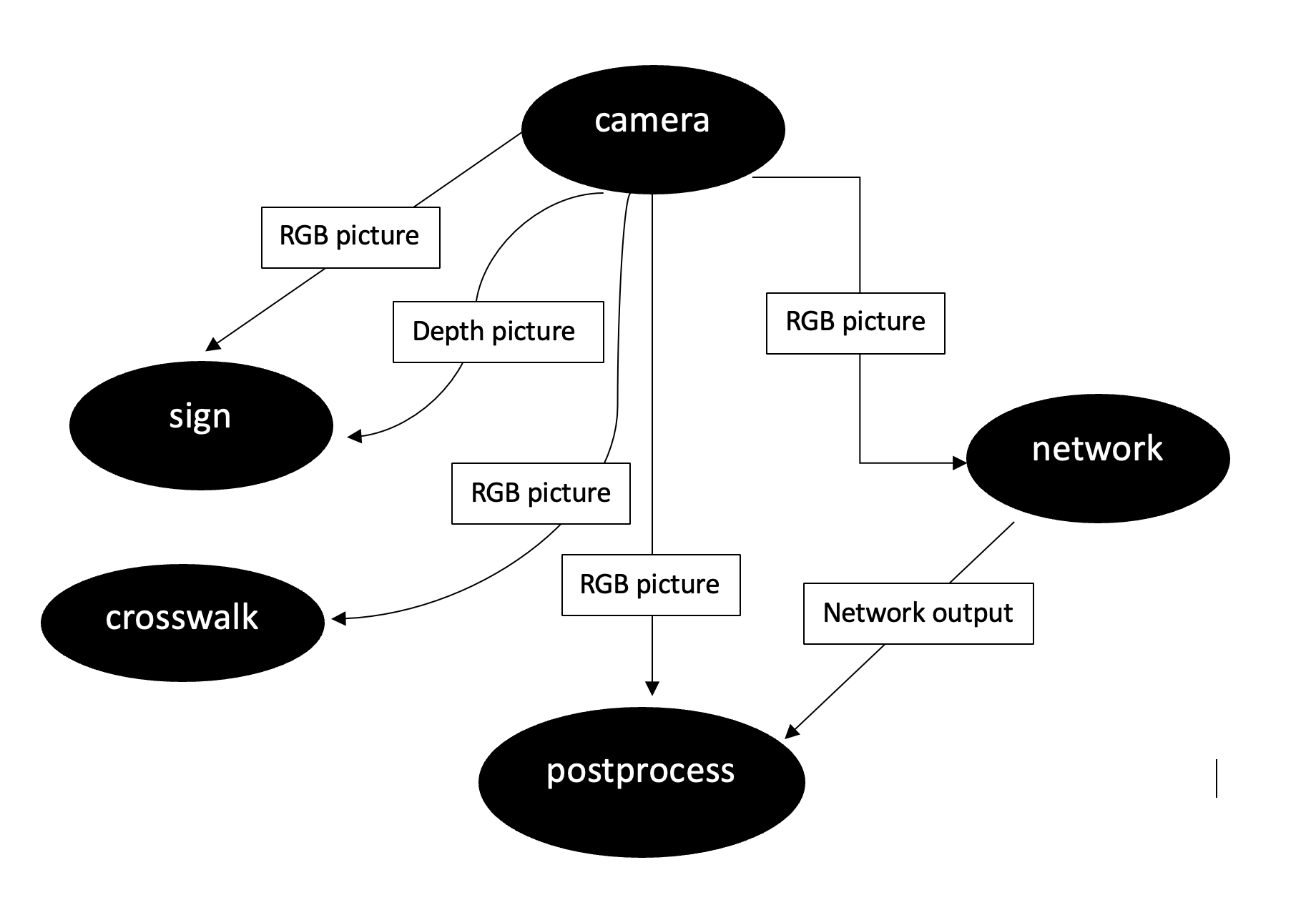}
    \caption{Shared memory and data exchange structure among ROS nodes.}
    \label{fig:shared_memory}
\end{figure}

The shared-memory relationships are shown in Fig.~\ref{fig:shared_memory}.

\subsubsection{Launch Configuration and Fault Handling}

All nodes are executed through a unified launch file in \texttt{launch/}.  
The configuration defines output behavior, restart rules, and global parameter initialization.

Because the Camera node allocates the shared memory region, if it restarts, the memory pointer changes.  
To avoid inconsistent pointers, the system is configured so that a Camera crash triggers a complete system restart, while crashes in other nodes cause only those nodes to restart individually.

\begin{figure}[H]
    \centering
    \includegraphics[width=0.85\linewidth]{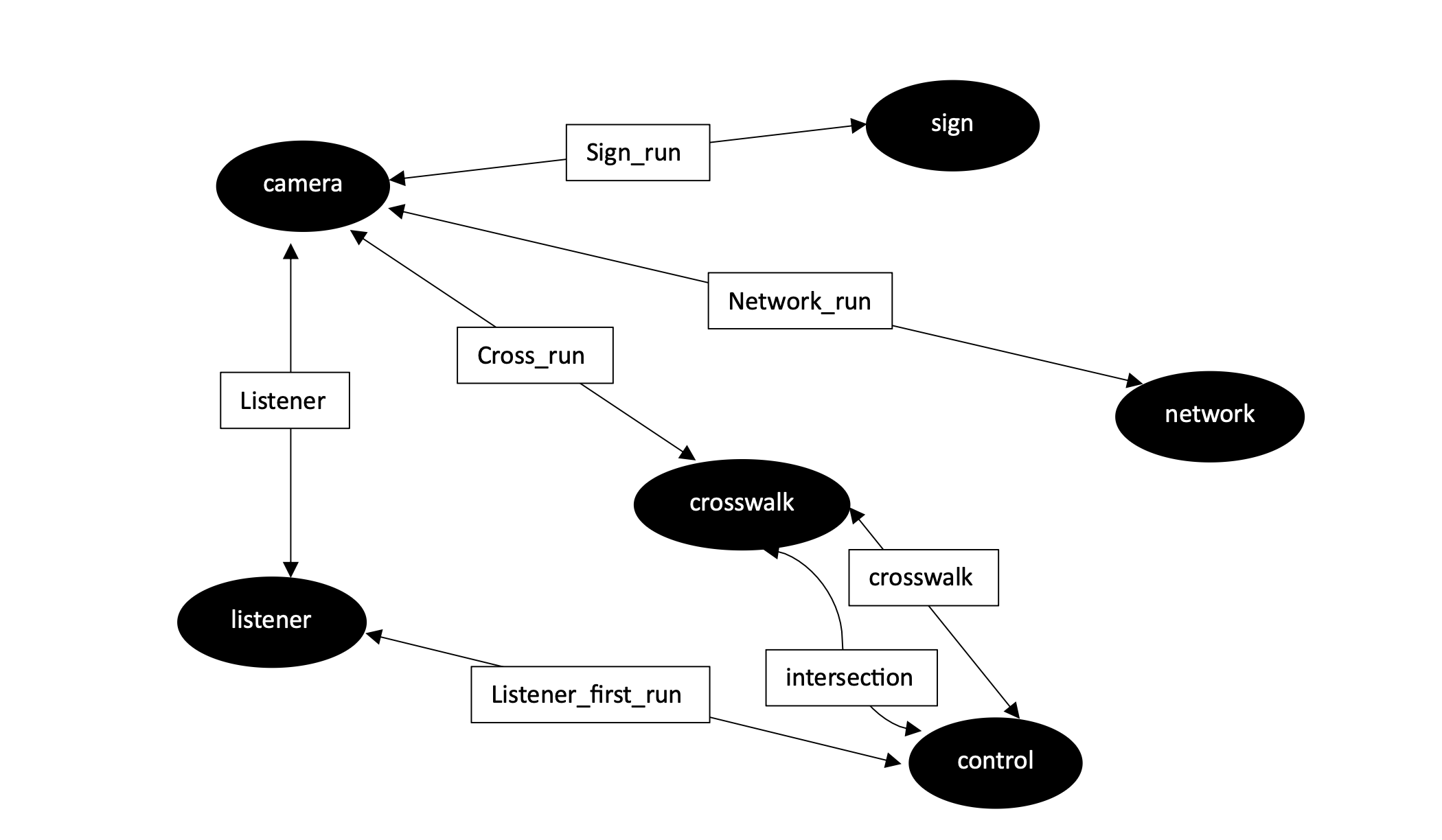}
    \caption{Global parameter exchange for cross-node synchronization.}
    \label{fig:ros_parameters}
\end{figure}

Global shared parameters (calibration constants, visualization toggles, and processing modes) are propagated through the same launch configuration and are accessible system-wide, as shown in Fig.~\ref{fig:ros_parameters}.

\subsubsection{Timing, Synchronization, and Performance}

Real-time operation requires matching node execution rates.  
The lane network (\texttt{Network}) runs at approximately 13\,Hz and is therefore the bottleneck for downstream modules. Nodes such as \texttt{Postprocess} and \texttt{Control} can reach 30\,Hz, but first verify whether new data is available before computing. If the network output has not changed, they defer processing, preventing redundant work and improving CPU efficiency.

Conversely, the Camera must operate faster than all dependent nodes to minimize data age and reduce perception-to-actuation latency.

This combined scheduling strategy produces deterministic timing, low redundant computation, and predictable behavior across the full perception and control pipeline.

\section{Computer Vision and Perception}
\label{sec:cv}
The perception stack serves as the primary sensory input for the autonomous vehicle, mapping the drivable area and enabling adherence to traffic rules. We designed a custom vision pipeline centered around a modified Point Instance Network (PINet)~\cite{ko2022pinet}, optimized specifically for the sharp curvature and unique lighting conditions of the competition track. The stack is entirely vision-driven: a single Intel RealSense D455 provides synchronized RGB and depth, and three cooperating perception paths---lane detection, traffic sign and signal recognition, and crosswalk detection---run concurrently and publish their state to the controller through a zero-copy shared-memory bus.

This section is organized to follow a single camera frame through the full pipeline. We first describe acquisition, then the lane-detection network architecture and its mathematical training objective, the deterministic geometric post-processing that converts network outputs into physical control variables, the sign and crosswalk perception paths, the inter-process communication layer that binds them, and finally the experimental validation of the trained model. Throughout, we emphasize the engineering rationale behind each design decision, since the dominant constraint---real-time inference on an embedded Jetson Orin at $30$~FPS---shaped nearly every choice in the stack.

\subsection{Acquisition System}
Visual data is acquired via an Intel RealSense D455 depth camera~\cite{keselman2017}, selected for its global-shutter RGB sensor and its wide-baseline active stereo depth module. The global shutter is a deliberate choice for a moving platform: at competition speeds a rolling-shutter sensor introduces skew artifacts on vertical lane markings during yaw motion, which would corrupt the sub-pixel key point regression discussed in section~\ref{subsec:loss}. The camera streams synchronized $480\times848$ RGB and depth frames at $30$~FPS, with the depth stream hardware-aligned to the color optical frame so that any RGB pixel can be dereferenced directly into a metric depth value. This alignment is what later enables the depth-gated sign validation of section~\ref{subsec:signs} without a separate extrinsic calibration step.

Unlike traditional pipelines that crop the input to a Region of Interest (ROI), we process the full sensor resolution (resized to the network input of $512\times256$). We deliberately avoided ROI cropping to maintain the maximum possible field of view. This ensures the network can utilize peripheral visual cues---essential for detecting the entry points of sharp hairpin turns---which would otherwise be discarded by aggressive cropping. The trade-off is a larger input tensor and therefore a heavier front-end; we absorb this cost in the aggressive stride-2 stem described in Table~\ref{tab:complexity} rather than by discarding field of view.

\begin{figure}[H]
    \centering
    \includegraphics[width=0.6\linewidth]{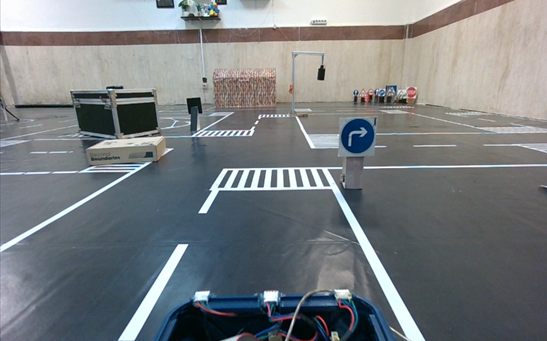}
    \caption{Raw RGB frame captured from the RealSense D455 on the competition track. The full, uncropped field of view retains peripheral cues (distant signs, crosswalk, hairpin markings) at the cost of a static foreground occlusion from the chassis, which the network learns to suppress.}
    \label{fig:rgb_sample}
\end{figure}

Fig.~\ref{fig:rgb_sample} shows a representative raw RGB frame captured on the competition track. The scene is intentionally cluttered---lane markings, a pedestrian crosswalk, a mandatory-direction sign, and a bank of distant traffic signs are all simultaneously visible---illustrating the range of cues the perception stack must disentangle from a single monocular frame. The lower portion of the frame is partially occluded by the vehicle's own chassis and wiring harness; because we do not crop, the network must (and does) learn to ignore this static foreground rather than having it removed by a hand-tuned mask.

\subsection{Lane Detection Architecture}
We utilized a deep learning approach based on the Stacked Hourglass architecture. Standard lane detectors often struggle with the tight turning radii ($<1\,\mathrm{m}$) found in 1:10 scale tracks. To address this, we engineered specific modifications into the PINet backbone.

\subsubsection*{Layer-wise Complexity Analysis}
The summary below details the structural evolution of the feature maps as they propagate through the network. The Stem layer performs aggressive downsampling to reduce the memory footprint. The architecture then transitions into feature extraction, where HG-Block 3 introduces specific dilated convolutions. Finally, the Heads project these deep features into the specific 7-channel output.

\begin{table}[H]
\centering
\caption{Layer-wise Complexity Analysis}
\label{tab:complexity}
\begin{tabular}{@{}l l l l p{5cm}@{}}
\toprule
\textbf{Block Name} & \textbf{Type} & \textbf{Input Res.} & \textbf{Out Ch.} & \textbf{Engineering Rationale} \\
\midrule
Stem & Conv2D + PReLU & $512\times256$ & 64 & Aggressive downsampling to reduce memory footprint. \\
\midrule
HG-Block 1 & Standard Hourglass & $64\times32$ & 128 & Initial coarse feature extraction. \\
\midrule
HG-Block 2 & Standard Hourglass & $64\times32$ & 128 & Refinement of features with skip connections. \\
\midrule
HG-Block 3 & Dilated Hourglass & $64\times32$ & 128 & Custom: Uses dilation rates (2, 4) to expand receptive field for sharp turns. \\
\midrule
Heads & Conv 1x1 & $64\times32$ & 7 & Projects features into: 1x Confidence, 2x Offsets, 4x Embeddings. \\
\bottomrule
\end{tabular}
\end{table}

\begin{figure}[H]
    \centering
    \parbox{0.8\textwidth}{\centering
    \includegraphics[width=0.6\textwidth]{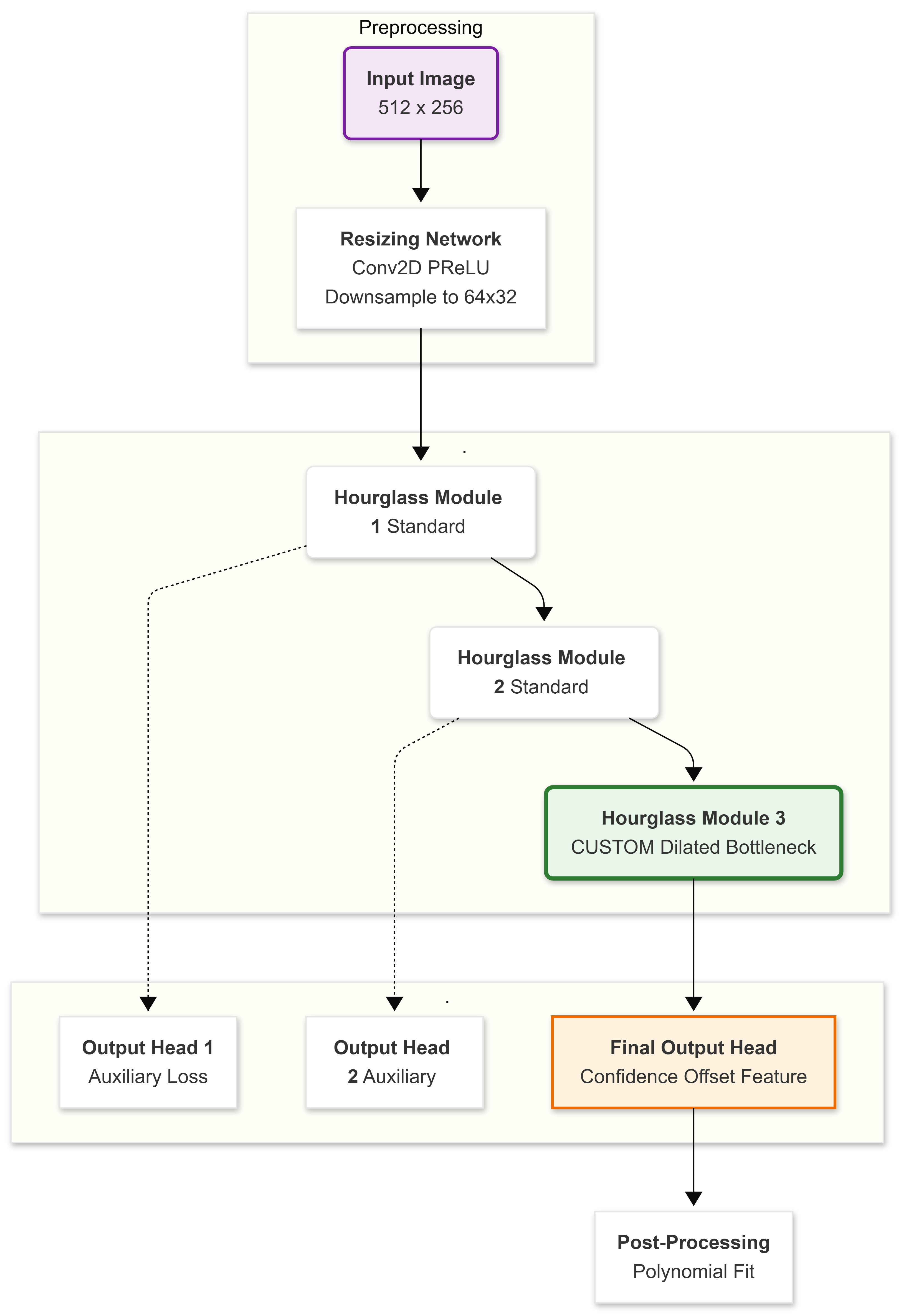}
    }
    \caption{Top-Down Network Data Flow. The high-resolution input ($512\times256$) is first processed by a Resizing Network to reduce spatial dimensions to $64\times32$, optimizing inference speed on the Jetson Orin. The core backbone consists of three Stacked Hourglass modules. The final module (Hourglass 3) is engineered with a Dilated Bottleneck.}
    \label{fig:hourglass-pipeline}
\end{figure}

\subsubsection{Network Design: The ``Dilated'' 3-Stack Hourglass}
Our model consists of three stacked hourglass modules. While the original PINet implementation uses standard convolutions, we engineered the bottleneck layers to utilize dilated convolutions~\cite{yu2016}. The overall data flow is shown in Fig.~\ref{fig:hourglass-pipeline}.

\textbf{The Engineering Challenge:} Standard convolutions in a compact network often lack the receptive field necessary to ``see'' the curvature of a sharp turn before the vehicle enters it.

\textbf{The Solution:} By applying dilation rates of 2 and 4 in the bottleneck layers, we expanded the network's receptive field without increasing the parameter count. This allows the model to capture the global context of the lane geometry (e.g., a sharp U-turn) while retaining local spatial resolution for precise localization.

\textbf{Reliability Strategy:} We opted to train a 3-Hourglass network from scratch rather than distilling a larger 4-Hourglass teacher network. This ``direct training'' approach ensured that the validation metrics observed during training perfectly matched the inference behavior on the embedded hardware.

\subsection{Mathematical Formulation of the Loss Function}
\label{subsec:loss}
To ensure precise key point estimation and instance segmentation, we implemented a multi-task loss function. This function is applied to the three output branches (Confidence, Offset, and Embedding) of each hourglass module.

\paragraph{Confidence Loss.}
The confidence branch predicts the probability of a key point existing in a given grid cell. We utilize a split loss formulation to handle the class imbalance. The Existence Loss $L_{\mathrm{exist}}$ penalizes errors only in cells where key points actually exist ($G_e$):

\begin{equation}
L_{\mathrm{exist}}=\frac{1}{N_{e}}\sum_{c_{e}\in G_{e}}(c_{c}^{*}-c_{c})^{2}
\end{equation}

The Non-Existence Loss ($L_{\mathrm{non\_exist}}$) minimizes false positives in background cells $G_{n}$ with a regularization term to prevent overfitting to zero:

\begin{equation}
L_{\mathrm{non\_exist}}=\frac{1}{N_{n}}\sum_{c_{c}\in G_{n}}(c_{c}^{*}-c_{c})^{2}+0.00001\cdot\sum_{c_{c}\in G_{n}}c_{c}^{2}
\end{equation}

\paragraph{Offset Loss.}
To mitigate quantization error, the network predicts normalized offsets in the range $0$--$1$. The loss is calculated only for cells containing key points:

\begin{equation}
L_{\mathrm{offset}}=\frac{1}{N_{e}}\sum_{c_{x}\in G_{e}}(c_{x}^{*}-c_{x})^{2}+\frac{1}{N_{e}}\sum_{c_{y}\in G_{e}}(c_{y}^{*}-c_{y})^{2}
\end{equation}

\paragraph{Embedding Feature Loss.}
To distinguish between multiple lane instances, we employ a discriminative loss function~\cite{debrabandere2017}. This pulls embeddings of the same instance ($I_{ij}=1$) closer together and pushes embeddings of different instances ($I_{ij}=0$) apart by a margin $K$:

\begin{equation}
L_{\mathrm{feature}}=\frac{1}{N_{e}^{2}}\sum_{i}^{N_{e}}\sum_{j}^{N_{e}}l(i,j)
\label{eq:feature}
\end{equation}

Where the pairwise distance term $l(i,j)$ is defined as:

\begin{equation}
l(i,j)=\begin{cases}
\lVert\vect{f}_{i}-\vect{f}_{j}\rVert_{2} & \text{if } I_{ij}=1 \\
\max(0,K-\lVert\vect{f}_{i}-\vect{f}_{j}\rVert_{2}) & \text{if } I_{ij}=0
\end{cases}
\end{equation}

\paragraph{Total Optimization Objective.}
The final objective function is a weighted sum of these components:

\begin{equation}
L_{\mathrm{total}} = \gamma_e L_{\mathrm{exist}} + \gamma_n L_{\mathrm{non\_exist}} + \gamma_o L_{\mathrm{offset}} + \gamma_f L_{\mathrm{feature}}
\end{equation}

\subsection{Data Curation and Augmentation Pipeline}
We developed a proprietary semi-automated data pipeline to solve the lack of labeled data.

\textbf{Manual Annotation:} We manually annotated 5,000 key frames.

\textbf{The Augmentation Engine:} We developed custom software to procedurally generate a dataset of 33,000 frames. The software applied randomized perturbations including \textbf{geometric} (random zooming, cropping, and horizontal flipping), \textbf{photometric} (HSV shifting to simulate different time-of-day lighting), and \textbf{noise injection} (Gaussian noise addition to simulate low-light sensor grain) transformations.

\subsection{Training Strategy}
Training was performed in two distinct stages.

\textbf{Transfer Learning:} Pre-trained for 20 epochs on the TuSimple public dataset~\cite{tusimple}.

\textbf{Domain Adaptation with Manual Scheduling:} Fine-tuned on our custom augmented dataset. We stepped the learning rate down (from $5\times10^{-4}$ to $1\times10^{-8}$) specifically when the loss value dropped below 1.00.

\begin{figure}[H]
    \centering
    \parbox{0.9\textwidth}{\centering
    \includegraphics[width=0.8\textwidth]{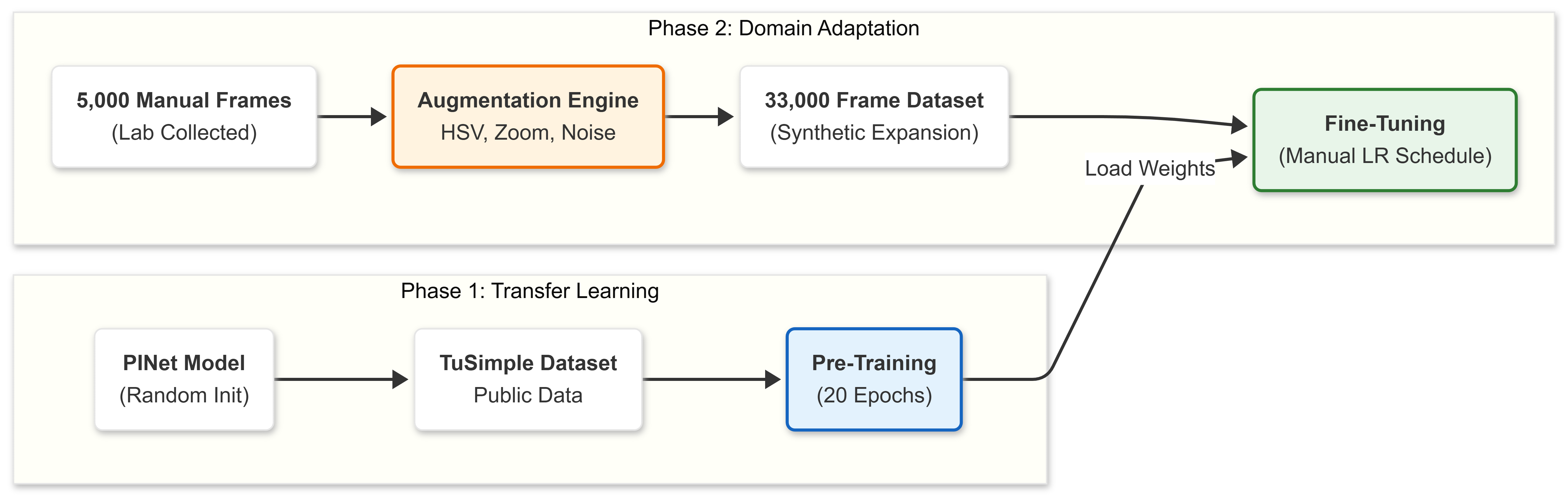}
    }
    \caption{Phase 1 and Phase 2 Training Diagram. A public dataset provided the base feature extraction weights, while aggressive domain adaptation using synthetic expansion tuned the model to the physical constraints of the competition track.}
    \label{fig:training-diagram}
\end{figure}

Fig.~\ref{fig:training-diagram} summarizes the two training stages.

\subsection{Lane Geometry Extraction and Post-Processing}
\label{subsec:postproc}
The raw outputs of the PINet network exist in perspective image space, which are non-linear and unsuitable for direct lateral control. To bridge the gap between perception and path planning, we engineered a deterministic post-processing pipeline that extracts physical state variables: Cross-Track Error (distance), Heading Error (degree), and Curvature.

\subsubsection{Inverse Perspective Mapping (IPM)}
To accurately measure distances and angles, the raw clustered lane points are projected into a Bird's Eye View (BEV) using a highly calibrated Homography matrix ($\bm{H}_{\mathrm{IPM}}$)~\cite{hartley2004}. Let $\vect{p}_i = [u, v, 1]^{\mathsf{T}}$ represent a point in image space. The corresponding physical ground plane point $\vect{p}_w = [x, y, 1]^{\mathsf{T}}$ is obtained via:
\begin{equation}
\vect{p}_w = \bm{H}_{\mathrm{IPM}} \cdot \vect{p}_i
\end{equation}
By executing this affine transformation \textit{after} network inference rather than before, we preserve the far-field visual horizon for the neural network while achieving geometrically accurate measurements in the near-field for the controller. The pipeline in fact maintains two parallel realizations of this transform: a full-frame $\texttt{cv2.warpPerspective}$ used only when a rasterized BEV image is required for visualization or the debug stream, and a lightweight point-domain realization that applies the homogeneous matrix product directly to the sparse cluster centroids. The latter avoids the cost of interpolated image resampling entirely, which matters when the transform is evaluated every frame at $30$ FPS on embedded hardware.

\subsubsection{Polynomial Path Fitting}
Filtered BEV points are isolated into left and right lane boundaries. We model the lane geometry by fitting a second-degree polynomial to the $x$ and $y$ coordinates of the lane boundaries:
\begin{equation}
x(y) = a_2 y^2 + a_1 y + a_0
\label{eq:poly}
\end{equation}
If both lane lines are visible, the algorithm computes a dynamic centerline trajectory by averaging the polynomial coefficients: $a_{\mathrm{mid}} = \frac{a_{\mathrm{left}} + a_{\mathrm{right}}}{2}$.

\subsubsection{Control Variable Generation}
With the centerline modeled mathematically, critical control variables are extracted dynamically relative to a look-ahead point ($y_{\mathrm{ref}}$). The \textbf{Heading Error ($\theta$)} is computed from the derivative of the polynomial:
\begin{equation} \theta = \arctan(2a_2 y_{\mathrm{ref}} + a_1) \label{eq:heading} \end{equation}
The \textbf{Curvature ($\kappa$)} is extracted using the standard formula for the radius of curvature:
\begin{equation} \kappa = \frac{|2a_2|}{(1 + (2a_2 y_{\mathrm{ref}} + a_1)^2)^{3/2}} \end{equation}
To maximize stability, the final reported Cross-Track Error and Heading Error are not sampled from a single look-ahead point, but rather derived from a weighted spatial average spanning $y \in [220, 300]$, heavily penalizing variations close to the vehicle chassis.

\subsubsection{Embedding-Guided Instance Clustering}
Raw detections emerging from the confidence, offset, and embedding heads arrive as an unordered point cloud; before any geometric fitting can occur, points must first be grouped into discrete lane instances. Rather than committing to a fixed-count clustering algorithm such as $k$-means, we implemented a greedy, similarity-threshold clustering scheme that mirrors the training-time discriminative loss (Eq.~\ref{eq:feature}) directly at inference time: each candidate point is compared, in embedding space, against every cluster centroid formed so far, and is merged into the nearest cluster whenever the $L_2$ distance falls under a fixed affinity threshold $\tau_{\mathrm{inst}}$; otherwise it seeds a new cluster, bounded to a maximum of 12 concurrent instances to guarantee bounded worst-case runtime. Cluster centroids are updated incrementally as a running mean rather than recomputed from scratch, keeping the per-frame cost linear in the number of detected points.

\begin{breakablealgorithm}
\caption{Greedy Embedding-Affinity Clustering}
\label{alg:clustering}
\begin{algorithmic}[1]
\Require Points $\{(x_i, y_i)\}$ passing the confidence threshold; embeddings $\{\vect{f}_i\}$; affinity threshold $\tau_{\mathrm{inst}}$; instance cap $K_{\max}=12$
\Ensure Instance-labeled point sets $\{X_k, Y_k\}$
\State $\mathcal{C} \gets \emptyset$
\For{each point $i$}
    \If{$\mathcal{C} = \emptyset$}
        \State create a new cluster with centroid $\bar{\vect{f}} \gets \vect{f}_i$
    \Else
        \State $k^\ast \gets \displaystyle\arg\min_{k \in \mathcal{C}} \lVert \vect{f}_i - \bar{\vect{f}}_k \rVert_2$
        \If{$\lVert \vect{f}_i - \bar{\vect{f}}_{k^\ast} \rVert_2 \le \tau_{\mathrm{inst}}$}
            \State assign point $i$ to cluster $k^\ast$
            \State $\bar{\vect{f}}_{k^\ast} \gets \dfrac{\bar{\vect{f}}_{k^\ast}\cdot n_{k^\ast} + \vect{f}_i}{n_{k^\ast}+1}$ \Comment{incremental running mean}
        \ElsIf{$|\mathcal{C}| < K_{\max}$}
            \State create a new cluster with centroid $\bar{\vect{f}} \gets \vect{f}_i$
        \EndIf
    \EndIf
\EndFor
\end{algorithmic}
\end{breakablealgorithm}

Two further filters run around this clustering stage. First, any cluster retaining three points or fewer is discarded as sensor noise prior to fitting, since a two-point set cannot support a statistically meaningful regression. Second, a region-of-interest mask, enforced via a calibrated polygon test against the drivable corridor, rejects points that fall outside the physically relevant track area both before and after clustering --- suppressing off-track clutter and adjacent-lane bleed-through at the image periphery. A dedicated far-field suppression rule additionally discards any cluster whose points lie overwhelmingly ($\geq 90\%$) above a fixed BEV row threshold, since such clusters typically correspond to unstable horizon-line artifacts rather than the near-field lane segment relevant to control.

\subsubsection{Adaptive-Order Fitting and Nearest-Boundary Selection}
Rather than committing every detected cluster to the fixed second-degree model of Eq.~\ref{eq:poly}, the pipeline selects the polynomial order according to point count: sparse clusters (6--16 points) receive a first-degree (linear) fit, while well-populated clusters ($>16$ points) receive the full second-degree fit. This prevents an under-constrained quadratic from chasing detector noise on short, partially-occluded lane segments --- a frequent occurrence on a 1:10 scale track where motion blur and occlusion routinely truncate the visible lane length.

Because the network can, and does, emit several simultaneous line-cluster candidates per frame --- occluded segments, opposing-straightaway bleed-through on hairpin exits, or spurious far-field fragments that survive the filters above --- the pipeline does not naively take the two most confident clusters as ``left'' and ``right.'' Instead, every surviving fit is evaluated at the vehicle's projected center row, and the pipeline independently selects the fit whose lateral position is nearest to the center line on the left and on the right, discarding all remaining candidates as scene clutter. This center-adjacency heuristic is what keeps the controller stable even in frames where the network reports four or five simultaneous line candidates.

When only one boundary survives this selection --- for example, at a hairpin apex where the outer line briefly leaves the field of view --- the centerline is reconstructed by offsetting the single available fit using a lane-width estimate. This width is tracked dynamically from the most recent frames in which both boundaries were visible, clamped to a physically plausible band of $60$--$100$ px in BEV space, and falls back to a fixed default of $70$ px whenever the running estimate leaves this band. This graceful-degradation path is what prevents a momentary single-line detection from producing a discontinuous jump in the reported centerline.

Finally, the pixel-domain lateral offset (Eq.~\ref{eq:poly}) and heading angle (Eq.~\ref{eq:heading}) are not physically meaningful on their own: both are converted to real-world units by calibration constants obtained from the same homography used for the IPM warp in section~\ref{subsec:postproc}, before being packed and transmitted over the $30$~Hz ROS control topic served by the shared-memory bus described in section~\ref{subsec:ipc}.

\begin{figure}[H]
    \centering
    \includegraphics[width=0.28\linewidth]{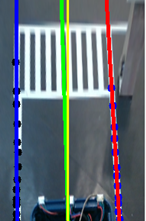}
    \caption{Bird's Eye View output of the post-processing pipeline. Clustered lane points are rectified via the homography $\bm{H}_{\mathrm{IPM}}$ and fitted with adaptive-order polynomials: left boundary (blue), right boundary (red, dashed), and the synthesized centerline (green). The near-vertical, parallel boundaries validate the IPM calibration.}
    \label{fig:bev_output}
\end{figure}

Fig.~\ref{fig:bev_output} shows a representative output of the full post-processing chain. After Inverse Perspective Mapping rectifies the perspective distortion, the greedy embedding-affinity clustering (Alg.~\ref{alg:clustering}) resolves the scene into discrete instances, and the adaptive-order polynomial fitter produces the three curves used by the controller: the left boundary (blue), the right boundary (red, dashed), and the synthesized centerline (green) computed by averaging the boundary coefficients. In the rectified BEV domain the physically parallel lane boundaries appear vertical and near-parallel, confirming that the homography is well calibrated; residual convergence at the top of the frame is the expected far-field noise that the weighted spatial averaging over $y \in [220, 300]$ is designed to attenuate.

\subsection{Traffic Sign and Signal Perception}
\label{subsec:signs}
Due to the scarcity of high-quality annotated data for 1:10 scale traffic signs within the target environment, relying strictly on single-frame neural network classification yielded unacceptable variance (flickering). To guarantee robust perception, we implemented a hybrid depth-gated approach coupled with an aggressive temporal smoothing architecture.

\paragraph{Depth-Gated Validation.}
The RealSense D455 provides aligned depth and RGB maps. Instead of relying purely on 2D bounding box dimensions to estimate sign proximity, we sample the raw depth tensor (scaled by $0.001$ meters/unit) at the centroid of the detected sign. Sign bounding boxes reporting distances beyond a physical limit of $0.85$ meters are immediately rejected. This depth-gating prevents the vehicle from reacting prematurely to signs in adjacent lanes or far down the track.

\paragraph{Temporal Voting Filter (FIFO Buffer).}
To overcome classification instability from the sparse training data, we treat the detection stream as a noisy time-series signal. We engineered a sliding-window First-In-First-Out (FIFO) buffer of length $N=5$. 

\begin{figure}[H]
    \centering
    \begin{tikzpicture}[
        >={Stealth[scale=1.2]},
        node distance=12mm and 14mm,
        box/.style={draw, rectangle, minimum height=1.3cm, align=center, rounded corners=2pt, font=\small, text width=2.3cm},
        buffer/.style={draw, rectangle, minimum height=1cm, minimum width=0.85cm, fill=white, font=\footnotesize}
    ]

    \node[box, fill=red!10] (input) {Raw Detections\\(Noisy)};
    \node[buffer, right=18mm of input] (b1) {$C_t$};
    \node[buffer, right=0mm of b1] (b2) {$C_{t-1}$};
    \node[buffer, right=0mm of b2] (b3) {$\dots$};
    \node[buffer, right=0mm of b3] (b4) {$\dots$};
    \node[buffer, right=0mm of b4] (b5) {$C_{t-4}$};
    \node[box, fill=orange!10, right=18mm of b5] (vote) {Majority\\Voting\\Filter};
    \node[box, fill=green!10, right=of vote] (output) {Stable Output\\Classification};

    \begin{scope}[on background layer]
        \node[draw, rectangle, rounded corners=2pt, fill=gray!10, fit=(b1)(b2)(b3)(b4)(b5), inner sep=3mm] (fifo_wrap) {};
    \end{scope}
    \node[above=1mm of fifo_wrap, font=\small] {FIFO Time-Series Buffer ($N=5$)};

    \draw[->, thick] (input) -- (fifo_wrap.west);
    \draw[->, thick] (fifo_wrap.east) -- (vote);
    \draw[->, thick] (vote) -- (output);

    \end{tikzpicture}
    \caption{The FIFO Temporal Filter mitigates single-frame detection errors, translating noisy, data-starved classifier outputs into robust, state-machine-ready signals.}
    \label{fig:fifo-filter}
\end{figure}
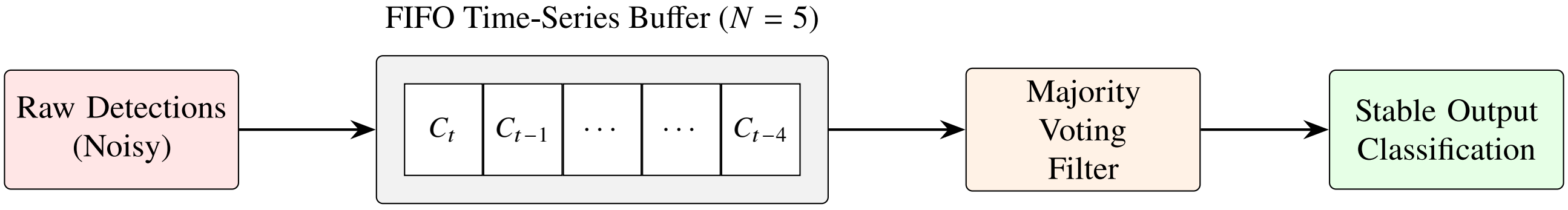

Let $C_t$ be the raw classification at time $t$. The final published classification $O_t$ is dictated by a strict majority voting threshold:
\begin{equation}
O_t = 
\begin{cases} 
C_i & \text{if } \sum_{k=0}^{N-1} \delta(B[k], C_i) > \lfloor N/2 \rfloor \\
17 \text{ (Null)} & \text{otherwise}
\end{cases}
\end{equation}
This hysteresis loop guarantees that temporary occlusions or motion blur do not trigger an erroneous state change, dramatically smoothing the behavioral transitions of the vehicle. Fig.~\ref{fig:fifo-filter} illustrates the filter.

\subsection{Crosswalk Detection}
\label{subsec:crosswalk}
Pedestrian crosswalks require a fundamentally different treatment from the parametric lane model. A crosswalk is not a continuous boundary to be tracked but a discrete, safety-critical region that must trigger a deterministic stop-and-yield behavior. We therefore run crosswalk detection as an independent perception path rather than folding it into the PINet lane head, so that a crosswalk can be flagged even in frames where the lane geometry is momentarily ambiguous.

\begin{figure}[H]
    \centering
    \includegraphics[width=0.55\linewidth]{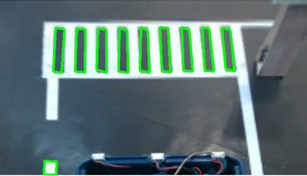}
    \caption{Crosswalk detection in the rectified BEV frame. Each validated stripe of the repeating ladder pattern is bounded by a green box; the isolated marking at the bottom of the frame is correctly excluded by the group-consistency test, preventing a single stripe or floor reflection from triggering a false stop.}
    \label{fig:crosswalk_detect}
\end{figure}

\paragraph{Detection and Geometric Validation.}
The detector operates on the rectified Bird's Eye View, where the characteristic ladder pattern of a crosswalk becomes a set of near-rectangular, regularly spaced stripes of consistent orientation and area. Candidate stripes are extracted and then subjected to a geometric consistency test: an isolated bright rectangle is rejected, whereas a group of stripes sharing a common orientation, comparable area, and uniform inter-stripe spacing is accepted as a crosswalk. This structural voting scheme is deliberately conservative---demanding a repeating pattern rather than a single marking---so that the isolated dashes of a lane boundary or a specular reflection on the floor cannot masquerade as a crossing. Fig.~\ref{fig:crosswalk_detect} shows a representative detection, in which each validated stripe is bounded by a green box in the BEV frame.

\paragraph{Behavioral Signaling.}
Upon a validated detection, the crosswalk node updates a boolean status flag in shared memory and immediately notifies the control node, which allows the vehicle to execute a controlled deceleration before the marked region rather than reacting only once the stripes are directly beneath the chassis. Because the crosswalk flag is transmitted over the same zero-copy shared-memory channel described in section~\ref{subsec:ipc}, this safety-critical signal incurs no serialization latency and remains synchronized with the lane and sign outputs on a per-frame basis.

\subsection{High-Throughput IPC Architecture}
\label{subsec:ipc}
A critical bottleneck in complex ROS 1 environments is the latency and CPU overhead caused by TCP/IP serialization of large image payloads. To resolve this, we bypassed the standard \texttt{sensor\_msgs/Image} pipelines. Both the Lane Perception and Sign Perception nodes access raw $480\times848\times3$ camera tensors directly via POSIX Shared Memory (\texttt{multiprocessing.shared\_memory}). This zero-copy Inter-Process Communication (IPC) strategy reduces frame distribution latency to sub-millisecond levels, ensuring the entire perception stack operates synchronously at $30$ FPS.

\subsection{Experimental Results and Validation}

\subsubsection{Training Convergence Analysis}
The model's learning trajectory validates our two-stage training strategy.

\textbf{Rapid Adaptation (Epochs 0-8):} Sharp initial decline in loss using pre-trained weights.

\textbf{Optimal Performance Window (Epochs 8-21):} Loss minimized to its lowest sustainable point. Our custom dataset prevented overfitting.

\textbf{Stability:} Post-epoch 21, the manual learning rate scheduler prevented ``exploding gradient'' issues.

\begin{figure}[H]
    \centering
    \parbox{0.8\textwidth}{\centering
    \includegraphics[width=0.8\textwidth]{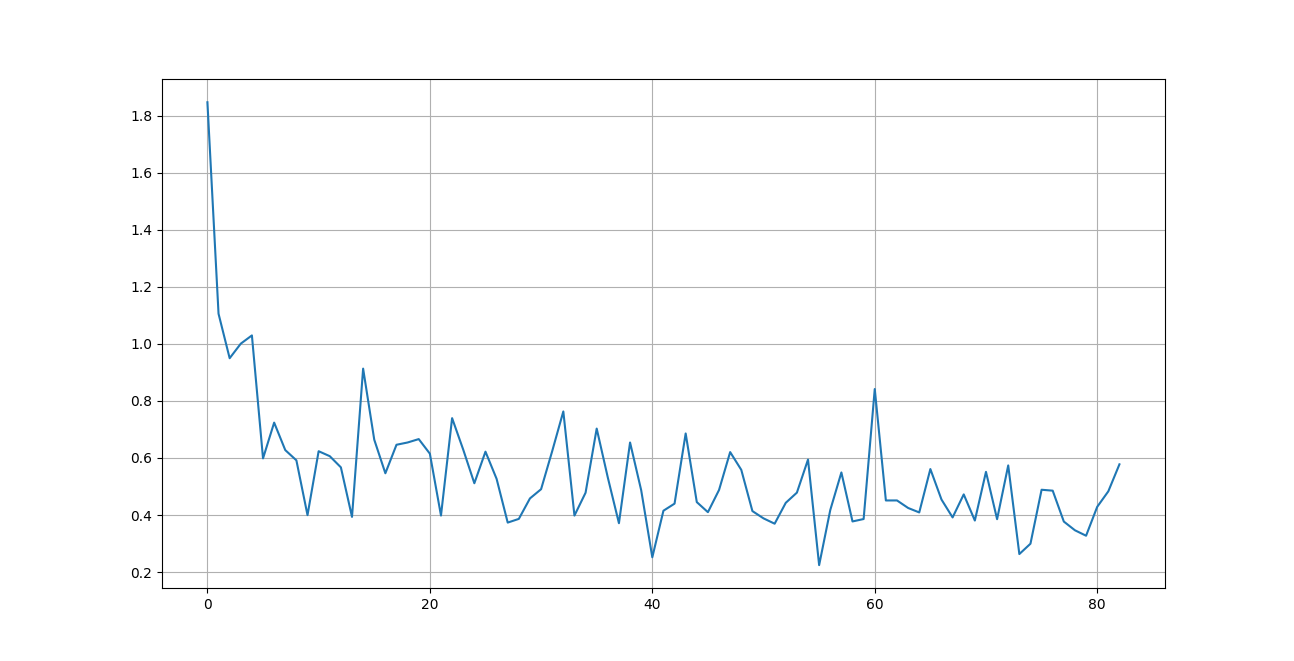}
    }
    \caption{Training loss over epochs. The period highlighted between Epochs 8 and 21 demonstrates the model achieving maximum accuracy without succumbing to overfitting constraints.}
    \label{fig:training-loss}
\end{figure}

Fig.~\ref{fig:training-loss} shows the training loss over epochs.

\subsubsection{Qualitative Domain Adaptation}
\begin{figure}[H]
    \centering
    \begin{subfigure}[b]{0.48\textwidth}
        \centering
        \parbox{0.9\textwidth}{\centering
            \includegraphics[width=0.8\textwidth]{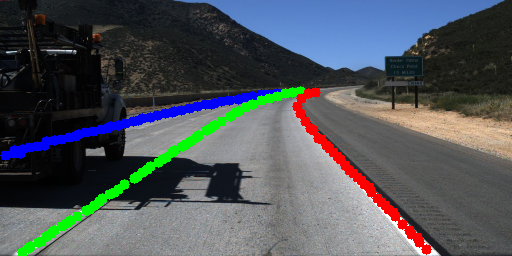}
        }
        \caption{TuSimple Dataset}
        \label{fig:tusimple}
    \end{subfigure}
    \hfill
    \begin{subfigure}[b]{0.48\textwidth}
        \centering
        \parbox{0.9\textwidth}{\centering
            \includegraphics[width=0.8\textwidth]{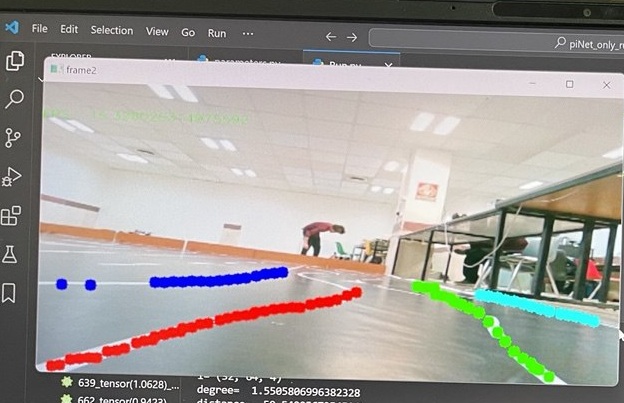}
        }
        \caption{Custom Lab Track}
        \label{fig:labtrack}
    \end{subfigure}
    
    \caption{Domain Adaptation Results. (a) Successful instance segmentation on the TuSimple pre-training dataset. (b) Successful transfer to the custom competition environment, handling distinct texture and lighting changes dynamically without ROI cropping.}
    \label{fig:domain_adaptation}
\end{figure}

Fig.~\ref{fig:tusimple} shows the model inference on the public TuSimple dataset. Fig.~\ref{fig:labtrack} demonstrates the model operating in our custom lab environment. Despite differences in lighting and texture, the network successfully generalized its lane instance extraction capabilities. The most demanding test of this transfer is not the dataset the network was pre-trained on, but the raw competition track under laboratory lighting, where reflective flooring, hard shadows, and the vehicle's own chassis occlusion are all present in a single frame.

\subsubsection{End-to-End Lab Inference and Instance Separation}
\begin{figure}[H]
    \centering
    \includegraphics[width=0.6\linewidth]{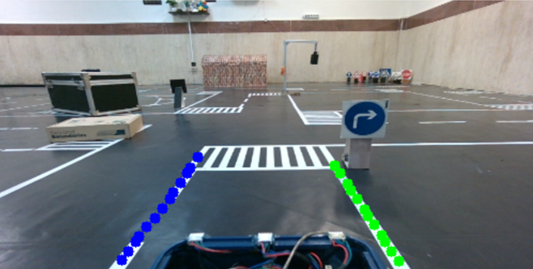}
    \caption{PINet output in the lab environment. Clustered key points are overlaid on a raw competition-track frame, with the two lane boundaries separated into distinct instances (blue: left, green: right) by the embedding-guided clustering stage. Foreground reflections and the chassis occlusion are correctly rejected.}
    \label{fig:pinet_output}
\end{figure}

The clearest evidence that the trained network operates correctly on the physical platform is its in-lab output on live frames. Fig.~\ref{fig:pinet_output} overlays the network's key point predictions---after the greedy embedding-affinity clustering of Algorithm~\ref{alg:clustering}---directly onto a raw track frame captured in the lab. The two lane boundaries are cleanly resolved into distinct instances, rendered here in blue (left boundary) and green (right boundary). Critically, the discriminative embedding loss (Eq.~\ref{eq:feature}) has done exactly what it was trained to do: even though the two point sets converge toward a common vanishing region near the crosswalk, the network assigns them to separate instances rather than merging them, and it correctly suppresses spurious detections on the reflective foreground and the static chassis in the lower frame. Read together with the domain-adaptation panels of Fig.~\ref{fig:domain_adaptation}, this confirms that both the feature extractor and the instance-separation head transferred successfully from the public pre-training data to the competition environment.

This image-space clustered output is precisely the input consumed by the post-processing pipeline of section~\ref{subsec:postproc}: the same blue/green instances, once passed through Inverse Perspective Mapping and adaptive-order polynomial fitting, yield the rectified Bird's Eye View trajectory of Fig.~\ref{fig:bev_output}. The two figures therefore represent consecutive stages of a single frame's journey from raw pixels to controller-ready state variables.

\section{Control and Localization}
\label{sec:control}

Accurate localization is essential for enabling the autonomous vehicle to determine its position, orientation, and motion relative to the environment. In this system, vehicle dynamics are modeled using the Ackermann steering mechanism and the standard Bicycle Kinematic Model~\cite{rajamani2012}, which provides a mathematically tractable representation of motion for ground vehicles with small steering angles. This model is used both for localization estimates and as a foundation for motion planning and feedback-based correction.

\subsection{Vehicle Kinematic Modeling}
\label{subsec:kinematics}

The bicycle model treats the vehicle as a single-track system with a front and rear wheel aligned along a common centerline. The steering input is applied at the front wheel, and the resulting motion is governed by geometric constraints imposed by the fixed wheelbase.

\begin{figure}[H]
  \centering
  \includegraphics[width=0.7\linewidth]{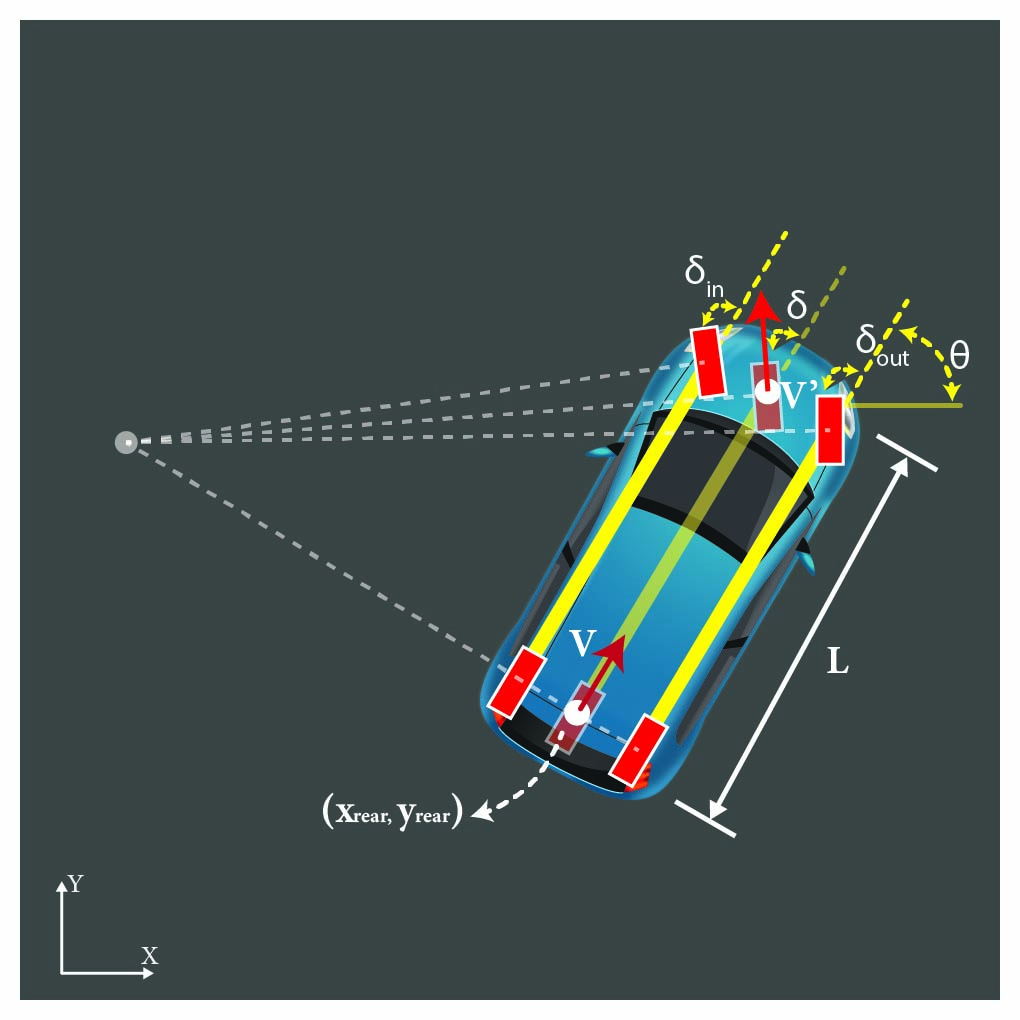}
  \caption{Bicycle model representation of the vehicle kinematics.}
  \label{fig:bicycle_model}
\end{figure}

Fig.~\ref{fig:bicycle_model} illustrates the kinematic abstraction of the vehicle used in this project.

Following the Ackermann steering relationship, the steering geometry for the equivalent bicycle model may be expressed as:
\begin{equation}
\cot(\delta) = \frac{\cot(\delta_{\mathrm{in}}) + \cot(\delta_{\mathrm{out}})}{2}
\label{eq:nonlin_steering_equation}
\end{equation}

Since $R = L/\tan\delta$ is the turning radius of the rear-axle centre, this is the mean of the two wheel radii measured on the rear-axle line, and it is exact when the wheel axes meet at a common point on that line.

Given the front steering angle $\delta$ and linear velocity $V$, the kinematic state of the vehicle is represented by three parameters: $x$ (longitudinal position), $y$ (lateral position), and $\theta$ (vehicle heading).

Depending on the reference point used (front axle or rear axle), the equations of motion are expressed as follows:

\paragraph{Front-Axle Model}
\begin{align}
\dot{x}_{\mathrm{front}} &= V' \cos(\theta - \delta) \\
\dot{y}_{\mathrm{front}} &= V' \sin(\theta - \delta) \\
\dot{\theta} &= \frac{V'}{L} \sin(\delta)
\end{align}

\paragraph{Rear-Axle Model}
\begin{align}
\dot{x}_{\mathrm{rear}} &= V \cos(\theta) \\
\dot{y}_{\mathrm{rear}} &= V \sin(\theta) \\
\dot{\theta} &= \frac{V}{L} \tan(\delta)
\end{align}

Here, $L$ denotes the vehicle wheelbase, $V$ is the linear velocity of the equivalent rear wheel (at the rear-axle center), $V'$ is the linear velocity of the equivalent front wheel (at the front-axle centre), $\theta$ is the vehicle orientation, and $\delta$ is the front steering angle. These expressions provide a motion-level representation for predicting the vehicle's pose based on its current control inputs.

\subsection{Steering-Limited Path Design}
\label{subsec:steering_limit}

Due to the mechanical constraints of the Ackermann steering layout, the vehicle cannot arbitrarily translate between arbitrary pose configurations with a single steering command. Instead, its feasible motion is limited by the wheelbase \(L\) and maximum steering angle \(\delta_{\mathrm{max}}\). As a result, transitioning between two configurations typically requires a sequence of bounded curvature maneuvers~\cite{dubins1957}.

\begin{figure}[H]
  \centering
  \includegraphics[width=0.65\linewidth]{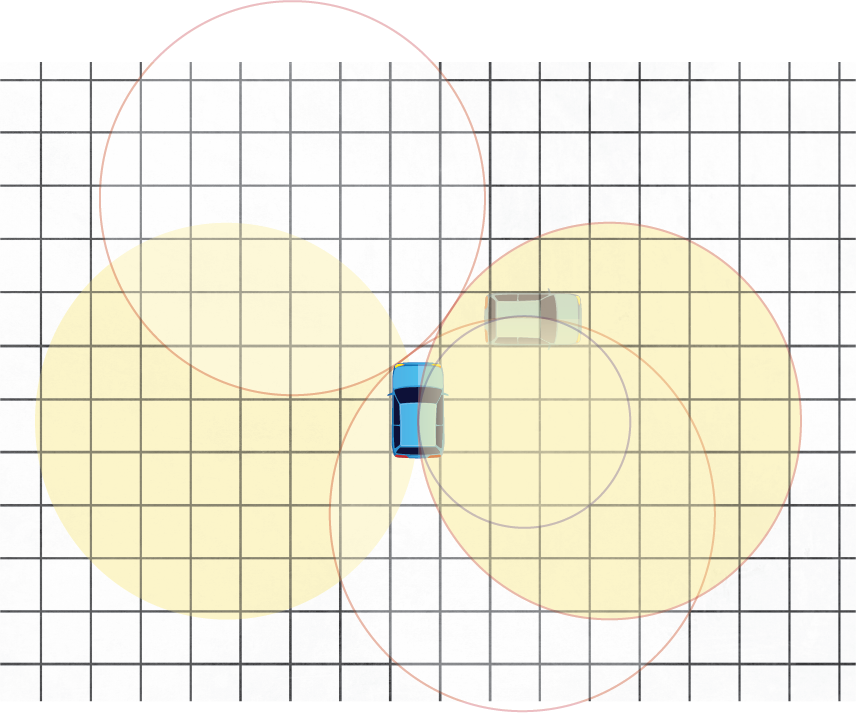}
  \caption{Vehicle relocation between two poses using up to three steering segments with maximum steering angle.}
  \label{fig:three_turn}
\end{figure}

Fig.~\ref{fig:three_turn} illustrates an example of repositioning the vehicle using up to three maximum-steering arcs, a common strategy when a single steering motion is insufficient.

The minimum turning radius imposed by the vehicle geometry is:
\begin{equation}
R_{\mathrm{min}} = \frac{L}{\tan(\delta_{\mathrm{max}})}
\label{eq:rmin}
\end{equation}

To represent motion between an initial pose \((x_i, y_i, \theta_i)\) and target pose \((x_j, y_j, \theta_j)\), the path may be decomposed into three sequential curvature segments of angular change \(\Delta\varphi_1\), \(\Delta\varphi_2\), and \(\Delta\varphi_3\), where:
\begin{equation}
\theta_i - \theta_j = \Delta\varphi_1 + \Delta\varphi_2 + \Delta\varphi_3
\end{equation}

Let the cumulative displacement vector be:
\begin{equation}
\vect{r} = 
\begin{bmatrix}
\Delta x \\[4pt]
\Delta y
\end{bmatrix}
=
R_{\mathrm{min}} 
\bigg(
\bm{M}_1 
\begin{bmatrix}
    \sin(\Delta\varphi_1) \\
\cos(\Delta\varphi_1) - 1 
\end{bmatrix}
-
\bm{M}_2 
\begin{bmatrix}
    \sin(\Delta\varphi_2) \\
\cos(\Delta\varphi_2) - 1 
\end{bmatrix}
+
\bm{M}_3 
\begin{bmatrix}
    \sin(\Delta\varphi_3)\\
    \cos(\Delta\varphi_3) - 1 
\end{bmatrix}
\bigg)
\label{eq:path_displacement}
\end{equation}

where \(\bm{M}_1\), \(\bm{M}_2\), and \(\bm{M}_3\) are rotation matrices corresponding to the vehicle orientation at the start of each segment:

\begin{equation}
\bm{M}_1 =
\begin{bmatrix}
    \cos(\theta_i) & -\sin(\theta_i) \\
     \sin(\theta_i) & \cos(\theta_i) 
\end{bmatrix}
\end{equation}

\begin{equation}
\bm{M}_2 =
\begin{bmatrix}
    \cos(\theta_i - \Delta\varphi_1) & -\sin(\theta_i - \Delta\varphi_1) \\
     \sin(\theta_i - \Delta\varphi_1) & \cos(\theta_i - \Delta\varphi_1)
\end{bmatrix}
\end{equation}

\begin{equation}
\bm{M}_3 =
\begin{bmatrix}
    \cos(\theta_i - \Delta\varphi_1 - \Delta\varphi_2) & -\sin(\theta_i - \Delta\varphi_1 - \Delta\varphi_2) \\ 
    \sin(\theta_i - \Delta\varphi_1 - \Delta\varphi_2) & \cos(\theta_i - \Delta\varphi_1 - \Delta\varphi_2)
\end{bmatrix}
\end{equation}

\begin{figure}[H]
  \centering
  \includegraphics[width=0.68\linewidth]{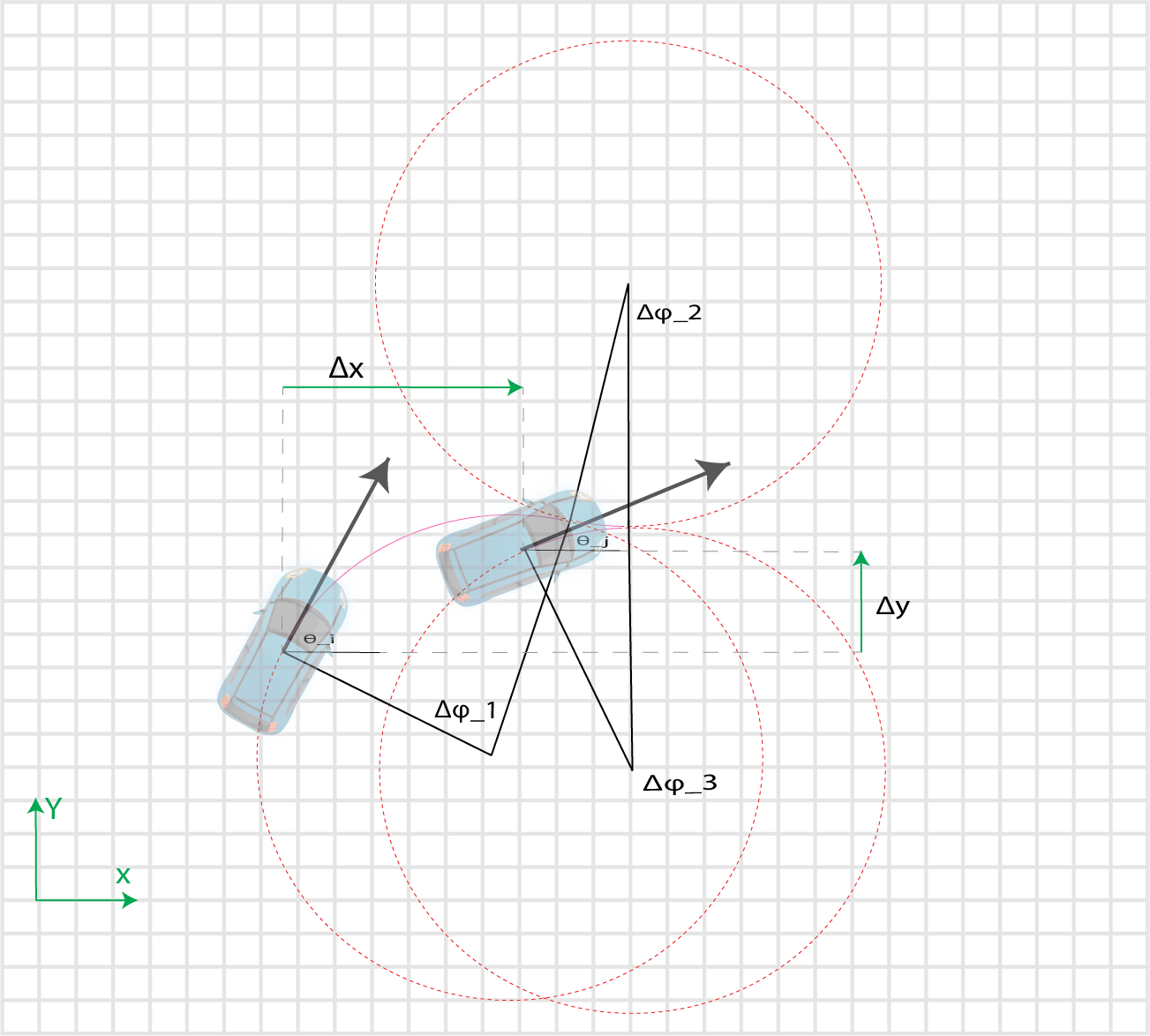}
  \caption{Geometric and mathematical representation of three-segment steering motion.}
  \label{fig:path_math}
\end{figure}

Fig.~\ref{fig:path_math} provides a geometric representation of the three-segment motion formulation.

\begin{figure}[H]
    \centering
    \includegraphics[width=0.9\linewidth]{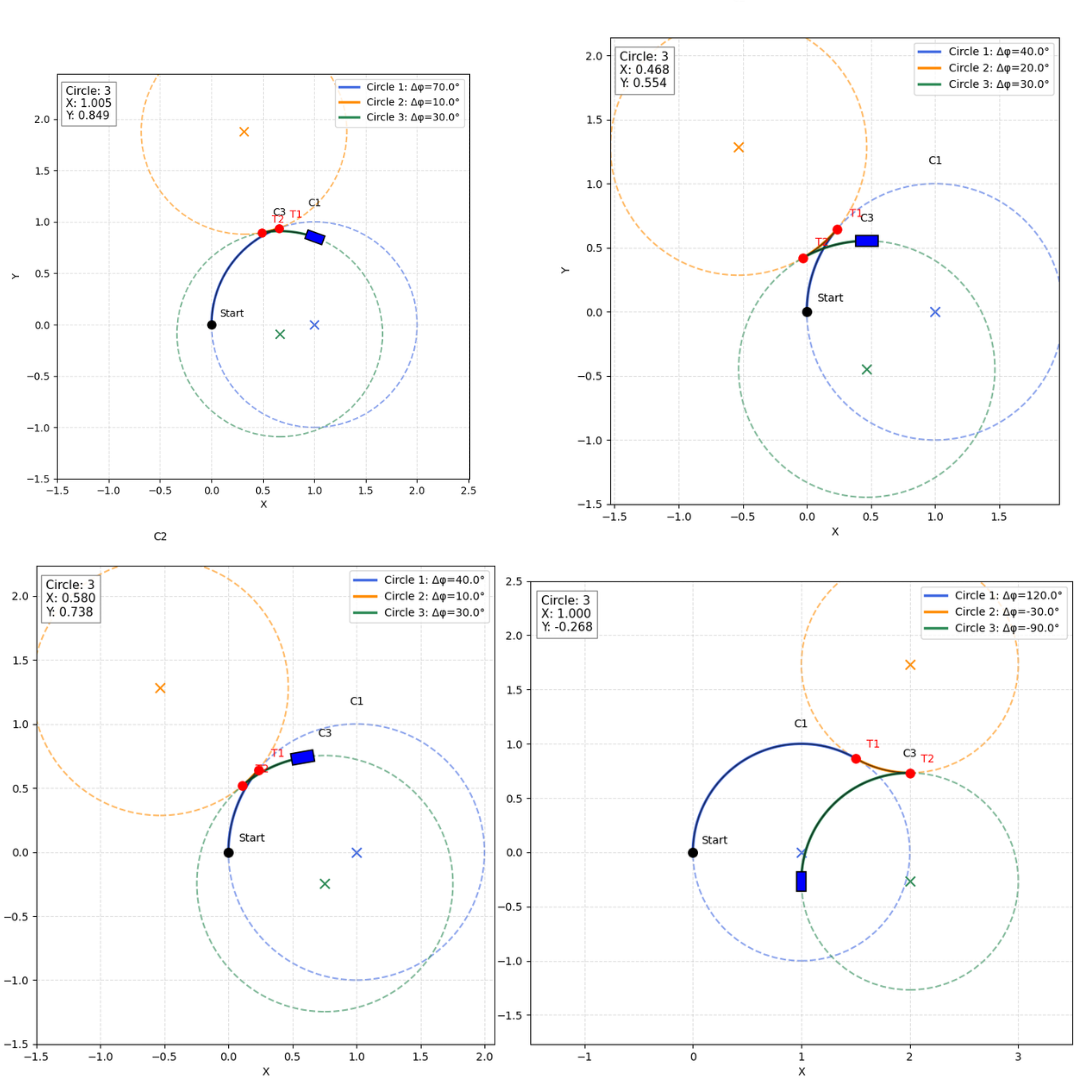}
    \caption{Example three-segment steering maneuvers (tangent circular arcs) for different $\Delta\varphi_1$, $\Delta\varphi_2$, $\Delta\varphi_3$ combinations, illustrating repositioning into regions unreachable by a single steering command.}
    \label{fig:AV_three_circle_demo}
  \end{figure}

This three-segment maneuver is specifically required for repositioning the vehicle into a region on its right side that cannot be reached through a single ordinary steering command, due to the geometric constraints imposed by the vehicle's minimum turning radius. Repositioning into the mirror-symmetric region on the left side does not require re-deriving the maneuver: it suffices to mirror the target displacement about the vehicle's longitudinal axis, solve Eq.~\ref{eq:path_displacement} for that mirrored displacement to obtain $\Delta\varphi_1$, $\Delta\varphi_2$, $\Delta\varphi_3$, and then negate all three resulting angular changes (i.e., use $-\Delta\varphi_1$, $-\Delta\varphi_2$, $-\Delta\varphi_3$) to obtain the angular changes that actually produce the desired displacement into the left-side region. See Fig.~\ref{fig:AV_three_circle_demo}

A special case frequently encountered in structured environments (e.g., four-way intersections) is the 90° right turn with constant curvature. In this maneuver:
\begin{equation}
|\Delta\varphi_1| = |\Delta\varphi_3|
\end{equation}

This symmetry simplifies analytic evaluation and enables computation of the turn using closed-form expressions for geometry and displacement. In practical use, this model is executed with steering saturated at \(\delta_{\mathrm{max}}\) and verified using gyroscope feedback.

\textbf{Asymmetric steering limits.} Eq.~\ref{eq:path_displacement} uses the same minimum radius $R_\text{min}$ for all three arcs.  
This is appropriate when the vehicle moves according to the bicycle model with Ackermann-type steering and the maximum steering angle is the same in both directions.  
If the maximum wheel angle differs between turning to the right (positive limit) and turning to the left (negative limit), the minimum radius $R_\text{min} = L/\tan|\delta_\text{max}|$ takes a different value for each direction, $R^{\text{pos}}_{\min,w}$ and $R^{\text{neg}}_{\min,w}$, where $w=r,l$ denotes the reference wheel: the right wheel for repositioning into a region on the right side of the vehicle, and the left wheel for repositioning into a region on the left side of the vehicle.  
Two cases are distinguished. 

\textbf{Case 1:} The difference is not negligible.  
Each arc uses the radius of its own steering direction: 
\begin{equation}
\begin{bmatrix}\Delta x\\ \Delta y\end{bmatrix}
= R^{\mathrm{pos}}_{\min,r}\bm M_1\begin{bmatrix}\sin(\Delta\varphi_1)\\ \cos(\Delta\varphi_1)-1\end{bmatrix}
- R^{\mathrm{neg}}_{\min,r}\bm M_2\begin{bmatrix}\sin(\Delta\varphi_2)\\ \cos(\Delta\varphi_2)-1\end{bmatrix}
+ R^{\mathrm{pos}}_{\min,r}\bm M_3\begin{bmatrix}\sin(\Delta\varphi_3)\\ \cos(\Delta\varphi_3)-1\end{bmatrix}
\label{eq:case_1_asymmetry}
\end{equation} 
For a region on the left side of the vehicle, the corresponding relation follows the same mirroring procedure described earlier: mirror the target displacement about the vehicle's longitudinal axis, solve as in Eq.~\ref{eq:case_1_asymmetry}, then negate the resulting $\Delta\varphi_1$, $\Delta\varphi_2$, $\Delta\varphi_3$, only now using the left-side radii $R^{\mathrm{neg}}_{\min,l}$, $R^{\mathrm{pos}}_{\min,l}$, $R^{\mathrm{neg}}_{\min,l}$ in place of $R^{\mathrm{pos}}_{\min,r}$, $R^{\mathrm{neg}}_{\min,r}$, $R^{\mathrm{pos}}_{\min,r}$. 

\textbf{Case 2:} The difference is small enough that the two radii can be averaged.  
\begin{equation}
\bar{R}_{\min,r}=\frac{R^{\mathrm{pos}}_{\min,r}+R^{\mathrm{neg}}_{\min,r}}{2}, \qquad \bar{R}_{\min,l}=\frac{R^{\mathrm{pos}}_{\min,l}+R^{\mathrm{neg}}_{\min,l}}{2}
\end{equation} 
For a region on the right side of the vehicle, Eq.~\ref{eq:path_displacement} is retained with $R_\text{min}$ replaced by $\bar{R}_{\min,r}$.  
For a region on the left side of the vehicle, following the same mirroring procedure described earlier, Eq.~\ref{eq:path_displacement} is retained with $R_\text{min}$ replaced by $\bar{R}_{\min,l}$. 

\subsection{Navigation and Steering Control Algorithms}
\label{subsec:navigation_control}

To maintain stable, lane-centered navigation under varying road topologies and speeds, multiple feedback controllers~\cite{paden2016} are implemented in the vehicle's decision and actuation loop. Each controller is selected for its specific performance characteristics and switching between them is performed according to the vehicle state and operational mode.

\begin{figure}[H]
  \centering
  \includegraphics[width=0.70\linewidth]{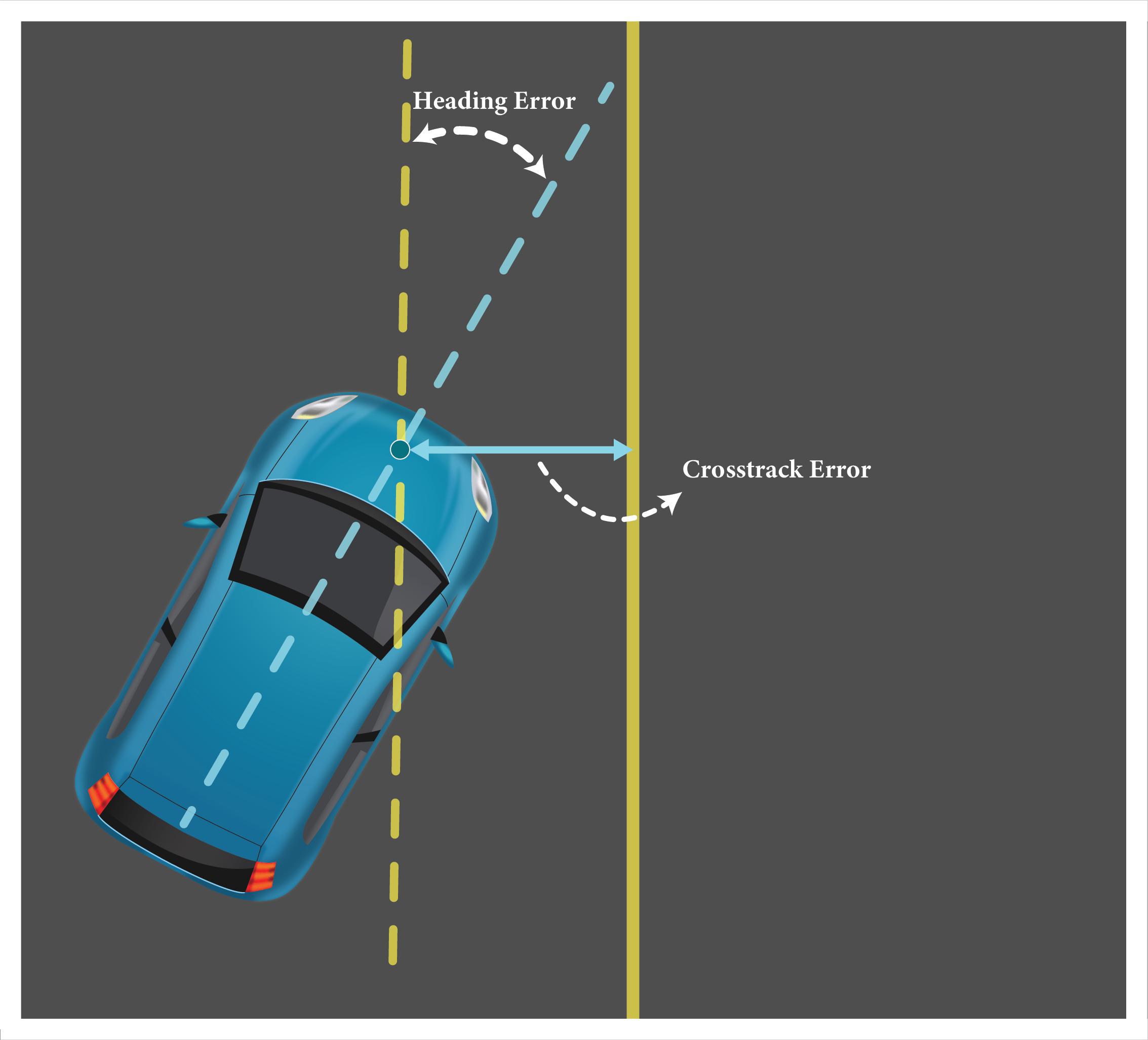}
  \caption{Geometric interpretation of cross-track and heading errors used in lane-keeping control.}
  \label{fig:ct_he}
\end{figure}

Fig.~\ref{fig:ct_he} illustrates the two primary geometric feedback quantities used in lane-keeping control: the \textbf{Cross-Track Error ($\mathrm{CE}$)}, the lateral deviation of the vehicle center from the lane reference centerline, and the \textbf{Heading Error ($\mathrm{HE}$)}, the angular misalignment between the vehicle's orientation and the tangent direction of the lane.

\paragraph{PID Controller.}

A PID controller~\cite{astrom2006} is used primarily as a backup steering mechanism and for regulating the forward velocity. The steering command is computed as:

\begin{equation}
\mathrm{Steer} = k_{\mathrm{p}} \cdot \mathrm{CE} + k_{\mathrm{d}} \cdot \mathrm{HE} + k_{\mathrm{i}} \int \mathrm{CE}\, \mathrm{d}t
\end{equation}

In final deployment, PID also regulates speed: when heavy curvature is detected, the proportional component automatically reduces forward velocity, improving cornering stability. A higher-priority obstacle-response P-regulator supersedes normal lane PID when vision or onboard sensors detect objects obstructing the path.

\paragraph{Stanley Controller.}

The Stanley controller~\cite{thrun2006,hoffmann2007} is the primary steering law used for continuous lane following due to its kinematic and speed-adaptive formulation. The steering angle is computed as:

\begin{equation}
\mathrm{Steer}
=
k_{\mathrm{he}} \cdot \mathrm{HE}
+
\tan^{-1}
\left(
\frac{k_{\mathrm{ce}} \cdot \mathrm{CE}}{v_{\mathrm{f}} + k_{\mathrm{s}}}
\right)
\end{equation}

where \(v_{\mathrm{f}}\) is the forward velocity and \(k_{\mathrm{s}}\) prevents singular behavior near zero speed.

This model-based formulation provides smoother transitions during acceleration and deceleration, reducing oscillation at higher speeds. Its direct grounding in the vehicle geometry makes it superior to a pure PID controller for continuous closed-loop lane tracking.

\paragraph{Pure Pursuit Controller.}

\begin{figure}[H]
    \centering
    \includegraphics[width=0.45\linewidth]{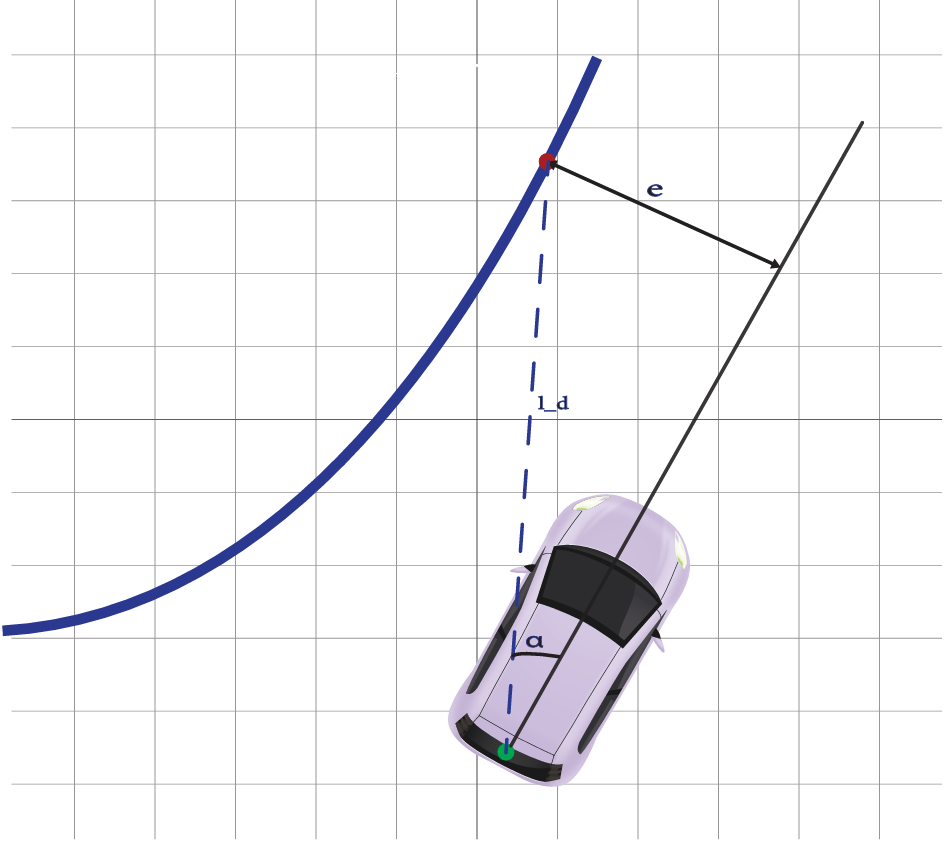}
    \caption{Pure Pursuit geometry illustrating the look-ahead point, look-ahead distance, and heading error angle.}
    \label{fig:pure_pursuit}
  \end{figure}

The Pure Pursuit controller~\cite{coulter1992,snider2009} computes the steering angle based on a geometric look-ahead point projected onto the lane reference, determined from the rear-axle position (Fig.~\ref{fig:pure_pursuit}). The steering angle is calculated as

\begin{equation}
\mathrm{Steer} =
\tan^{-1}
\left(
\frac{2 \sin\alpha \cdot L}{k_{\mathrm{pp}} \cdot v_{\mathrm{f}}}
\right)
\end{equation}

where $L$ is the vehicle wheelbase, $l_d=k_\mathrm{pp}\cdot v_f$ is the look-ahead distance, $k_\mathrm{pp}$ is the look-ahead gain, and $v_f$ is the forward velocity. The angle $\alpha$ is defined as the angle between the vehicle's longitudinal heading direction and the line connecting the rear-axle position to the look-ahead point on the lane reference.

This method is particularly effective for tracking curved roads and complex trajectories, as it determines the steering command from the geometric relationship between the vehicle and the look-ahead point. The speed-dependent look-ahead distance allows the controller to adapt its tracking behavior to the vehicle's forward speed.

\paragraph{Gain Calibration.}

Controller parameters for PID, Stanley, and Pure Pursuit were tuned empirically through iterative testing. Real-time visualization of control variables was used while adjusting responsiveness, lateral settling time, overshoot, and high-speed stability. The final gains delivered fast, non-oscillatory velocity control, minimal steady-state lane deviation at moderate/high speeds, and smooth cornering without abrupt steering.

\subsection{Decision Making, Parking, and Intersection Handling}
\label{subsec:decision_making}

The decision layer of the vehicle converts high-level perception outputs into discrete driving actions. All logic is structured as a Finite State Machine (FSM) that evaluates lane geometry, traffic signs, crosswalk signals, and feedback from onboard sensors, then transitions between states such as \textit{lane following}, \textit{parking}, and \textit{intersection traversal}.

\subsubsection{Parking Behavior}
\label{subsubsec:parking}
\begin{figure}[H]
  \centering
  \includegraphics[width=0.96\linewidth]{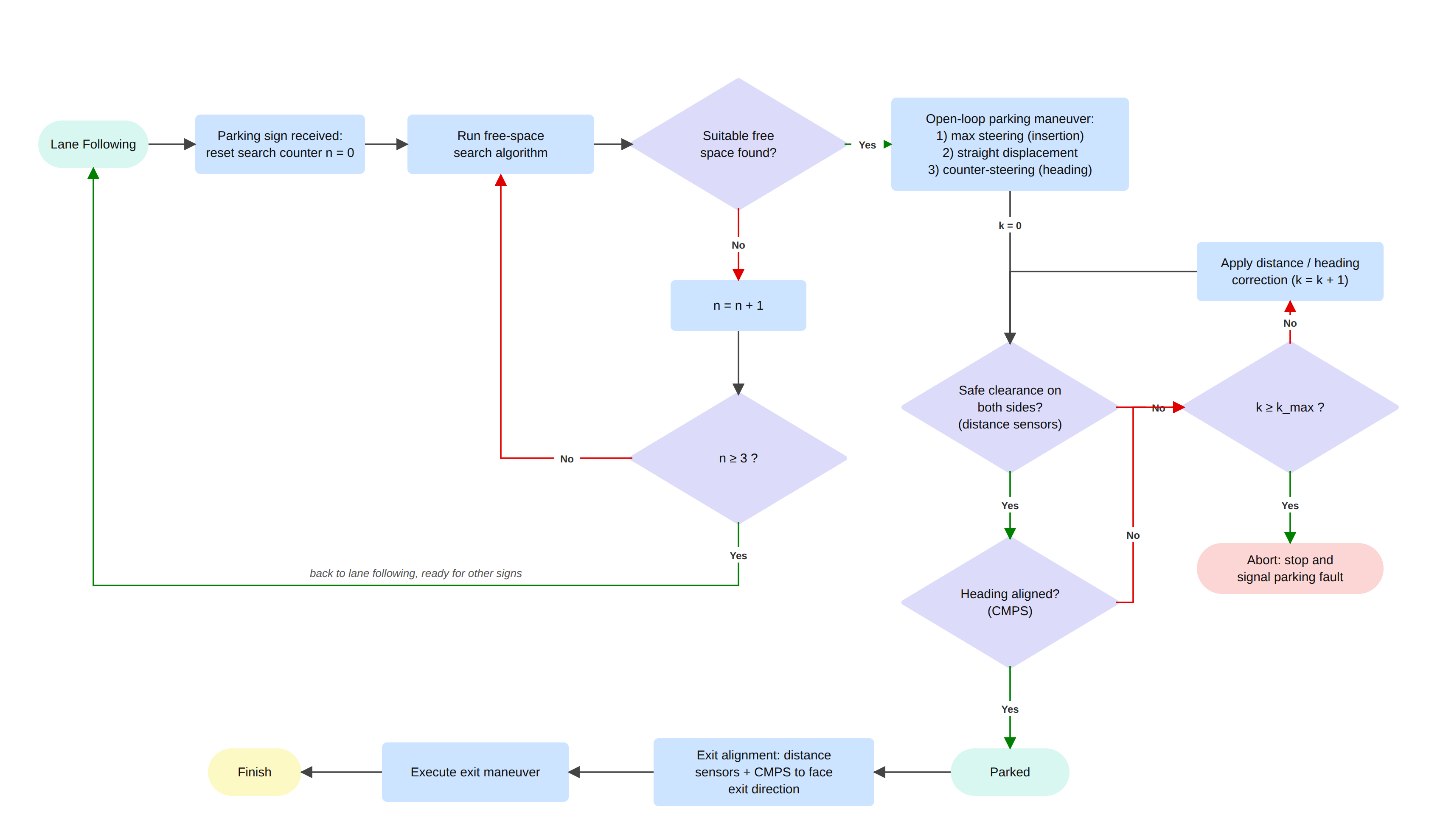}
  \caption{Finite State Machine for autonomous parking behavior.}
  \label{fig:parking_fsm}
\end{figure}

During normal navigation, the vehicle continues lane following until a parking sign is detected by the vision subsystem. Once detected, the FSM transitions into the parking sequence shown in Fig.~\ref{fig:parking_fsm}. The vehicle searches for a valid free space using onboard sensors and begins the maneuver as soon as a suitable location is detected.

The parking process consists of two main phases.

\textbf{Open-Loop Geometric Maneuver:}  
An open-loop geometric maneuver was developed to achieve the required vehicle alignment after turning. The maneuver consists of three sequential phases: maximum steering for trajectory insertion, straight displacement for position adjustment, and counter-steering for final heading correction. The motion parameters were calculated using the kinematic bicycle model and further refined through empirical calibration to compensate for practical effects such as tire slip, actuator delay, and mechanical backlash, resulting in repeatable and accurate maneuver execution.

\textbf{Feedback-Based Correction:}  
Once the initial maneuver completes, sensor data (vision and IMU) are used to refine the final position. The negative feedback loop corrects residual lateral and angular error to ensure precise alignment within the parking space.

\begin{figure}[H]
  \centering
  \includegraphics[width=0.7\linewidth]{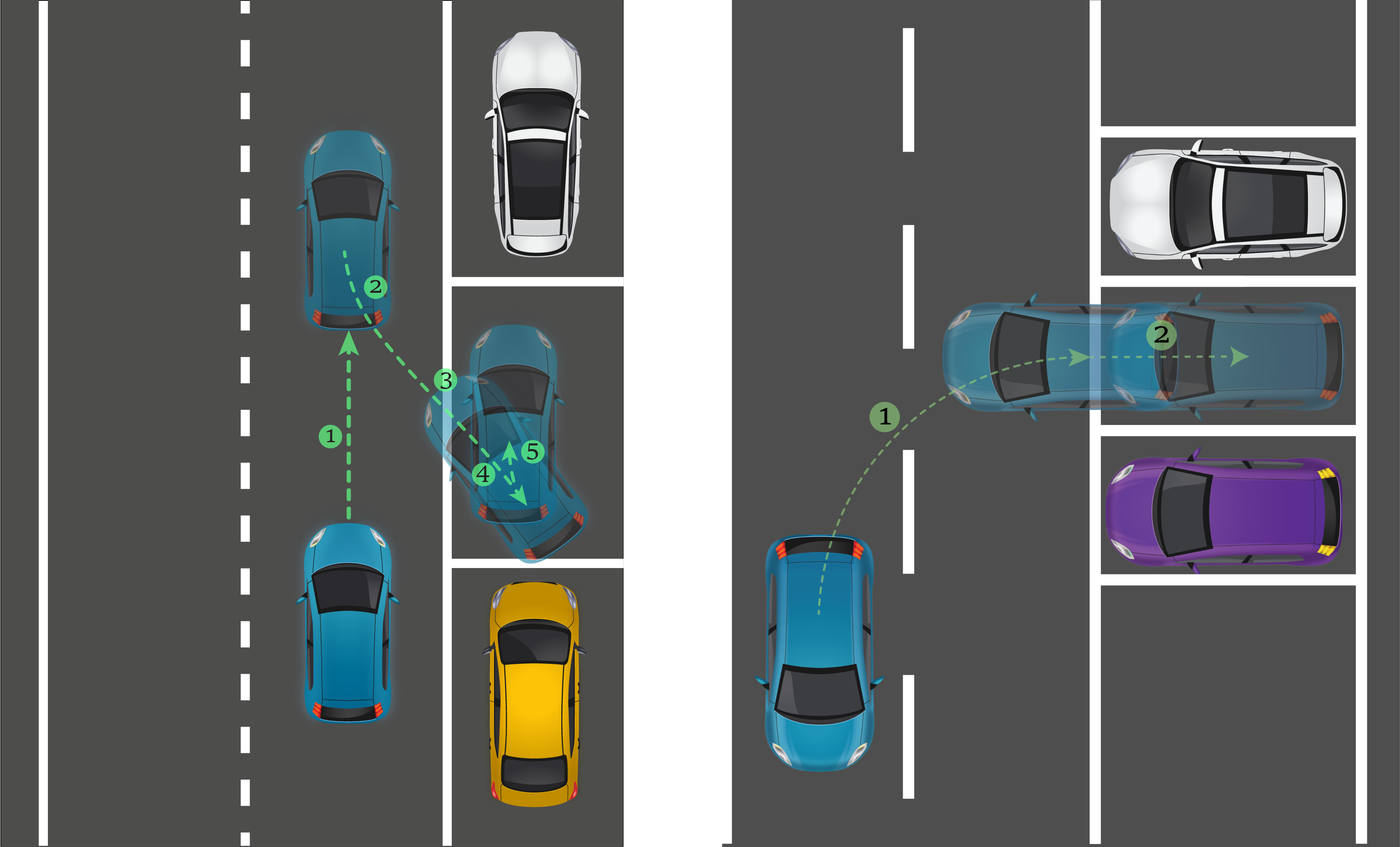}
  \caption{Example parking results: left -- parallel parking; right -- perpendicular (cross) parking.}
  \label{fig:parking_modes}
\end{figure}

Fig.~\ref{fig:parking_modes} illustrates example end states for parallel and perpendicular (cross) parking.

\subsubsection{Sign-Approach Distance Estimation on Curved Paths}
\label{subsubsec:sign_approach}

Correctly estimating the remaining distance to a sign matters beyond the intersection maneuver of section~\ref{subsubsec:intersection_turning}: the same estimate also drives the initial approach of the parking maneuver (section~\ref{subsubsec:parking}). Once the park sign is detected, the vehicle must advance by the corresponding encoder distance so that its side-facing distance sensor reaches approximately the correct position; specifically, the free-space search algorithm for parking is initiated about 10~cm past the point where the park sign was first sighted. In this case, as in the intersection approach, the straight-line distance reported by the vision pipeline is not, in general, the distance the vehicle will actually travel: whenever the road curves during the approach, the vehicle follows a curved path rather than a straight line toward the sign, which the procedure below accounts for.

As long as the vision pipeline reports a valid sign distance, the vehicle drives directly on that live reading. From the moment this data is lost --- or is deliberately no longer used --- the required travel distance is instead estimated from three quantities available onboard: the yaw rate $\dot{\theta}$ obtained from the compass (CMPS), the vehicle speed $V$, and the last valid sign distance $D$ reported by vision before tracking was lost. Three cases are distinguished, summarized in Table~\ref{tab:sign_cases} and detailed below.

\begin{table}[H]
\centering
\caption{Sign-approach distance estimation: case selection based on yaw rate and the vision-reported distance.}
\label{tab:sign_cases}
\small
\renewcommand{\arraystretch}{1.2}
\begin{tabular}{@{}cp{3.6cm}p{9.4cm}@{}}
\toprule
\textbf{Case} & \textbf{Condition} & \textbf{Action} \\ \midrule
1 & $|\dot{\theta}| < \dot{\theta}_{\mathrm{min}}$ & Path assumed locally straight: advance the last known sign distance $D$ directly, commanded by the encoder. \\ \midrule
2 & $|\dot{\theta}| > \dot{\theta}_{\mathrm{min}}$ and $D \geq R$ & Curve detected, but the arcsine relation of Case 3 is outside its valid domain ($D/R \geq 1$); fall back to the same straight-line action as Case 1. \\ \midrule
3 & $|\dot{\theta}| \geq \dot{\theta}_{\mathrm{min}}$ and $D \leq R$ & Curve detected and the sign lies within the estimated turn radius: use the arc-length correction of Eqs.~\ref{eq:arc_angle}--\ref{eq:arc_length} instead of the raw distance $D$. \\
\bottomrule
\end{tabular}
\end{table}

\textbf{Case 1 --- straight path.} If the magnitude of the yaw rate falls below a minimum threshold $\dot{\theta}_{\mathrm{min}}$, the road is assumed to be locally straight for the remainder of the approach. The vehicle then simply continues toward the sign using the last known distance $D$, commanding the encoder to cover exactly that distance.

\textbf{Cases 2--3 --- curved path.} Otherwise the vehicle is in a curve, and the local radius of curvature is estimated from the current motion state as
\begin{equation}
R = \frac{V}{\dot{\theta}}
\label{eq:radius_curv}
\end{equation}
If $D > R$ (Case 2), the sign lies beyond the estimated turn radius; the correction below is outside its valid range, so the vehicle falls back to the same straight-line action as Case 1, using $D$ directly.

\begin{figure}[H]
    \centering
    \includegraphics[width=0.4\linewidth]{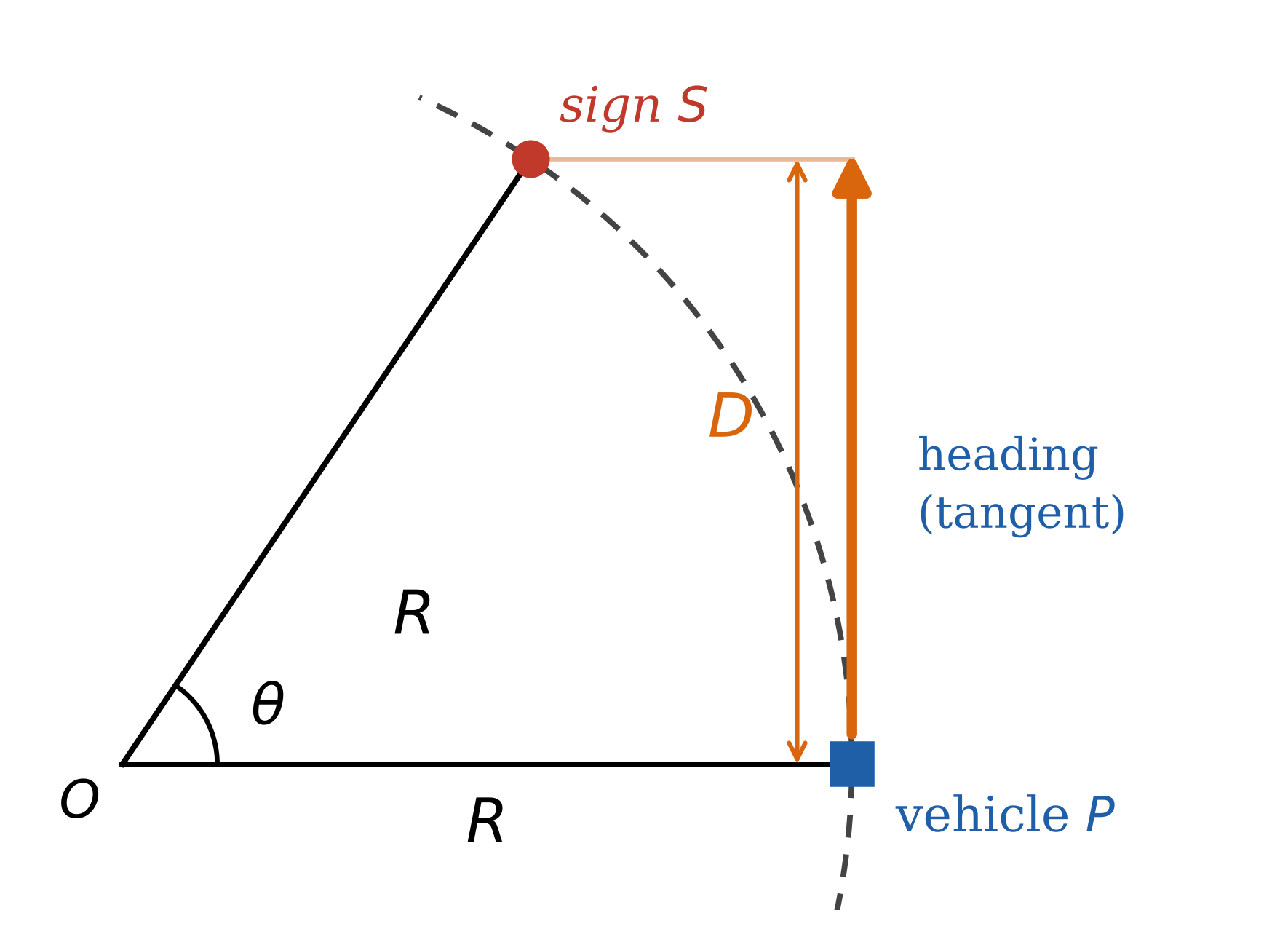}
    \caption{Geometry relating the estimated turn radius $R$, the vision-reported forward distance $D$, and the subtended angle $\theta$.}
    \label{fig:sign_geometry}
\end{figure}

\textbf{Case 3 --- within-radius correction.} If $D \leq R$, the sign lies within the estimated turn radius, and the raw distance $D$ increasingly underestimates the true travel distance as the path curves more sharply. Assuming the curvature does not change appreciably before the sign is reached (Fig.~\ref{fig:sign_geometry}), the subtended arc angle is

\begin{equation}
\theta = \sin^{-1}\!\left(\frac{D\cdot\dot{\theta}}{V}\right) = \sin^{-1}\!\left(\frac{D}{R}\right)
\label{eq:arc_angle}
\end{equation}
and the encoder is commanded to travel the corresponding arc length rather than the chord distance $D$:
\begin{equation}
d_{\mathrm{encoder}} = R\cdot\theta
\label{eq:arc_length}
\end{equation}
which gives a more accurate stopping distance than driving the raw vision distance $D$ whenever the path curvature is significant relative to the remaining distance to the sign.

\subsubsection{Intersection Control and Turning Logic (Version 4)}
\label{subsubsec:intersection_turning}

\begin{figure}[H]
  \centering
  \includegraphics[width=0.60\linewidth]{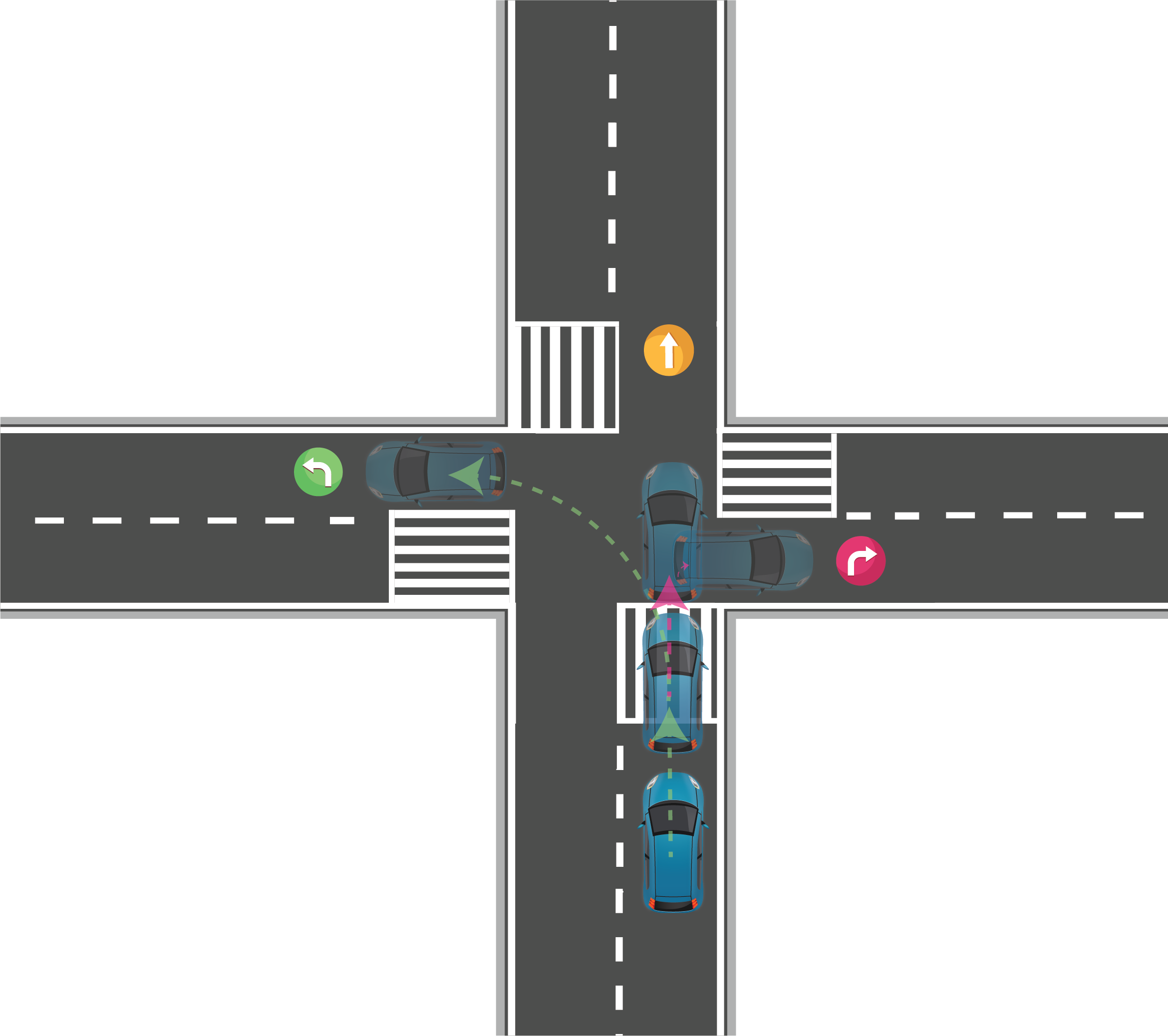}
  \caption{Available intersection actions: right turn, left turn, forward crossing.}
  \label{fig:intersection_options}
\end{figure}

When approaching an intersection, the correct maneuver is determined based on the recognized sign. The three supported actions (turn right, turn left, and go straight) are shown in Fig.~\ref{fig:intersection_options}.

\begin{figure}[H]
    \centering
    \includegraphics[width=0.65\linewidth]{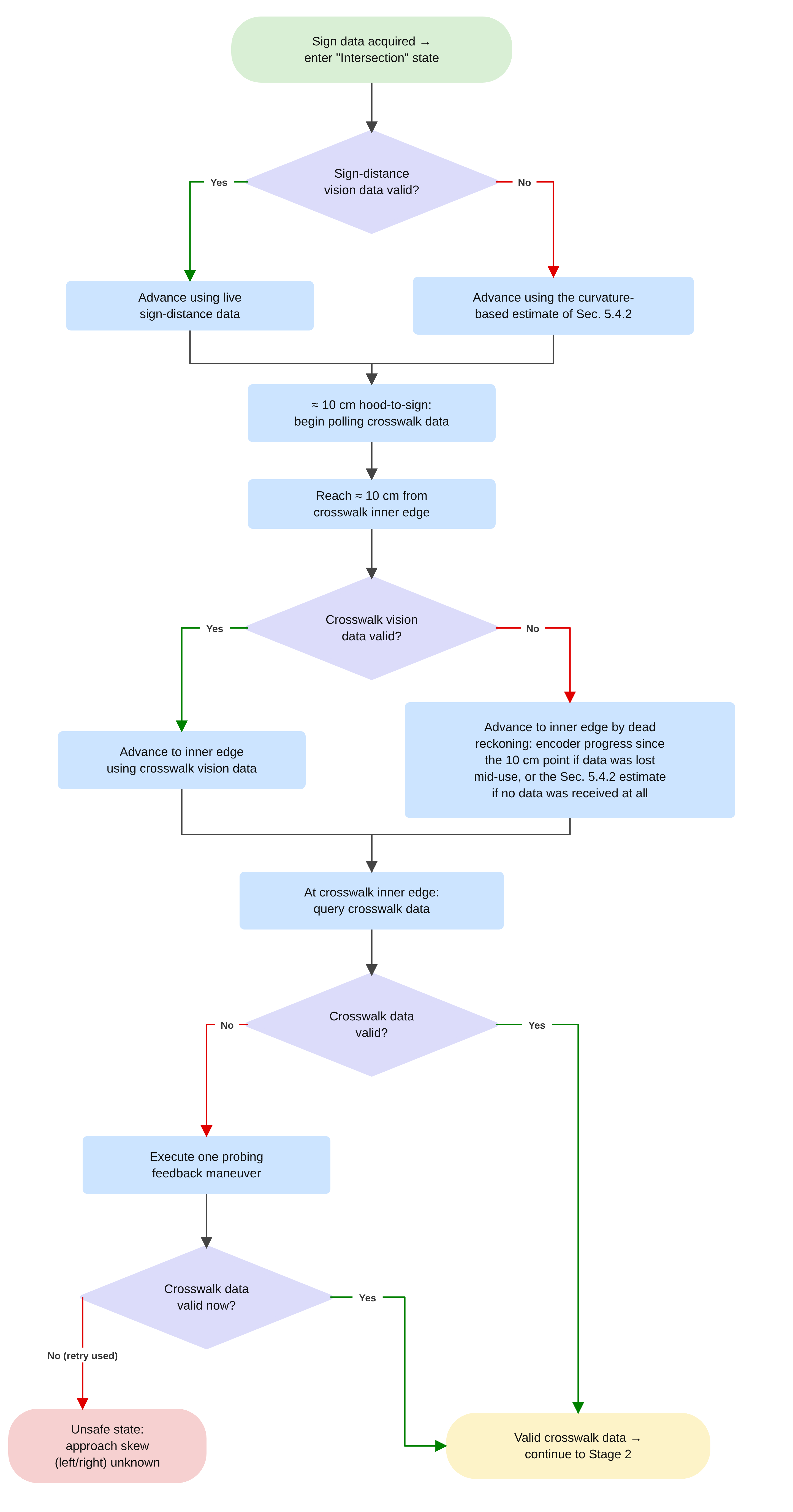}
    \caption{Stage 1 --- sign-distance approach and crosswalk-data acquisition at the crosswalk inner edge, with a single probing retry.}
    \label{fig:stage1_flow}
\end{figure}

\begin{figure}[H]
    \centering
    \includegraphics[width=0.7\linewidth]{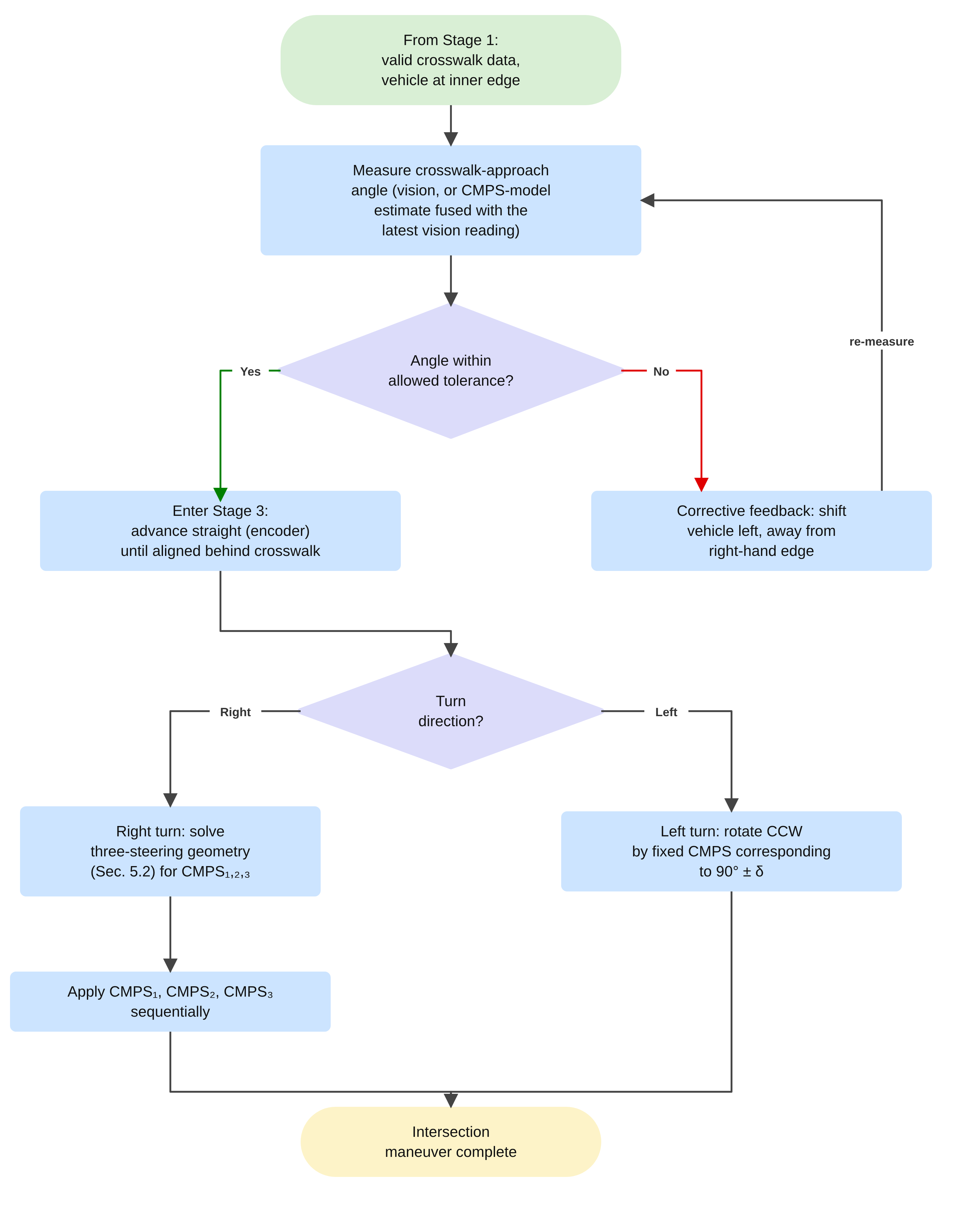}
    \caption{Stage 2 --- iterative crosswalk pose correction, and Stage 3 --- turn execution (unchanged).}
    \label{fig:stage23_flow}
\end{figure}

Once sign data is first acquired, the vehicle enters the ``Intersection'' state, and the maneuver proceeds through three stages: (1) approach and crosswalk-data acquisition, (2) pose correction relative to the crosswalk, and (3) turn execution. Each stage is described below and illustrated in Figures~\ref{fig:stage1_flow} and~\ref{fig:stage23_flow}.

\paragraph{Stage 1 --- Approach and crosswalk-data acquisition (Fig.~\ref{fig:stage1_flow}).}
Because the probability of losing the sign from the camera's field of view increases as the vehicle gets closer to it, the approach to the sign is not driven by live vision data alone: at each control step the vehicle checks whether current sign-distance vision data is valid, advancing on that live reading if so, and otherwise falling back to the curvature-based estimate of section~\ref{subsubsec:sign_approach}. As the vehicle nears the hood-to-sign offset ($\approx 10$~cm), it begins polling for crosswalk data as well.

When the vehicle subsequently reaches $\approx 10$~cm from the crosswalk's inner edge, it checks whether crosswalk vision data is currently valid. If so, it advances the remaining distance to the inner edge on that data directly. If not --- whether because crosswalk data had been in use and was suddenly lost, or because no crosswalk data had been received at all since the 10~cm point --- the vehicle instead completes the approach by dead reckoning: in the first case, from the encoder progress accumulated since the 10~cm point; in the second, from the full curvature-based estimate of section~\ref{subsubsec:sign_approach}.

Once at the crosswalk inner edge, the vehicle queries crosswalk data once more. If it is valid, the vehicle proceeds directly to Stage 2. If not, a single probing feedback maneuver is executed and the query is repeated exactly once; if crosswalk data is still unavailable after this one retry, the vehicle is considered to be in an unsafe state, since the direction and magnitude of any approach skew remain unknown.

\begin{figure}[H]
    \centering
    \includegraphics[width=0.85\linewidth]{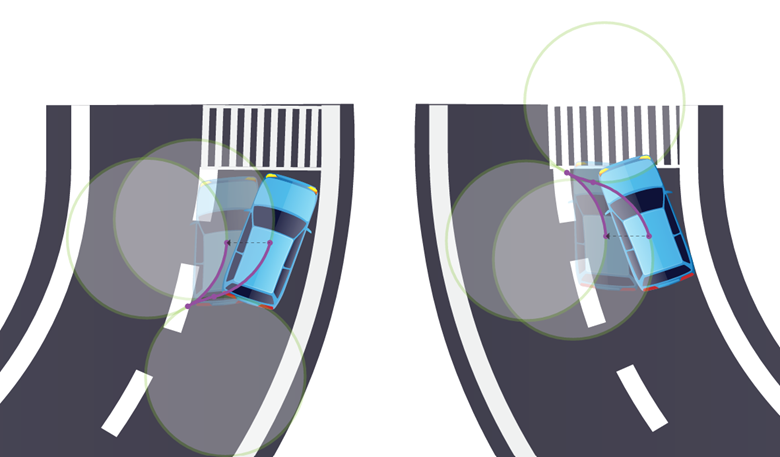}
    \caption{Feedback maneuver design for crosswalk alignment and lateral correction. The illustrated maneuver positions the vehicle behind the crosswalk while shifting it slightly to the left. Alternative feedback maneuvers can achieve the same positioning and lateral shift without necessarily aligning the vehicle perpendicular to the crosswalk.}
    \label{fig:intersection_illustration}
\end{figure}

\paragraph{Stage 2 --- Pose correction relative to the crosswalk (Fig.~\ref{fig:stage23_flow}).}
With valid crosswalk data available, the crosswalk-approach angle is measured. If it exceeds the allowed tolerance, a corrective feedback maneuver shifts the vehicle a modest amount to the left, away from the right-hand corner of the intersection, and crosswalk data is queried again to re-measure the angle; this correction loop repeats until the angle falls within tolerance (see Fig.~\ref{fig:intersection_illustration}). Rather than requiring a fresh valid vision reading at every iteration, the angle change produced by each feedback step can instead be estimated from the commanded CMPS rotation and fused with the most recent vision reading, reducing dependence on repeated crosswalk detections.

\paragraph{Stage 3 --- Turn execution (Fig.~\ref{fig:stage23_flow}).}
Once the crosswalk-approach angle is within tolerance, the vehicle first advances straight, by encoder count, until it is aligned just behind the crosswalk. The subsequent action depends on the turn direction: for a right turn, the measured crosswalk-approach angle is substituted into the three-steering-wheel geometry of section~\ref{subsec:steering_limit} to solve for the three required encoder (CMPS) values, which are then applied to the vehicle in sequence; for a left turn, the vehicle instead performs a single rotation of fixed magnitude, counter-clockwise, corresponding to an encoder travel of $90^\circ \pm \delta$, where $\delta$ accounts for the observed approach angle.

\subsubsection{Evolution of the Intersection-Handling Strategy}
\label{subsubsec:intersection_evolution}

The intersection-handling logic was developed iteratively over four versions, moving from a fixed, open-loop maneuver to a closed-loop, kinematically grounded approach. Table~\ref{tab:intersection_gens} summarizes the progression; the following subsections describe each version in more detail.

\begin{table}[p]
\centering
\caption{Summary of the four versions of the intersection-handling algorithm.}
\label{tab:intersection_gens}
\footnotesize
\setlength{\tabcolsep}{3pt}
\renewcommand{\arraystretch}{1.15}
\begin{tabular}{@{}p{1.1cm}p{4.9cm}p{3.9cm}p{5.6cm}@{}}
\toprule
\textbf{Version} & \textbf{Core Approach} & \textbf{Key Strengths} & \textbf{Key Limitations} \\ \midrule
Ver.~1 &
Open-loop, hardcoded actuation triggered at a fixed 10~cm distance from the sign (e.g., advance 30~cm, then turn until a preset heading change is sensed by the compass). &
Minimal development time. &
Only IMU feedback; almost entirely dependent on a fixed approach geometry. \\ \midrule
Ver.~2 &
Added intersection-approach angle. A PID loop first corrects heading error, then the same hardcoded turning actions as Ver.~1 are executed. &
Introduces road-relative feedback. Simpler algorithm; fewer tuneable parameters. &
High perception noise from a low-quality, poorly mounted camera; the vision data was only reliable in a narrow crosswalk viewing position, so the feedback loop had to re-sample it 2--3 times. Failed when the vehicle approached skewed toward the right edge of the road. Could not handle intersections on curved roads. Two-lane validation was added but was not accurate enough to enable feedback after the crosswalk or inside the intersection. \\ \midrule
Ver.~3 &
Generalized further: road curvature ($\kappa$) was introduced to estimate the distance the vehicle must travel through a bend to reach the sign, with turn direction (CW/CCW) taken from perception. &
Broader applicability across road geometries. &
More parameters to tune and longer development time than earlier versions. Increased generality reduced reliability: at plain, straight intersections the algorithm could still unnecessarily factor in crosswalk angle and curvature, causing failures over part of the intersection design space. \\ \midrule
Ver.~4 &
Adopted the vehicle's bicycle-model kinematics to reach the target lane more reliably and directly, motivated by a reduced maximum steering angle that had been increasing the distance travelled in the opposing lane. Repositioning between an initial and target pose is solved as a three-segment steering maneuver (section~\ref{subsec:steering_limit}), with the required encoder (CMPS) travel computed for each segment. The same model-based CMPS estimation is used to judge how far the vehicle must travel through a curve to reach a set distance from the sign, replacing noisy curvature-based perception. A dedicated feedback behavior biases the vehicle slightly left to counteract drift toward the right edge; once the vehicle reaches the crosswalk at a shallow, safe angle with no risk of edge proximity, the three-segment maneuver is executed using the measured crosswalk-approach angle. &
More reliable: grounded in the actual vehicle kinematic model rather than hardcoded actions. More accurate sign approach in sharp intersection turns, by relying on the vehicle model and CMPS changes instead of noisy curvature perception. &
Highly sensitive to the mechanical design, steering calibration (section~\ref{subsec:validation_control}) and measurement accuracy (section~\ref{sec:embedded}--\ref{sec:control}); imprecise geometry directly degrades the open-loop maneuver. Tends to look mechanically stiff, and a reverse segment is less visually natural --- a potential drawback for real-world user acceptance. \\
\bottomrule
\end{tabular}
\end{table}

\paragraph{Version 1 --- Open-Loop Baseline.}
The first version implemented the simplest possible behavior: once the vehicle was within 10~cm of the sign, a hardcoded sequence of actions executed the turn open-loop, with no feedback --- for example, advancing 30~cm and then turning until the onboard compass (CMPS) sensed a preset change in heading, at which point the maneuver was considered complete. Its only advantage was a very short development time; it could not compensate for any deviation from the assumed approach geometry.

\paragraph{Version 2 --- Heading-Error Feedback.}
The second version added the intersection-approach angle to the pipeline and began handling the heading error caused by a skewed arrival at the intersection. A PID controller first corrected this heading error, after which the vehicle executed the same type of hardcoded turning actions as version 1. This version incorporated road-relative feedback for the first time and kept the algorithm simple, with few design parameters. However, the camera in use was of low quality and poorly positioned on the chassis, so the processed vision data carried substantial error; because valid-looking data was only obtained from one specific viewing position relative to the crosswalk, the motion-feedback loop had to re-acquire vision data two to three times per approach. The version performed poorly whenever the vehicle's pose relative to the road was already skewed toward the right edge, and it could not handle intersections located on curved roads. Two-lane detection was added to verify that the target lane had been reached, but its accuracy was insufficient, so feedback after the crosswalk and inside the intersection itself remained unavailable.

\paragraph{Version 3 --- Curvature Generalization.~\cite{auriga2024}}
The third version generalized the approach further by introducing an initial implementation of road curvature ($\kappa$) to estimate the distance the vehicle needed to travel through a bend to reach the sign, with the turn direction (clockwise or counter-clockwise) taken from the perception pipeline. This broadened the range of road geometries the algorithm could address, but at the cost of more tunable parameters and a longer development time than the previous versions. More importantly, the added generality reduced reliability: because the logic could still weight crosswalk angle and curvature even at plain, straight intersections, it occasionally failed on a subset of intersection types that the earlier, simpler logic had handled correctly.

\paragraph{Version 4 --- Bicycle-Model-Based Maneuvering.}
The fourth version targeted more reliable motion: reaching the target lane faster and minimizing time spent in the opposing lane. A reduction in the vehicle's maximum steering angle had made this worse, forcing a long excursion in the wrong lane before the vehicle could correct itself. To address this, the version adopted the vehicle's motion geometry --- the bicycle model (section~\ref{subsec:kinematics}) --- based on the observation that the vehicle's motion lies on a circular arc. This raised the geometric question of how to move between two poses with defined initial and target headings when a single steering action cannot satisfy the constraint; the resulting solution was the three-segment steering maneuver described in section~\ref{subsec:steering_limit}, in which the required encoder (CMPS) travel for each of the three segments is computed and applied to the vehicle. The same model-based CMPS estimation was reused to judge how far the vehicle should travel through a curve to reach a specified distance from the sign, replacing the noisy curvature-based perception of version 3. A dedicated feedback behavior was added to counter drift toward the right edge of the road: it biases the vehicle slightly to the left and is shaped so that its behavior does not itself tend toward the right edge. Once the vehicle reaches the crosswalk at a shallow, safe angle with no risk of edge proximity, the three-segment maneuver is executed using the measured crosswalk-approach angle.

Grounding the maneuver in the bicycle model made the behavior more reliable overall, since it replaced hardcoded actions with the actual vehicle kinematic model. It also improved accuracy when approaching the sign through sharp intersection turns, since the vehicle no longer depended on curvature data derived from noisy road-curve perception, relying instead on the vehicle model and CMPS changes. The main drawback is that the three-segment maneuver is strongly dependent on the vehicle's mechanical design and on measurement accuracy (Sections~\ref{sec:hardware}--\ref{sec:control}): imprecise geometry directly reduces maneuver accuracy. In addition, the maneuver tends to look mechanically stiff, and when a reverse segment is required it can appear visually unnatural --- a consideration for acceptance in real-world use.

\subsection{Experimental Results and Validation}
\label{subsec:validation_control}
\subsubsection{Steering Calibration}
\label{subsec:ackermann_valid}

To calibrate the steering mechanism fabricated by the mechanical engineering team and to relate the commanded input u to the wheel angles, an empirical data collection procedure was conducted in coordination with the embedded control subsystem.

At every setpoint, the combined (equivalent single-track) steering angle $\delta_{\mathrm{comb}}$ reported below is computed directly from Eq.~\ref{eq:nonlin_steering_equation}, i.e. as $\cot^{-1}\!\left(\frac{\cot \delta_{\mathrm{left}} + \cot \delta_{\mathrm{right}}}{2}\right)$ applied to each individual trial, rather than as the simple arithmetic mean of the two measured wheel angles, so that it is consistent with the bicycle model of section~\ref{subsec:kinematics}.

The embedded controller maps control commands $u\in\{0, 1, \dots, 255\}$ to the steering actuator (servo motor). The servo hardware is configured to operate across a nominal angular range of $-40^{\circ}$ to $+40^{\circ}$ with an estimated positioning variance of $\sigma^2_{\mathrm{servo}}=1.0^{\circ}$. A total of 13 discrete control input setpoints were evaluated, spanning from 255 down to 0. At each setpoint, 3 independent measurements were collected for both the left wheel angle ($\delta_{\mathrm{left}}$) and the right wheel angle ($\delta_{\mathrm{right}}$).

\begin{figure}[H]
    \centering
    \includegraphics[width=0.8\linewidth]{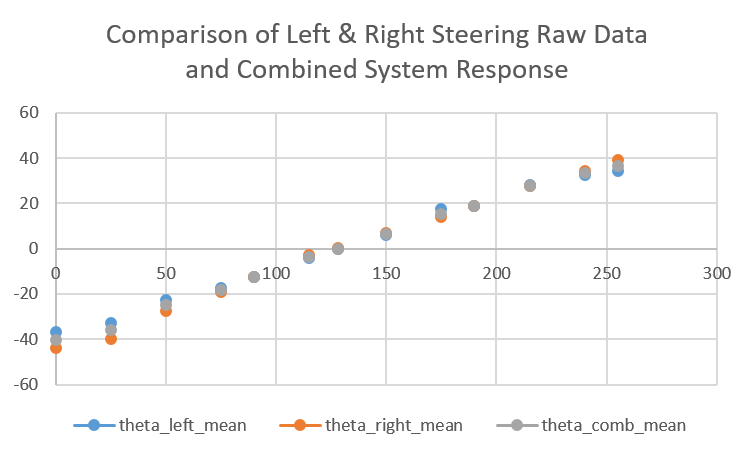}
    \caption{Front-left and front-right wheel steering angles, alongside the equivalent steering angle derived from the Ackermann relationship.}
    \label{fig:steering_raw_data}
  \end{figure}

At extreme steering inputs, such as $u = 0$ and $u = 255$, the two front wheels turn by different angles (Fig.~\ref{fig:steering_raw_data}), so the steering is not parallel. The equivalent angle $\delta_\text{comb}$ of Eq.~\ref{eq:nonlin_steering_equation} is defined for any pair of wheel angles; it therefore calibrates the input of the bicycle model but does not by itself verify the Ackermann condition.

\begin{figure}[H]
        \centering
        \includegraphics[width=0.9\linewidth]{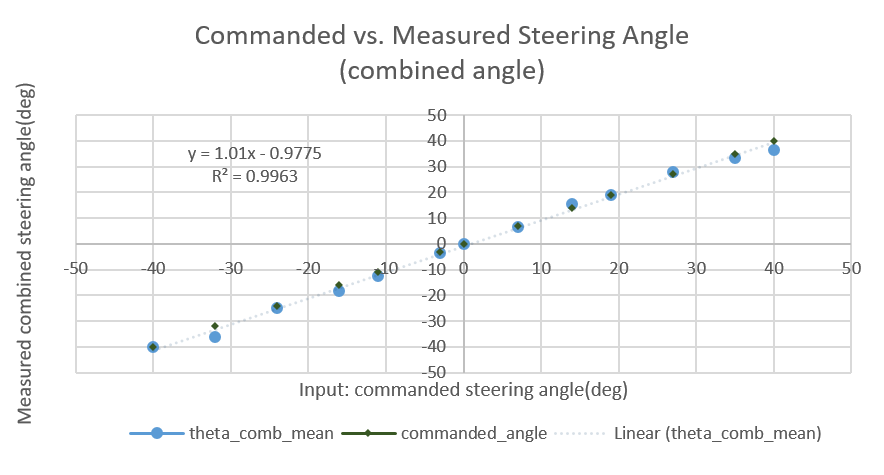}
        \caption{Commanded (input) steering angle versus the combined measured steering angle (mean\_steer), computed solely from the cot-average of Eq.~\ref{eq:nonlin_steering_equation} --- the individual per-wheel means are not used in this comparison. The dashed line marks perfect agreement (measured = commanded); error bars show $\pm 1$ StdDev across the 3 trials.}
        \label{fig:steering_command_vs_angle}
      \end{figure}

  Across the 13 tested setpoints, a one-sample t-test with a significance level of $\alpha=0.05$ showed that the combined steering angle ($\delta_{\mathrm{comb}}$), computed from the per-trial wheel angles via the cot-average relation of Eq.~\ref{eq:nonlin_steering_equation} rather than a simple arithmetic average, does not have a statistically significant deviation from the desired servo setpoint in 92.3\% (12 of 13 setpoints) of the operating range, with P-values greater than 0.05 at every setpoint except for $u=25$. This indicates a high level of agreement between the commanded and measured steering angles. Table~\ref{tab:steering_ttest} and Fig.~\ref{fig:steering_command_vs_angle} show this result.
  
  \begin{table}[H]
    \centering
    \caption{Commanded steering angle versus the combined measured steering angle $\delta_{\mathrm{comb}}$, computed from the per-trial wheel angles via the cot-average relation of Eq.~\ref{eq:nonlin_steering_equation}. Statistics are over 3 trials per setpoint; the one-sample t-test uses $\alpha=0.05$.}
    \label{tab:steering_ttest}
    \small
    \setlength{\tabcolsep}{4pt}
    \renewcommand{\arraystretch}{1.1}
    \resizebox{\textwidth}{!}{%
    \begin{tabular}{@{}rrrrrrrrc@{}}
    \toprule
    \textbf{Control} & \textbf{Commanded} & \textbf{Combined} & \textbf{Combined} & \textbf{Combined} & \textbf{Mean} & \textbf{T-} & \textbf{P-} & \textbf{Significant} \\
    \textbf{Input $u$} & \textbf{Angle ($^\circ$)} & \textbf{Mean ($^\circ$)} & \textbf{StdDev ($^\circ$)} & \textbf{Var ($^\circ$)} & \textbf{Diff. ($^\circ$)} & \textbf{Statistic} & \textbf{Value} & \textbf{($\alpha=0.05$)} \\ \midrule
    0   & $-40$ & $-40.198$ & 2.204 & 4.857 & $-0.198$ & $-0.156$ & 0.8904 & No  \\
    25  & $-32$ & $-35.988$ & 1.068 & 1.140 & $-3.988$ & $-6.469$ & 0.0231 & Yes \\
    50  & $-24$ & $-24.917$ & 1.013 & 1.027 & $-0.917$ & $-1.567$ & 0.2577 & No  \\
    75  & $-16$ & $-18.249$ & 1.374 & 1.887 & $-2.249$ & $-2.835$ & 0.1051 & No  \\
    90  & $-11$ & $-12.321$ & 1.888 & 3.565 & $-1.321$ & $-1.211$ & 0.3494 & No  \\
    115 & $-3$  & $-3.552$  & 2.854 & 8.146 & $-0.552$ & $-0.335$ & 0.7696 & No  \\
    128 & 0     & 0         & 0     & 0     & 0        & 0        & 1.0000 & No  \\
    150 & 7     & 6.382     & 2.197 & 4.825 & $-0.618$ & $-0.488$ & 0.6741 & No  \\
    175 & 14    & 15.217    & 2.791 & 7.792 & 1.217    & 0.755    & 0.5290 & No  \\
    190 & 19    & 18.991    & 1.501 & 2.253 & $-0.009$ & $-0.010$ & 0.9931 & No  \\
    215 & 27    & 27.825    & 1.528 & 2.335 & 0.825    & 0.935    & 0.4484 & No  \\
    240 & 35    & 33.480    & 1.815 & 3.293 & $-1.520$ & $-1.451$ & 0.2839 & No  \\
    255 & 40    & 36.515    & 2.158 & 4.657 & $-3.485$ & $-2.797$ & 0.1076 & No  \\
    \bottomrule
    \end{tabular}%
    }
    \end{table}

\begin{figure}[H]
        \centering
        \includegraphics[width=0.7\linewidth]{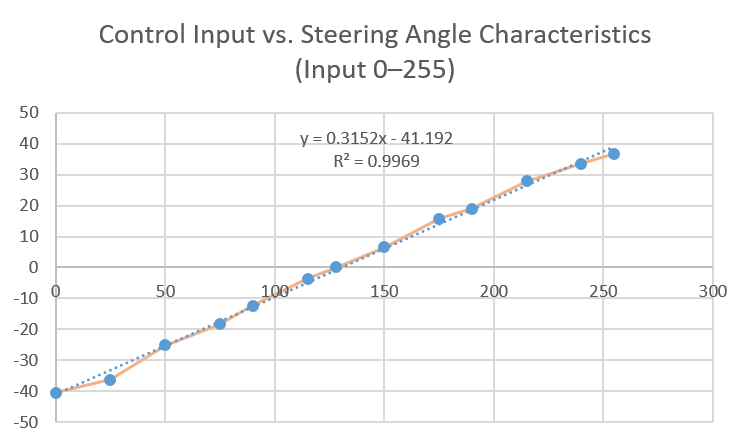}
        \caption{Linear regression model for control input $u$ vs. steering angle $\delta$.}
        \label{fig:linear_reg_steering}
      \end{figure}

Additionally, the alignment of the steering system around the neutral position was checked. The left wheel crosses $0^\circ$ at $u \approx 129$ and the right wheel at $u \approx 133$; at $u = 128$ the right wheel reads $-1.5^\circ$ and the left wheel $-0.2^\circ$.  Larger angular offsets between the calibrated and the actual wheel angles appear at large $|\delta|$ as residuals outside the $\pm 2\, \mathrm{cm}$ band in Fig.~\ref{fig:bicycle_residual_plot} and as the systematic trends of Fig.~\ref{fig:residual_vs_angle} (fitted offsets in Table~\ref{tab:residual_stats}).
For closed-loop path tracking and high-level trajectory planning the relationship between the 8-bit integer control input $u$ and the equivalent bicycle steering angle $\delta$ is modeled by Eq.~\ref{eq:input_vs_steer_model}. This fit is obtained by a direct linear regression of $u$ against the combined (equivalent single-track) steering angle of Eq.~\ref{eq:nonlin_steering_equation}, evaluated from the per-trial measurements at each setpoint (as shown in Fig.~\ref{fig:linear_reg_steering}) --- not by averaging the two individual wheel calibrations below, since a simple arithmetic average of $\delta_{\mathrm{left}}$ and $\delta_{\mathrm{right}}$ is not equivalent to Eq.~\ref{eq:nonlin_steering_equation}.
The separate per-wheel calibrations $\delta_{\mathrm{right}}(u)$ and $\delta_{\mathrm{left}}(u)$ are also present in Eq.~\ref{eq:input_vs_steer_model}, but are not used to derive $\delta(u)$.
\begin{equation}
    \delta(u)=0.3133u-40.923^{\circ}, \quad \begin{cases}
    \delta_{\mathrm{right}}(u)=0.3326u-44.089^{\circ} \\ \delta_{\mathrm{left}}(u)=0.2977u-38.295^{\circ}
    \end{cases}
    \label{eq:input_vs_steer_model}
    \end{equation}    

\subsubsection{Experimental Validation of the Bicycle-Model Geometry}

Reliable maneuvering in confined spaces (parking and intersection alignment, section~\ref{subsubsec:intersection_turning}) relies on the vehicle's kinematic geometry. Our planner assumes the bicycle model of section~\ref{subsec:kinematics}, in which the motion is fully determined by the wheelbase $L$ and the front steering angle $\delta$ (for example $R_{\mathrm{min}} = L / \tan \delta_{\mathrm{max}}$, Eq.~\ref{eq:rmin}). Before building maneuver planning on this assumption, we tested whether the relation actually holds on our physical vehicle, how large the deviations are, and whether they are random or systematic. In short: how does the real vehicle behave compared with the model?
The dimensions of the vehicle used in the tests are listed in Table~\ref{tab:vehicle_dims}. The wheelbase $L = 25.0\pm 1.0 \, \mathrm{cm}$ is the only parameter entering the model prediction.

\begin{figure}[H]
    \centering
    \includegraphics[width=0.4\linewidth]{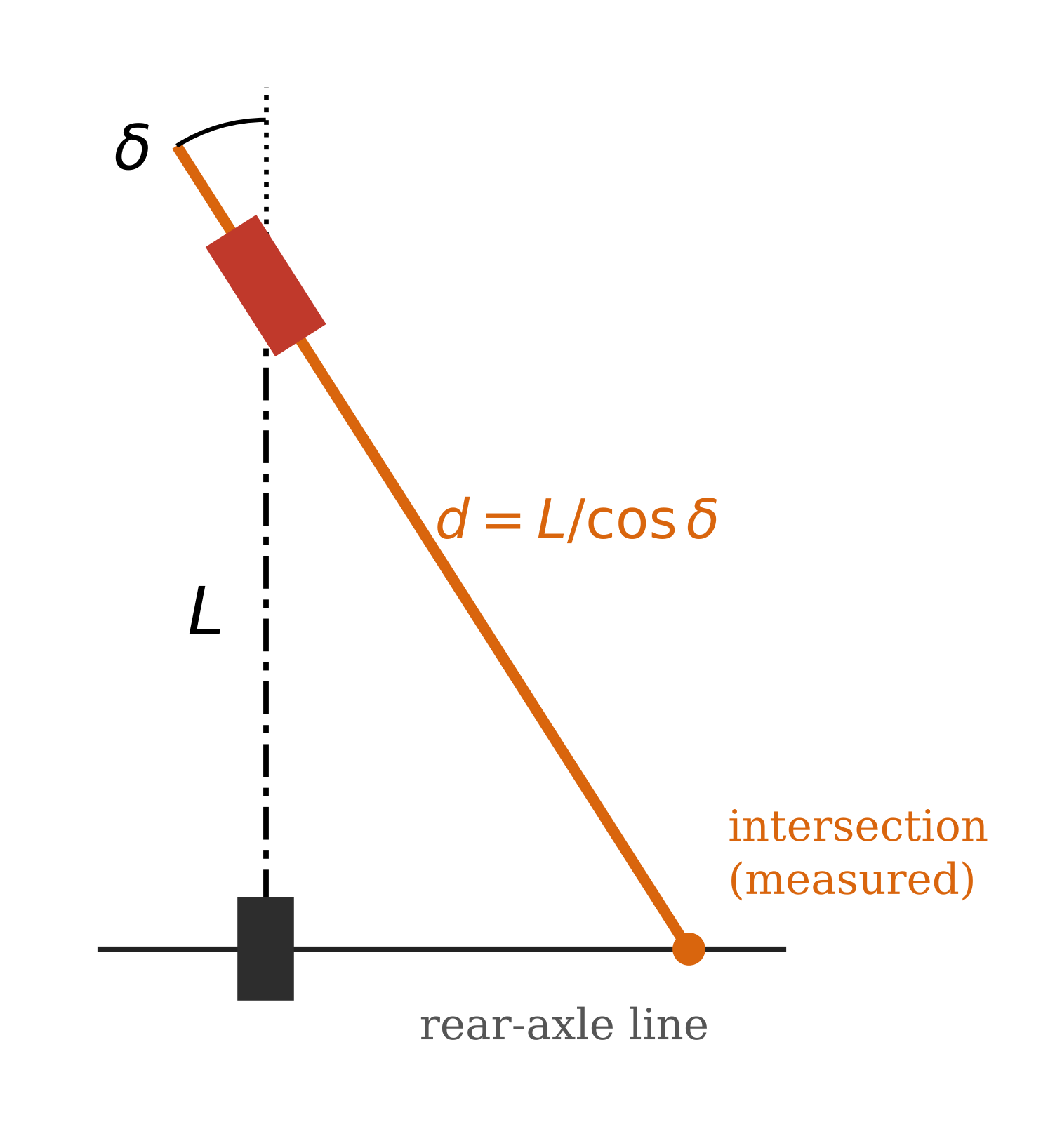}
    \caption{Measured quantity: the segment $d$ from the front wheel, along its rolling direction, to the rear-axle line.}
    \label{fig:bicycle_validation}
  \end{figure}

\begin{figure}[H]
        \centering
        \includegraphics[width=0.95\linewidth]{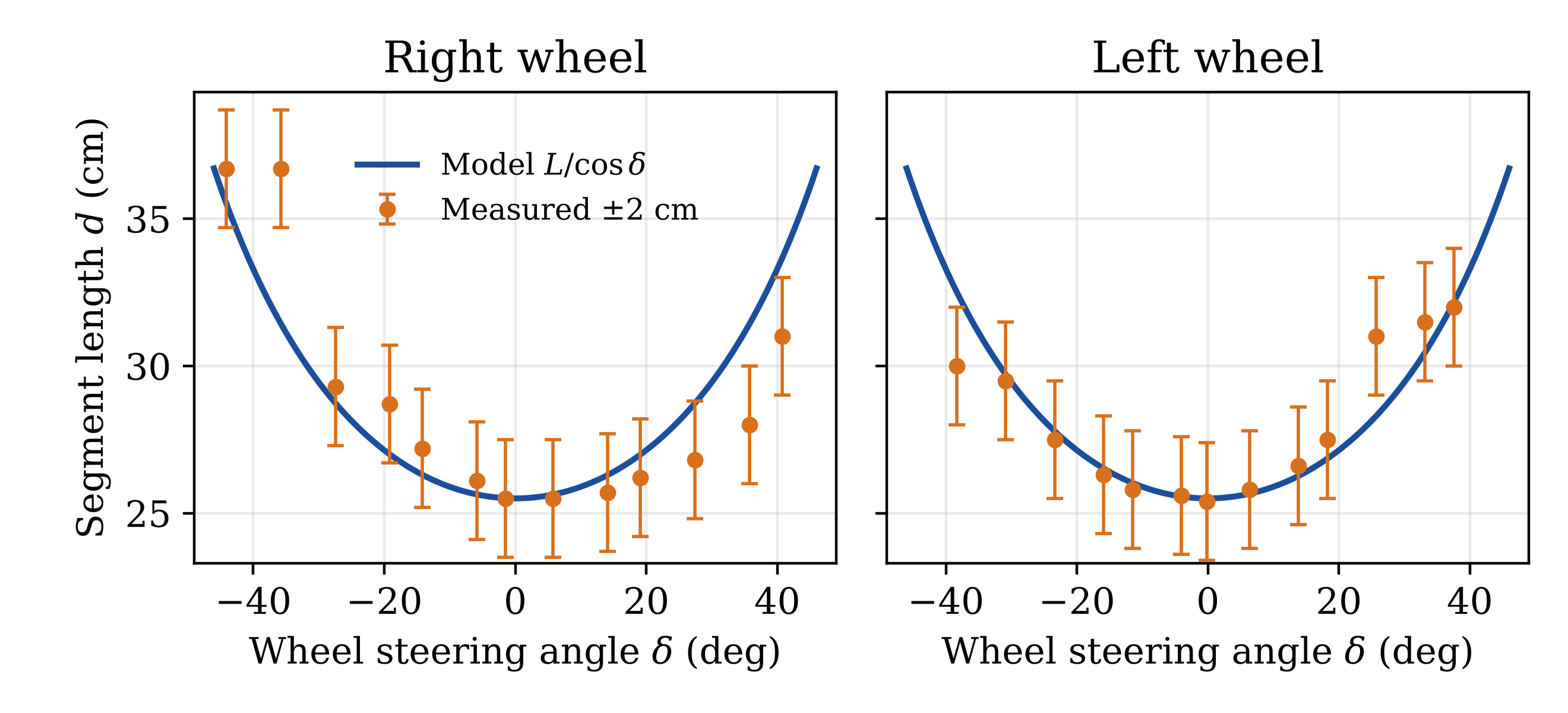}
        \caption{Model prediction $L/\cos\delta$ (line) and measured lengths with $\pm2$~cm error bars (points).}
        \label{fig:bicycle_residual_plot}
    \end{figure}

The steering command was swept from 0 to 255 at 13 setpoints. The steering angle of each front wheel follows the linear calibration of the previous section (section~\ref{subsec:ackermann_valid}).
At every setpoint and for each front wheel separately, the rolling direction of the wheel was extended on the ground until it met the line through the two rear wheels (the rear-axle line), as illustrated in Fig.~\ref{fig:bicycle_validation}. The length $d$ of this segment, from the wheel to the intersection point, was measured with an estimated uncertainty of $\pm 2 \, \mathrm{cm}$. One measurement was taken per wheel and setpoint ($n = 13$ per wheel).
  In the bicycle model the wheelbase $L$, the rear-axle line and the wheel's heading form a right triangle, so the predicted length is as Eq.~\ref{eq:predict_d}, which equals $L = 25.5\pm 1.0 \, \mathrm{cm}$ at $\delta=0$ and grows as the steering angle increases. The measured value was compared with this prediction at the calibrated angle of each wheel. Table~\ref{tab:residuals} lists the model prediction, the measured length and the residual at every setpoint, and Fig.~\ref{fig:bicycle_residual_plot} compares them graphically.
  \begin{equation}
    d_{\mathrm{model}}(\delta)=L\sec(\delta)
    \label{eq:predict_d}
  \end{equation}
  The results show that the mean model residual was only $-0.05 \, \mathrm{cm}$ for the right wheel and $-0.09 \, \mathrm{cm}$ for the left wheel, with no statistically significant mean bias ($p = 0.941$ and $p = 0.776$, respectively). The RMSE was $2.09\, \mathrm{cm}$ for the right wheel and $1.09\, \mathrm{cm}$ for the left wheel, while 21 of the 26 measurements (81\%) were within the $\pm 2\, \mathrm{cm}$ measurement tolerance (Table~\ref{tab:residual_stats}).
\begin{table}[H]
    \centering
    \caption{Model prediction, measured length and residual $e = d_{\mathrm{model}} - d_{\mathrm{meas}}$ per steering command $u$. Values with $|e| > 2$~cm are in bold.}
    \label{tab:residuals}
    \small
    \setlength{\tabcolsep}{4pt}
    \renewcommand{\arraystretch}{1.1}
    \begin{tabular}{@{}r rrrr rrrr@{}}
    \toprule
    & \multicolumn{4}{c}{\textbf{Right wheel}} & \multicolumn{4}{c}{\textbf{Left wheel}} \\
    \cmidrule(lr){2-5} \cmidrule(l){6-9}
    $u$ & $\delta$ ($^\circ$) & Model (cm) & Meas. (cm) & $e$ (cm) & $\delta$ ($^\circ$) & Model (cm) & Meas. (cm) & $e$ (cm) \\ \midrule
    0   & \textminus44.09 & 35.50 & 36.7 & \textminus1.20 & \textminus38.30 & 32.49 & 30.0 & \textbf{+2.49} \\
    25  & \textminus35.77 & 31.43 & 36.7 & \textbf{\textminus5.27} & \textminus30.85 & 29.70 & 29.5 & +0.20 \\
    50  & \textminus27.46 & 28.74 & 29.3 & \textminus0.56 & \textminus23.41 & 27.79 & 27.5 & +0.29 \\
    75  & \textminus19.14 & 26.99 & 28.7 & \textminus1.71 & \textminus15.97 & 26.52 & 26.3 & +0.22 \\
    90  & \textminus14.15 & 26.30 & 27.2 & \textminus0.90 & \textminus11.50 & 26.02 & 25.8 & +0.22 \\
    115 & \textminus5.84  & 25.63 & 26.1 & \textminus0.47 & \textminus4.06  & 25.56 & 25.6 & \textminus0.04 \\
    128 & \textminus1.52  & 25.51 & 25.5 & +0.01 & \textminus0.19  & 25.50 & 25.4 & +0.10 \\
    150 & 5.80  & 25.63 & 25.5 & +0.13 & 6.36  & 25.66 & 25.8 & \textminus0.14 \\
    175 & 14.12 & 26.29 & 25.7 & +0.59 & 13.80 & 26.26 & 26.6 & \textminus0.34 \\
    190 & 19.11 & 26.99 & 26.2 & +0.79 & 18.27 & 26.85 & 27.5 & \textminus0.65 \\
    215 & 27.42 & 28.73 & 26.8 & +1.93 & 25.71 & 28.30 & 31.0 & \textbf{\textminus2.70} \\
    240 & 35.73 & 31.41 & 28.0 & \textbf{+3.41} & 33.15 & 30.46 & 31.5 & \textminus1.04 \\
    255 & 40.72 & 33.65 & 31.0 & \textbf{+2.65} & 37.62 & 32.19 & 32.0 & +0.19 \\
    \bottomrule
    \end{tabular}
    \end{table}
    
    \begin{table}[H]
    \centering
    \caption{Summary statistics of the residuals (all lengths in cm unless noted).}
    \label{tab:residual_stats}
    \small
    \renewcommand{\arraystretch}{1.1}
    \begin{tabular}{@{}lccc@{}}
    \toprule
    \textbf{Statistic} & \textbf{Right wheel} & \textbf{Left wheel} & \textbf{Pooled} \\ \midrule
    Sample size $n$ & 13 & 13 & 26 \\
    Mean residual $\bar{e}$ (cm) & \textminus0.05 & \textminus0.09 & \textminus0.07 \\
    Std. deviation $s$ (cm) & 2.18 & 1.13 & 1.70 \\
    Std. error of mean (cm) & 0.60 & 0.31 & 0.33 \\
    95\% CI of mean (cm) & [\textminus1.36, 1.27] & [\textminus0.77, 0.59] & [\textminus0.75, 0.62] \\
    $t$-test vs.\ zero, $p$-value & 0.941 & 0.776 & 0.839 \\
    90\% CI of mean (cm) & [\textminus1.12, 1.03] & [\textminus0.65, 0.47] & [\textminus0.64, 0.50] \\
    Equivalence to $\pm2$~cm (TOST~\cite{lakens2017} $p$) & 0.004 & $<$0.001 & $<$0.001 \\
    Bootstrap~\cite{efron1979} 95\% CI of mean (cm) & [\textminus1.23, 1.04] & [\textminus0.69, 0.50] & [\textminus0.73, 0.55] \\
    RMSE (cm) & 2.09 & 1.09 & 1.67 \\
    MAE (cm) & 1.51 & 0.66 & 1.09 \\
    Max $|e|$ (cm) & 5.27 & 2.70 & 5.27 \\
    Limits of agreement~\cite{bland1986} (cm) & [\textminus4.31, 4.22] & [\textminus2.30, 2.12] & [\textminus3.40, 3.26] \\
    Points within $\pm2$~cm [95\% CI, \%] & 10/13 = 77\% [46, 95] & 11/13 = 85\% [55, 98] & 21/26 = 81\% [61, 93] \\
    Slope of $e$ vs.\ $\delta$ (cm/deg) & $+0.069 \pm 0.012$ & $-0.032 \pm 0.010$ & --- \\
    Correlation $r$ ($p$-value) & 0.86 ($<$0.001) & \textminus0.69 (0.009) & --- \\
    Shapiro--Wilk~\cite{shapiro1965} $p$ (normality of $e$) & 0.440 & 0.020 & 0.020 \\
    $|z| \le 2$ (uncertainty-weighted) & 12/13 & 13/13 & 25/26 \\
    $\chi^2/n$ ($p$-value) & 0.77 (0.689) & 0.22 (0.999) & 0.49 (0.985) \\
    Best-fit angular offset $\varphi$ (deg) & $-5.4 \pm 1.1$ & $+2.9 \pm 0.9$ & --- \\
    RMSE after offset (cm) & 1.21 & 0.80 & --- \\
    \bottomrule
    \end{tabular}
    \end{table}
    
\begin{figure}[H]
    \centering
    \includegraphics[width=0.95\linewidth]{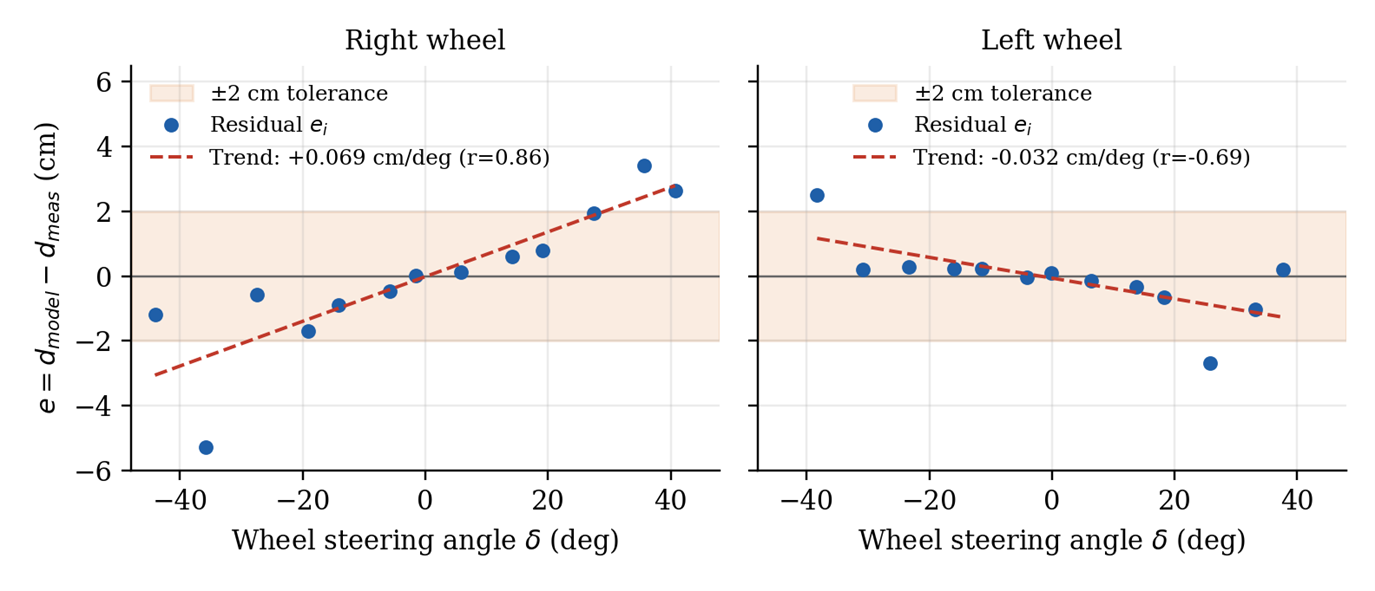}
    \caption{Residual $e = d_{\mathrm{model}} - d_{\mathrm{meas}}$ versus wheel steering angle $\delta$ for the right (left panel) and left (right panel) front wheels. The shaded band marks the $\pm2$~cm measurement tolerance and the dashed line is the linear trend of the residuals (right wheel: $+0.069$~cm/deg, $r=0.86$; left wheel: $-0.032$~cm/deg, $r=-0.69$).}
    \label{fig:residual_vs_angle}
\end{figure}

Although the average error was small, the residuals showed a systematic dependence on steering angle, as shown in Fig.~\ref{fig:residual_vs_angle}. The right-wheel error increased with steering angle ($r = 0.86$), while the left-wheel error showed an opposite trend ($r = -0.69$); the fitted slopes and correlations are listed in Table~\ref{tab:residual_stats}. This indicates that the main deviation is not purely random and is likely related to a small offset between the calibrated and actual wheel angles. Consistently, fitting a single angular offset $\varphi$ per wheel reduces the RMSE from 2.09 to 1.21~cm (right wheel) and from 1.09 to 0.80~cm (left wheel), with best-fit offsets of $-5.4^\circ$ and $+2.9^\circ$, respectively (Table~\ref{tab:residual_stats}).

Overall, the measured wheel directions agree with the bicycle-model geometry within the measurement tolerance for most setpoints, with a systematic residual consistent with small angular calibration offsets. The bicycle model is therefore used as an approximate model. This experiment checks each wheel separately; whether the vehicle as a whole follows circular arcs is examined in the next subsection through the Ackermann condition.

\paragraph{Implication for control.}
Since $R = L/\tan\delta$, the sensitivity
\begin{equation}
    \frac{dR}{d\delta}=\frac{L}{\sin^2(\delta)}
\end{equation}
is about $0.9$--$1.2\, \mathrm{cm}$ per degree over the calibrated extremes of Table~\ref{tab:rmin}, i.e. by roughly $3.5\%$ of $R$. This holds in both steering directions: at the negative limit ($\delta= -44.1^\circ$, right wheel) $R_\text{min} = 26.3 \,\mathrm{cm}$, while at the positive limit ($\delta = +40.7^\circ$) $R_\text{min} = 29.6 \,\mathrm{cm}$. An angular offset of $3$--$5^\circ$, as suggested by Fig.~\ref{fig:residual_vs_angle}, therefore changes $R_\text{min}$ by roughly $3$--$6 \,\mathrm{cm}$.

\subsubsection{Experimental Validation of the Ackermann Model}
Although the vehicle was designed and controlled under the assumption of an Ackermann steering model, experimental testing revealed significant deviations between the predicted and observed vehicle motion. Therefore, the validity of the Ackermann model for the present vehicle configuration was investigated experimentally.

The calibration in this section validates the steering mechanism through the equivalent single-track angle of Eq.~\ref{eq:nonlin_steering_equation}. This quantity can be computed for any pair of wheel angles, including parallel (non-Ackermann) steering, so agreement between $\delta_\text{comb}$ and the commanded angle shows that the equivalent angle is calibrated. It does not show that the two wheels obey the Ackermann condition. Likewise, the observation that the inner and outer wheels turn by different angles shows only that the wheels are not parallel; it does not show that the difference is the correct one.
The geometric Ackermann condition relates the two angles to the vehicle dimensions:
\begin{equation}
    \left| \cot |\delta_o|-\cot |\delta_i|\right| =\frac{w}{L}
    \label{eq:deltacotabsval}
\end{equation}
With the front track width $w = 18 \pm 0.5 \,\mathrm{cm}$ and the wheelbase $L = 25.5 \pm 1.0 \,\mathrm{cm}$, the required value is $w/L = 0.706 \pm 0.034$.

\begin{figure}[H]
    \centering
    \includegraphics[width=0.70\linewidth]{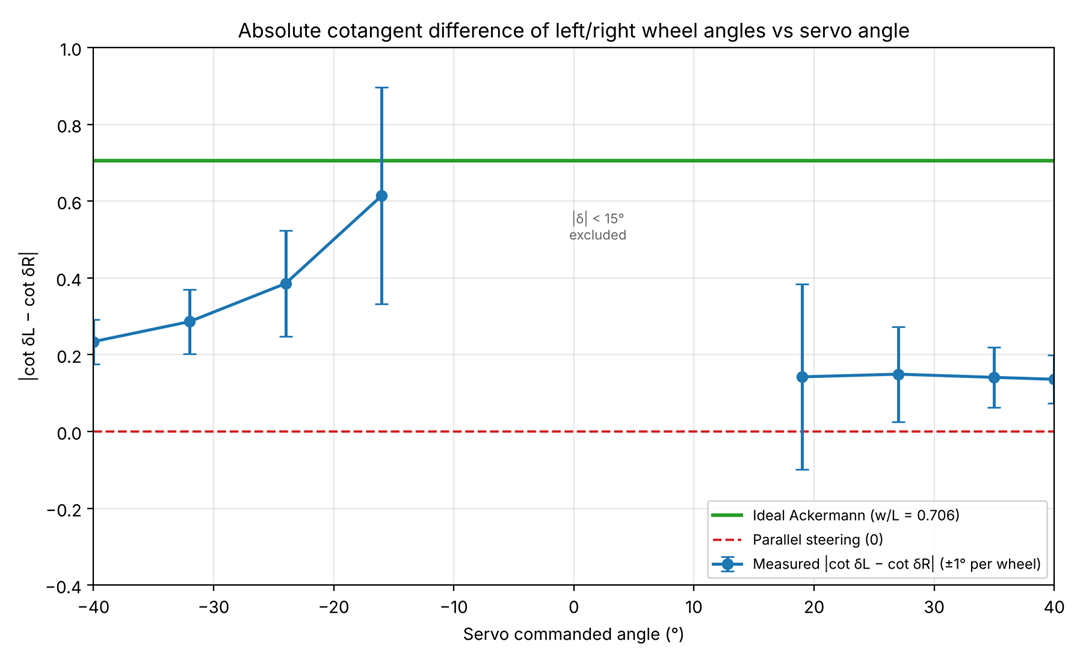}
    \caption{Measured $\left|\cot|\delta_L| -\cot|\delta_R|\right|$ (blue, with $\pm 1^\circ$ uncertainty bars) against the servo commanded angle from $-40^\circ$ to $+40^\circ$. The green line is the ideal Ackermann value $w/L = 0.706$ and the dashed red line is parallel steering (0). The measured points lie below the ideal line at every command. On the negative side the outer wheel turns more than the inner wheel (negative sign in Table~\ref{tab:ackermann_geometry}), which the absolute value does not show.}
    \label{fig:abs_cotangent}
\end{figure}

The calibrated mean wheel angles of the 13 setpoints ($u=0,\dots,255$) were used, with an angular uncertainty of about $1^\circ$ per wheel (see the calibration above). Each setpoint is plotted against the servo commanded angle, which spans $-40^\circ$ to $+40^\circ$ ($u = 0$ corresponds to $-40^\circ$, $u = 128$ corresponds to $0^\circ$ and $u = 255$ corresponds to $+40^\circ$). The quantity plotted is the absolute value $\left|\cot|\delta_L|-\cot|\delta_R|\right|$ (Eq.~\ref{eq:deltacotabsval}) of the difference between the cotangents of the left and right wheel angles, which equals $\left|\cot|\delta_o| - \cot|\delta_i|\right|$. Ideal Ackermann steering gives the constant value $w/L = 0.706$ on both sides of neutral and parallel steering gives 0. The absolute value does not show which wheel turns more; this sign is kept in Table~\ref{tab:ackermann_geometry}, where a negative value means that the outer wheel turns more than the inner wheel. The inner wheel was taken as the right wheel for positive commands and the left wheel for negative commands, consistent with the smaller right-wheel radius at $u = 255$ in Table~\ref{tab:rmin}. For each setpoint the measured value was compared with $w/L$, and the outer-wheel angle required by the Ackermann condition was computed from the measured inner-wheel angle. The Ackermann ratio is the measured value divided by $w/L$: $100\%$ is ideal Ackermann and $0\%$ is parallel steering. Setpoints with a wheel angle below $15^\circ$ (servo commands from about $-11^\circ$ to $+14^\circ$) are excluded from the chart and the table, because the difference of two cotangents is dominated by noise near straight-ahead.
Table~\ref{tab:ackermann_geometry} and Fig.~\ref{fig:abs_cotangent} summarise the comparison. Fig.~\ref{fig:abs_cotangent} is the main result: it shows the measured $\left|\cot|\delta_L| -\cot|\delta_R|\right|$ against the servo command together with the ideal Ackermann line ($0.706$) and the parallel-steering line ($0$).

\begin{table}[H]
    \centering
    \caption{Measured versus required Ackermann geometry. A negative sign means the outer wheel turns more than the inner wheel.}
    \label{tab:ackermann_geometry}
    \begin{tabular}{c c c c c c c}
        \hline
        $u$ & $\delta_L$ (°) & $\delta_R$ (°) &
        $\cot\delta_o-\cot\delta_i$ &
        Required $\delta_o$ (°) &
        Measured $\delta_o$ (°) &
        Ackermann (\%) \\
        \hline

        0   & $-38.3$ & $-44.1$ & $-0.234 \pm 0.06$ & $26.9$ & $44.1$ & $-33$ \\
        25  & $-30.9$ & $-35.8$ & $-0.286 \pm 0.08$ & $22.8$ & $35.8$ & $-41$ \\
        50  & $-23.4$ & $-27.5$ & $-0.385 \pm 0.14$ & $18.3$ & $27.5$ & $-55$ \\
        75  & $-16.0$ & $-19.1$ & $-0.614 \pm 0.28$ & $13.4$ & $19.1$ & $-87$ \\
        190 & $+18.3$ & $+19.1$ & $+0.142 \pm 0.24$ & $15.6$ & $18.3$ & $+20$ \\
        215 & $+25.7$ & $+27.4$ & $+0.149 \pm 0.12$ & $20.8$ & $25.7$ & $+21$ \\
        240 & $+33.2$ & $+35.7$ & $+0.141 \pm 0.08$ & $25.5$ & $33.2$ & $+20$ \\
        255 & $+37.6$ & $+40.7$ & $+0.136 \pm 0.06$ & $28.2$ & $37.6$ & $+19$ \\

        \hline
    \end{tabular}
\end{table}

\begin{figure}[H]
    \centering
    \includegraphics[width=0.70\linewidth]{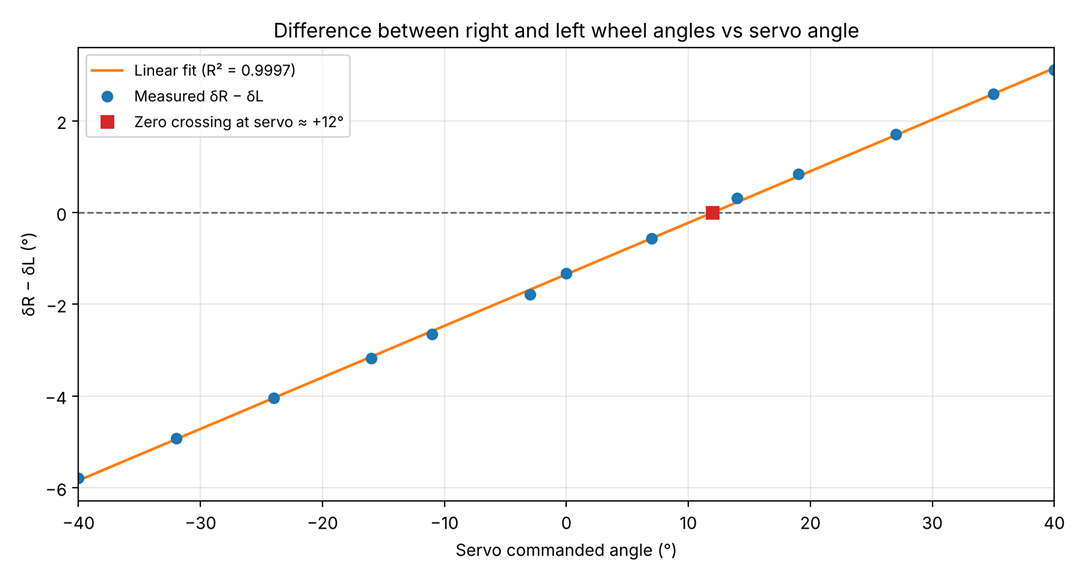}
    \caption{Difference between the right and left wheel angles, $\delta_R-\delta_L$, against the servo commanded angle (points) with a linear fit. The difference changes linearly with the command ($R^2 = 0.9997$) and crosses zero near $+12^\circ$ ($u \approx 166$), not at $0^\circ$.}
    \label{fig:right_left_servo}
\end{figure}

At the steering limits the right wheel turns about $8\%$ more than the left wheel at $+40^\circ$ ($40.7^\circ$ against $37.6^\circ$) and about $15\%$ more at $-40^\circ$ ($44.1^\circ$ against $38.3^\circ$), and the difference $\delta_R-\delta_L$ changes linearly with the servo command and crosses zero near $+12^\circ$ ($u \approx 166$), not at $0^\circ$ (Fig.~\ref{fig:right_left_servo}). This pattern is typical of two wheels driven with unequal effective gain plus a small neutral offset (the right wheel reads $-1.5^\circ$ at $u = 128$). It is not the pattern produced by a steering trapezoid. The equivalent-angle result of Eq.~\ref{eq:nonlin_steering_equation} is therefore still valid for the bicycle model, but the physical mechanism does not implement Ackermann geometry.
Consequence. The two front wheels do not roll about a common centre on the rear-axle line, so the outer wheel is partly dragged sideways, most strongly near the steering limits. This is expected to cause tyre slip, a turning radius different from $L / \tan\delta$, and different behaviour in left and right turns. These effects were not measured here. Because errors of this kind accumulate in open-loop maneuvers, the feedback correction of Section~\ref{subsubsec:intersection_turning} remains necessary.

\subsubsection{Turning-Radius Estimate}
\label{subsubsec:turning_radius}

For the maneuver planning of section~\ref{subsec:steering_limit} we convert the validated wheel geometry into a turning radius using the standard bicycle-model relation
\begin{equation}
    R_{\mathrm{wheel}}(\delta) = \frac{L}{\tan\delta},
    \label{eq:R_wheel}
\end{equation}
applied directly to each front wheel's own calibrated angle $\delta$, rather than to the single equivalent bicycle angle of Eq.~\ref{eq:nonlin_steering_equation}. This gives the physical radius traced by that wheel, which is what determines the space the vehicle occupies while turning.

\begin{figure}[H]
    \centering
    \includegraphics[width=0.8\linewidth]{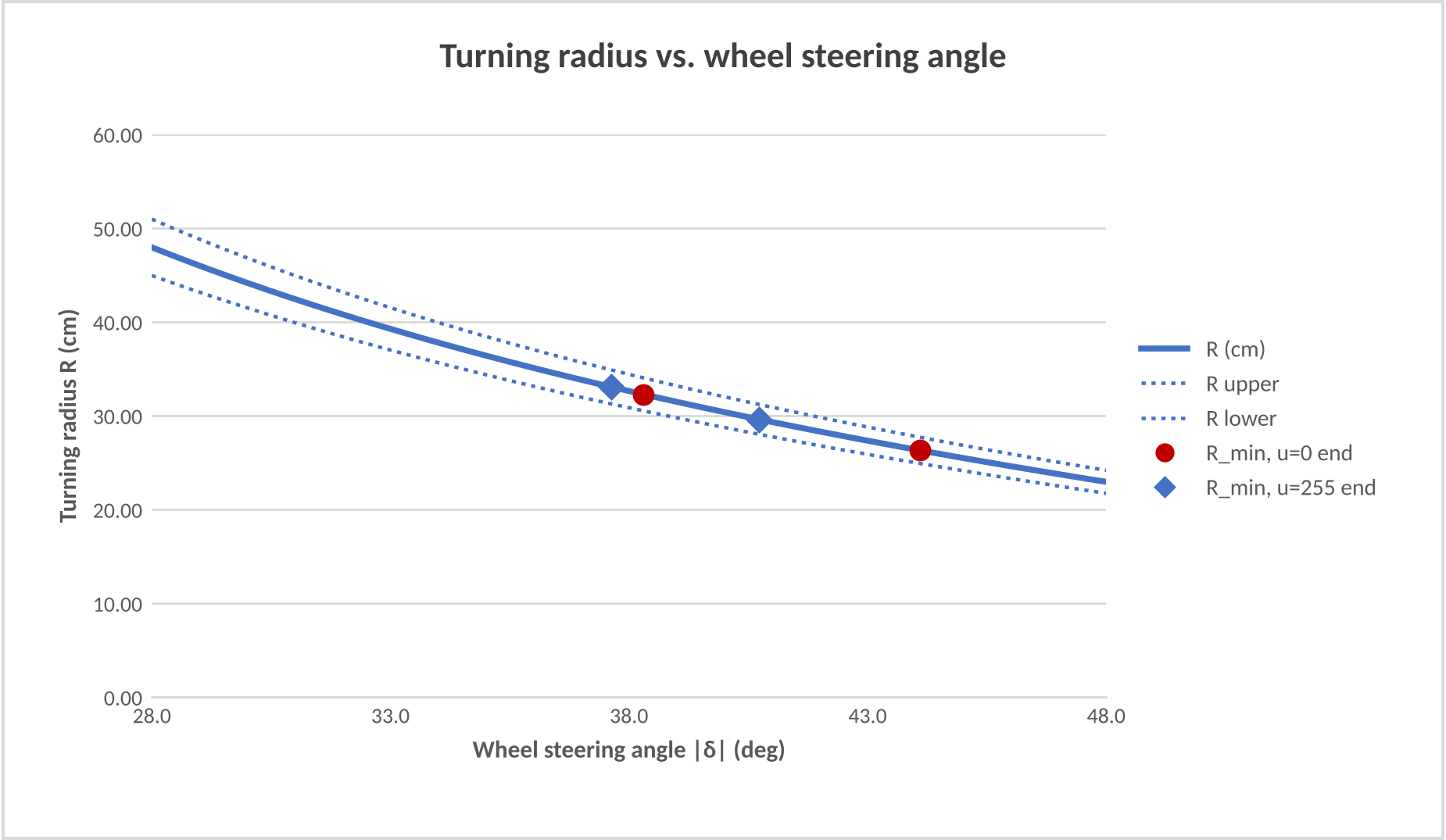}
    \caption{Turning radius $R = L/\tan\delta$ over the wheel steering angle $|\delta| = 28$--$48^\circ$, with the propagated uncertainty band (R\textsubscript{upper}/R\textsubscript{lower}) and the four extreme operating points ($u = 0$ and $u = 255$ for each wheel, plotted at $|\delta|$).}
    \label{fig:turning_radius}
\end{figure}

\paragraph{Why the minimum matters.}
The space required for a turn is set by the smallest achievable radius, which occurs at the largest steering angle the linkage can reach. Because each front wheel has its own calibrated angle, and the two ends of the steering range do not give the same angle, we evaluate $R$ at both extreme calibrated angles of each wheel in Table~\ref{tab:residuals}: the minimum angle ($\delta_{\mathrm{min}}$, at $u=0$) and the maximum angle ($\delta_{\mathrm{max}}$, at $u=255$), consistent with Eq.~\ref{eq:rmin}. The wheelbase tolerance ($\sigma_L = 0.5$~cm) and the angular calibration uncertainty ($\approx 1^\circ$, section~\ref{subsec:ackermann_valid}) are propagated through Eq.~\ref{eq:R_wheel} to obtain $\sigma_R$. The result is reported per wheel and per steering end, giving four radii, because the two wheels and the two steering directions are not interchangeable for this purpose.  The results are listed in Table~\ref{tab:rmin} and plotted in Fig.~\ref{fig:turning_radius}.

\begin{table}[H]
\centering
\caption{Estimated minimum turning radius for each front wheel at both extreme calibrated steering angles ($\delta_{\mathrm{min}}$ at $u=0$ and $\delta_{\mathrm{max}}$ at $u=255$) over the tested command range ($u \in [0, 255]$).}
\label{tab:rmin}
\renewcommand{\arraystretch}{1.15}
\begin{tabular}{@{}lccc@{}}
\toprule
\textbf{Wheel} & \textbf{Extreme angle $\delta$ (deg)} & \textbf{Setpoint} & \textbf{$R_{\mathrm{min}}$ (cm)} \\ \midrule
Right & $-44.09$ & $u=0$   & $26.3 \pm 1.4$ \\
Right & $+40.72$ & $u=255$ & $29.6 \pm 1.6$ \\
Left  & $-38.30$ & $u=0$   & $32.3 \pm 1.7$ \\
Left  & $+37.62$ & $u=255$ & $33.1 \pm 1.8$ \\
\bottomrule
\end{tabular}
\end{table}

\section{Future Work}
\label{sec:future_work}
While the current sensor fusion and adaptive Kalman filter architecture is designed to isolate and reject transient magnetic anomalies, several limitations remain regarding long-term calibration stability and dynamic environmental changes. Future development of the heading estimation and magnetometer calibration pipeline will focus on the following key areas:

\paragraph{Dynamic Motor-Current Compensation.}
The current magnetometer calibration (Stage 4) is entirely static; it estimates the hard- and soft-iron distortions caused by the physical chassis but does not account for the dynamic magnetic field generated by the vehicle's own DC motor. In future iterations, we plan to model this motor-induced field. By correlating the commanded motor PWM signal---or data from a dedicated current sensor---with the induced magnetic deviation, this dynamic hard-iron offset can be actively subtracted from the raw AK8963 readings before they enter the calibration buffer.

\paragraph{Autonomous Calibration Recovery.}
As demonstrated in the magnetic disturbance tests, a persistent, lasting anomaly (such as a strong magnet attached to the chassis or transitioning into a structurally different indoor environment) pollutes the active sliding window. Currently, this forces the Kalman filter into an extended dead-reckoning state and necessitates a manual reset and a full $360^\circ$ rotation to re-establish the baseline. Future work will investigate dual-buffer or multiple-hypothesis calibration schemes. By maintaining a shadow buffer of the disturbed field, the system could autonomously detect when a permanent shift has occurred and seamlessly transition to the new calibration ellipsoid without requiring human intervention or halting the vehicle.

\paragraph{Passive 3D Ellipsoid Population for Ground Vehicles.}
Because a planar ground vehicle does not naturally pitch or roll, the $z$-axis of the magnetometer experiences almost zero variance, forcing the system to rely on a constrained 2D calibration mode to avoid singular matrix inversions. This 2D mode is vulnerable to tilt errors if the vehicle encounters uneven terrain. A future improvement will involve harvesting the minor pitch and roll variations caused by longitudinal acceleration, hard braking, and suspension dynamics. By selectively filtering and storing these rare out-of-plane samples over extended driving periods, the algorithm could slowly accumulate enough $z$-axis variance to passively solve the full 3D ellipsoid without requiring the operator to manually tumble the vehicle in 3D space.

\paragraph{Full Localization.}
The set of sensors, actuators, and perception pipeline are theoretically enough to evaluate a robot full state: $[x,y,\psi]^T$. This will be developed in the future.

Additionally in the control section, because the Ackermann condition is not satisfied, the bicycle model describes the real vehicle less accurately than assumed: the path actually followed for a given steering command differs from the one predicted by $R = L / \tan \delta$, and the difference is not the same for left and right turns. Any algorithm whose result depends on a model-based prediction of the vehicle motion therefore inherits this error. For the current mechanism we recommend relying on feedback wherever possible and keeping model-based algorithms to a minimum.

If the vehicle is to keep this steering structure, a more accurate motion model should be identified from data instead of derived from geometry. We therefore propose to record the motion of the vehicle (displacement and rotation) in many experiments with different steering commands, speeds and turn directions, and to train a neural network on this supervised dataset to predict the motion from the vehicle state and command. Such a model would include the left--right asymmetry and wheel slip that the bicycle model does not contain, and could replace the bicycle model in the planner. This was not performed in this work.
\section*{Acknowledgments}

The authors wish to note that the submission of this manuscript was delayed due to prolonged disruptions to research activities caused by civil unrest and subsequent military conflict in Iran. These circumstances significantly restricted access to laboratory facilities and delayed manuscript preparation. The work presented here corresponds to the robotic system as completed in early 2025.

Additionally, the authors used AI language models to assist with English-language editing, \LaTeX{} formatting, and manuscript preparation, owing to limited fluency in English. All technical content, experiments, data, and conclusions are the authors' own, and the authors have reviewed the text and take full responsibility for its contents.

\newpage

\begin{center}
    \large\textbf{Author Contributions}
    
    \vspace{0.75em}
    \small
    \renewcommand{\arraystretch}{1.25}
    \begin{tabularx}{\textwidth}{|>{\raggedright\arraybackslash\bfseries}p{4.3cm}|>{\raggedright\arraybackslash}X|}
    \hline
    \textbf{Name(s)} & \textbf{Role and Contribution Summary} \\
    \hline
    
    Marya Sabouri Dodaran &
    Design of motion-geometry-based maneuvering algorithms for sign and signal response; experimental calibration and validation of the vehicle kinematic model; contributions to lane-keeping and cruise control \\
    \hline

    Ali Taheri Garakani &
    Implementation of AI-based lane detection model, and model data post-processing \\
    \hline

    Mandana Shabanzadeh Nasrollah-Abadi &
    Electronics, hardware design and integration \\
    \hline

    Yousef Ashrafi &
    ROS architecture and system communication \\
    \hline

    Mohammad Mahdi Azimi &
    Embedded firmware design, sensor data handling, heading estimation algorithm, actuator interfacing and control, and inter-processor communication system design \\
    \hline

    Erfan Nekouei Ghahfarokhi &
    Design and training of AI-based lane detection model \\
    \hline

    Amirhossein Kheiri Holighi &
    Sign and traffic light detection (vision algorithms) \\
    \hline

    Seyyed Sobhan Hosseini Hajibakandeh &
    Implementation of control algorithms to maintain lane position and prevent vehicle drift \\
    \hline

    Matin Jamshidi &
    Mechanical chassis, car body and drivetrain system design \\
    \hline

    Amirhossein Gholizadeh Behbahani &
    Implementation of an adaptive cruise control system for vehicle speed regulation under various conditions and collision prevention \\
    \hline

    Mohammad Hossein Moaiyeri &
    Team supervision \\
    
    \hline
    \end{tabularx}
\end{center}

\end{document}